\documentclass{article}

\usepackage{arxiv}

\usepackage[utf8]{inputenc} 
\usepackage[T1]{fontenc}    
\usepackage{hyperref}       
\usepackage{url}            
\usepackage{booktabs}       
\usepackage{amsfonts}       
\usepackage{amssymb}
\usepackage{nicefrac}       
\usepackage{microtype}      
\usepackage{graphicx}
\usepackage{natbib}
\usepackage{doi}
\usepackage{framed}
\usepackage[inline]{enumitem}
\usepackage{longtable}

\usepackage{tikz}
\newcommand{\tikzxmark}{%
\tikz[scale=0.23] {
    \draw[line width=0.7,line cap=round] (0,0) to [bend left=6] (1,1);
    \draw[line width=0.7,line cap=round] (0.2,0.95) to [bend right=3] (0.8,0.05);
}}
\newcommand{\tikzcmark}{%
\tikz[scale=0.23] {
    \draw[line width=0.7,line cap=round] (0.25,0) to [bend left=10] (1,1);
    \draw[line width=0.8,line cap=round] (0,0.35) to [bend right=1] (0.23,0);
}}

\title{Combining Machine Learning and Ontology: A Systematic Literature Review}

\date{February 19, 2024}	

\author{ \href{https://orcid.org/0000-0003-2030-5824}{\includegraphics[scale=0.06]{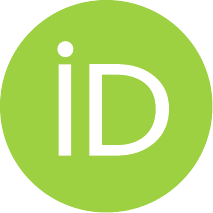}\hspace{1mm}Sarah Ghidalia} \\
	CIAD UMR 7533\\
	Université de Bourgogne, UB\\
	F-21000, Dijon, France \\
	\texttt{sarah.ghidalia@u-bourgogne.fr} \\
	\And
	\href{https://orcid.org/0000-0001-5521-0126}{\includegraphics[scale=0.06]{orcid.pdf}\hspace{1mm}Ouassila Labbani Narsis} \\
	CIAD UMR 7533\\
	Université de Bourgogne, UB\\
	F-21000, Dijon, France \\
	\texttt{ouassila.narsis@u-bourgogne.fr} \\
    \And
	\href{https://orcid.org/0000-0001-9969-2569}{\includegraphics[scale=0.06]{orcid.pdf}\hspace{1mm}Aurélie Bertaux} \\
	CIAD UMR 7533\\
	Université de Bourgogne, UB\\
	F-21000, Dijon, France \\
	\texttt{aurelie.bertaux@u-bourgogne.fr} \\
    \And
	\href{https://orcid.org/0000-0002-8118-5005}{\includegraphics[scale=0.06]{orcid.pdf}\hspace{1mm}Christophe Nicolle} \\
	CIAD UMR 7533\\
	Université de Bourgogne, UB\\
	F-21000, Dijon, France \\
	\texttt{christophe.nicolle@u-bourgogne.fr} \\
}

\hypersetup{
pdftitle={Combining Machine Learning and Ontology: A Systematic Literature Review},
pdfsubject={cs.AI},
pdfauthor={Sarah Ghidalia, Ouassila Labbani Narsis, Aurélie Bertaux, Christophe Nicolle},
pdfkeywords={machine learning, ontology, artificial reasoning, hybrid reasoning},
}

\begin{document}
\maketitle

\begin{abstract}
	Motivated to explore the process of combining inductive and deductive reasoning, we conducted a systematic literature review of articles investigating the integration of machine learning and ontologies. The objective was to identify diverse techniques incorporating inductive reasoning (performed by machine learning) and deductive reasoning (performed by ontologies) into artificial intelligence systems. Our review, which included the analysis of 128 studies, allowed us to identify three main categories of hybridization between machine learning and ontologies: \textit{learning-enhanced ontologies}, \textit{semantic data mining}, and \textit{learning and reasoning systems}. We provide a comprehensive examination of all these categories, emphasizing the various machine learning algorithms utilized in the studies. Furthermore, we compared our classification with similar recent work in the field of hybrid AI and neuro-symbolic approaches.
\end{abstract}


\section{Introduction}
\label{ref:introduction}
Artificial intelligence (AI) has become part of our daily lives, transforming every economic sector, from industry 4.0 to healthcare and smart cities. However, there is still a lack of consensus among researchers regarding the precise definition of AI, a term coined more than half a century ago~\citep{nilsson_quest_2009,wang_defining_2019}. For example,~\cite{minsky_society_1986} suggests that AI refers to the capability of machines to \textit{solve complex problems}.~\cite{dobrev_definition_2012} compares AI to human beings, defining it as \textit{a program which in an arbitrary world will cope not worse than a human}, which recalls the original definition of~\cite{mccarthy_proposal_1955}. On the other hand,~\cite{kaplan_siri_2019} defines AI as \textit{a system's ability to correctly interpret external data, to learn from such data, and to use those learnings to achieve specific goals and tasks through flexible adaptation}, potentially narrowing AI to the realm of machine learning. Alternatively,~\cite{wang_defining_2019} emphasizes AI's capability to adapt to its environment, even with limited knowledge and resources. Another perspective, similar to~\cite{russell_artificial_2009}'s, suggests defining AI as the scientific field that enables machines to perceive, understand, and interact with the real world in a way that is very close to human beings.

To understand this perspective, we can call upon the allegory of the cave exposed by~\cite{plato_republic_1888}. Like the prisoners chained to the bottom of the cave, who only perceive shadows and echoes of the intelligible world, machines have a perception of our world limited to the data they are provided. 
How can we enable machines to have an impact on the tangible world if they can't  make sense of the various aspects, complexities, and nuances of the real world? Gradually, we would have to get machines out of the cave in which they are chained. The initial step involves imparting problem-solving thinking abilities to machines. In this endeavor, the human cognitive process primarily relies on two forms of reasoning: induction and deduction. Inductive reasoning facilitates the discovery of general knowledge (such as laws, theorems, correlations, etc.) from specific observations. Deductive reasoning, on the other hand, permits the application of pre-existing general knowledge to specific instances~\citep{rafanelli_position_2022}.
In the 19th century, Charles Sanders Peirce identified a third type of reasoning called abduction, which is employed to generate hypotheses that explain specific observations~\citep{roudaut_comment_2017, rafanelli_position_2022}. Abduction holds substantial scientific significance as it pertains to issues of causality, explainable artificial intelligence (XAI), and potentially even trustworthiness. However, this study will not focus on abduction, but rather on the combination of inductive and deductive reasoning for problem-solving purposes. Nevertheless, the matter of explainability remains intriguing and will be addressed in the study's conclusion.

Socrates, Plato's disciple, defines \emph{induction} as a way of reasoning that consists in drawing a general conclusion from several particular cases. Inductive reasoning is a form of ampliative reasoning, i.e. one draws conclusions that go beyond the information contained in the premises \citep{roudaut_comment_2017}.
Inductive reasoning is close to mechanisms of machine learning: establishing a reasoning (model) from explicit facts (experiments). Thus, the model is not explicitly written, on the contrary, it is deduced from the input data in order to extract information (general laws). 
In the opposite, deductive reasoning is based on syllogism defined by Aristotle  such as a
\textit{speech (logos) in which, certain things having been supposed, something different from those supposed results of necessity because of their being so\footnote{Prior Analytics I.2, 24b18–20}.} In other words, deductive reasoning is the ability to draw conclusions about individual facts (experiences) from generic knowledge (general law). When Aristotle writes \textit{things supposed} this corresponds to the premise of the argument, and when he writes \textit{results of necessity} this corresponds to the conclusion of the argument~\citep{smith_aristotles_2020}. Within the field of AI, deductive reasoning is primarily associated with symbolic approaches, commonly referred to as Good Old-Fashioned AI (GOFAI)~\citep{haugeland_artificial_1989}. GOFAI encompasses a range of techniques including knowledge-based systems (e.g., expert systems), multi-agent systems, and constraint-based reasoning systems. These approaches leverage symbolic tools such as knowledge graphs, logical rules, ontologies and algebraic computation to facilitate deductive reasoning. We have placed particular focus on ontologies due to their ability to formalize knowledge by establishing associations between a knowledge graph and logic rules using inference engines. These deductive reasoning engines can use the axioms describing the concepts in the TBox (Terminology Box - general laws) to deduce new knowledge on the ABox (Assertional Box - specific facts) part of the ontology. 
In modern information systems, ontologies, such as formal and explicit specifications of shared conceptualizations~\citep{guarino_what_2009}, prove to be highly valuable. They address the need for data interoperability and the formalization of business rules, which are essential aspects of contemporary information systems. 

The exploration of the fusion between machine learning and ontologies arises from the aim to investigate the mechanisms that facilitate the integration of inductive reasoning with deductive reasoning. 
Unlike recent advancements in neuro-symbolic approaches~\citep{hitzler_neuro-symbolic_2022, garcez_neurosymbolic_2020, henry_kautz_third_2020, van_bekkum_modular_2021}, our study did not exclusively focus on neural networks. Therefore, we chose to concentrate on the study of ontologies, as they encompass two vital categories of symbolic methods: knowledge graphs and logic rules. In our review, we ensured to highlight these two concepts to assist readers interested in either subject in navigating the research more effectively. 

The main question addressed in this systematic literature review is whether it is possible to combine these two paradigms, and if so, how? The objective is to study how machine learning methods and ontologies can be effectively combined. In this study, our focus is on providing an overview of techniques that integrate learning and reasoning to realize a hybrid AI system capable of learning, building knowledge, and performing reasoning to tackle complex tasks and simulate the human cognitive process.

The Systematic Literature Review (SLR) is a research method based on the identification, evaluation, and interpretation of all relevant research results related to a particular topic area. It is a very popular analysis tool in the medical field, and was then adapted in computer science by~\cite{kitchenham_systematic_2010} who define it as \textit{a form of secondary study that uses a well-defined methodology to identify, analyze and interpret all available evidence related to a specific research question in a way that is unbiased and (to a degree) repeatable}~\citep{kitchenham_guidelines_2007}. 
By \emph{secondary study}, the authors mean \textit{a study that reviews all the primary studies relating to a specific research question with the aim of integrating/synthesizing evidence related to a specific research question}. In this context, a \emph{primary study} describes new original research and aims to answer questions that haven’t been answered or even asked before.
The main objectives of this SLR are:
\begin{enumerate*}[label=(\alph*)]
  \item providing an overview of existing approaches combining ontology and machine learning,
  \item understanding the motivations of each research work and addressed issues,
  \item identifying the weaknesses and difficulties encountered, and
  \item facilitating the positioning of new studies in this field.
\end{enumerate*}

To the best of our knowledge, a comprehensive overview encompassing the combination of ontology and machine learning techniques has not been previously conducted. While there have been studies focusing on specific aspects such as ontology learning~\citep{al-aswadi_automatic_2020, khadir_ontology_2021}, semantic data mining~\citep{dou_semantic_2015, sirichanya_semantic_2021}, and more recently, neuro-symbolic~\citep{hitzler_neuro-symbolic_2022, garcez_neurosymbolic_2020, henry_kautz_third_2020, van_bekkum_modular_2021}\footnote{Note that neuro-symbolic methods are not limited to ontologies}. There is no systematic review providing a holistic mapping of the different approaches that integrate ontologies and machine learning. Therefore, this study aims to fill this gap by providing a comprehensive overview of the combinations of ontology and machine learning, shedding light on the potential synergies and insights gained from their integration.

This document is structured as follows. Section~\ref{S:2} describes the used methodology to conduct this systematic literature review. Section~\ref{S:3} presents the overview results, and sections~\ref{knowledge_aquisition_discovery},~\ref{semantic_data_mining} and~\ref{combining_learning_reasoning} detail each of the three major groups combining ontologies and machine learning. Finally, section \ref{conclusion} presents the conclusions and a discussion of challenges and research directions.
\section{Methodology}
\label{S:2}

\begin{figure}[h]
\centering
\includegraphics[width=0.8\textwidth]{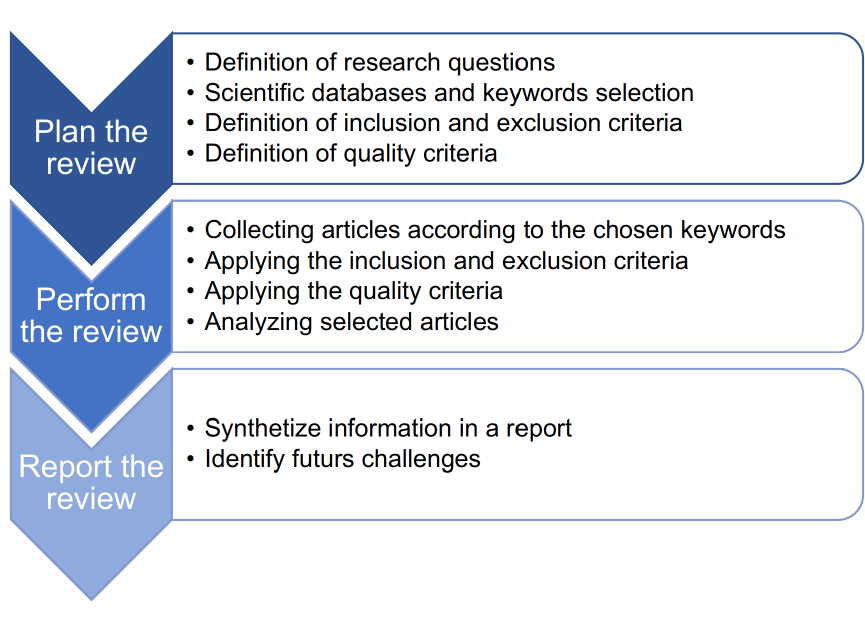}
\caption{Systematic Literature Review (SLR) methodology}\label{fig:methodology_SLR}
\end{figure}

The conducted SLR methodology is mainly inspired by~\cite{kitchenham_systematic_2010}, and is depicted in Figure~\ref{fig:methodology_SLR}. 
The first step is to plan the review after identifying the need to conduct it. We start by specifying the research questions. Then, we elaborate the adopted protocol to conduct the review by identifying keywords and selecting inclusion and exclusion criteria. 
The second step is performing the review by conducting the activities planned in the protocol and selecting primary studies to be included in the review, as well as the actions related to their evaluation. 
The final step is the reporting step, which consists of documenting, explaining, and summarizing the results to answer each research question.

\subsection{Planning the review}
This SLR aims to study current work combining ontology and machine learning. First, we define the research questions to be studied in this review and then detail the protocol used to select the primary studies.

\subsubsection{Definition of the research questions}
The objective of this review is to answer the following five research questions:

\begin{itemize}[itemindent=1em]
\item[\textbf{RQ1}:] To understand the motivation of current work and used methodology, we ask the following question: \textit{Why and how ontology and machine learning are combined?}
    \item[\textbf{RQ2}:] There are various machine learning models using different approaches and algorithms. To better understand which algorithms are used, we ask the following question: \textit{What are the machine learning algorithms used in each work?}
    \item[\textbf{RQ3}:] Ontologies are typically used in the reasoning process to ensure model consistency and to infer new knowledge. Most often, ontologies infer on subsumption links, but some works go further by adding rules that go beyond simple hierarchical relationships. To find out whether works use non-hierarchical reasoning rules, we ask the following question: \textit{Are there works that use rules other than subsumption links?}
    \item[\textbf{RQ4}:] Several main themes, defined by the ACM Computer Classification System\footnote{\url{https://dl.acm.org/ccs}}, are recurrent in the field of AI, such as computer vision or natural language processing (NLP). The treatment of each theme is particular and often involves techniques that are different from each other: \textit{What are the different types of main AI themes present in work combining machine learning and ontology?}
    \item[\textbf{RQ5}:] AI is used in different domains and plays an important role in helping humans to work with better performance. To identify the application domain addressed by current work, we ask the following question: \textit{What are covered application domains?}
\end{itemize}

\subsubsection{Scientific databases and keywords selection}
\label{scientific_databases_keywords_selection}
Once the research questions are defined, we identify the scientific databases and search keywords that will be used to select primary studies.
This SLR uses three scientific databases to cover a large panel of articles: Web Of Science\footnote{\url{https://clarivate.com/webofsciencegroup/solutions/web-of-science/}}, ACM Digital Library\footnote{\url{https://dl.acm.org/}} and Science Direct\footnote{\url{https://www.sciencedirect.com/}}.
These three search engines are recommended by~\cite{gusenbauer_which_2020}. They satisfy the reproducibility criteria of the searches, as well as the use of boolean terms in the query. 
Google Scholar is not used because it does not guarantee good reproducibility \citep{gusenbauer_which_2020}. 

To query selected scientific databases, a combination of the two main keywords, \textit{ontology} and \textit{machine learning} is used. However, the number of articles obtained was too large and all of these studies are not relevant to our review. Then, we refined the query by restricting the search of these same keywords in \textit{title, abstract, and keywords}, which gave a much smaller number of articles that seem more relevant.

After a first analysis, we find that the term \textit{deep learning} is often used instead of \textit{machine learning}, even though it is a subset of this technique. We, therefore, decided to add the \textit{deep learning} keyword to our query, with the same restrictions applied to the \textit{machine learning} term.
Also, some authors use directly the term \textit{neural network} (in particular for the most recent articles) without explicitly mentioning the \textit{machine learning} or \textit{deep learning} terms. We, therefore, decided to add this keyword to our query, even if it concerns a minority of articles (10 to 15\%) in each query.

For the \textit{ontology} keyword, we decided to use only this term and not combine it with other keywords representing different semantic techniques, such as \textit{taxonomy}, \textit{knowledge modeling}, or \textit{knowledge graph}. In this review, we are mainly interested in the use of ontologies, and the possibility to perform logical reasoning.

Finally, we have also included in our query the \textit{artificial intelligence} keyword. This allows us to restrict obtained results to our research domain. Unlike previous keywords, the presence of this term is looked for anywhere in the article to be less restrictive.

Based on our search and selected keywords, we obtain the following query: 
\begin{framed}
\begin{center}
\textit{``ontology" AND (``machine learning" OR \\``deep learning" OR ``neural network") AND ``artificial intelligence"}
\end{center}
\end{framed}

This query allows targeting primary studies concerned by our research questions.

\subsubsection{Definition of inclusion and exclusion criteria}
To filter the returned articles from the keyword search and keep relevant papers that will be used to answer our research questions, we defined a set of inclusion and exclusion criteria.

\paragraph{Inclusion criteria}
\begin{itemize}[itemindent=2em]
    \item[\textbf{InC1}:] The paper describes an approach that combines at least one ontology with at least one machine learning technique. 
    \item[\textbf{InC2}:] The paper does not use ontologies and machine learning only to compare them.
\end{itemize}

\paragraph{Exclusion criteria}
\begin{itemize}[itemindent=2em]
    \item[\textbf{ExC1}:] Posters or demonstrations that do not provide enough details about their contribution.
    \item[\textbf{ExC2}:] Duplicate papers returned from various search engines.
    \item[\textbf{ExC3}:] Papers that are not written in English.
    \item[\textbf{ExC4}:] Non-accessible papers that can not be online recovered. 
    \item[\textbf{ExC5}:] Books (or book chapters) detailing previously collected papers.
    \item[\textbf{ExC6}:] Extended paper by the same authors. In this case, the most recent paper is kept.
    \item[\textbf{ExC7}:] Existing survey or not a primary study (it may be a secondary or tertiary study).
\end{itemize}

\subsubsection{Definition of quality criteria}
The quality of an SLR depends on the quality of the reviewed articles. It is then important to rigorously assess the papers included in our SLR by considering the following quality criteria:
\begin{enumerate*}[label=(\alph*)]
  \item studies are conducted in higher research institutions
  \item studies are published in good quality international revues and conferences and referenced by well-known electronics libraries
  \item motivations and contributions are clearly defined.
\end{enumerate*}

To evaluate the quality of our SLR, we used the Quality Assessment Instrument for Software Engineering systematic literature Reviews (QAISER) developed by~\cite{usman_quality_2021}.

\subsection{Performing the review}
This section describes how we performed the review according to the defined protocol. We follow four steps:
\begin{enumerate*}[label=(\roman*)]
  \item collecting articles according to the chosen keywords,
  \item applying the inclusion and exclusion criteria,
  \item applying the quality criteria, and
  \item analyzing selected articles.
\end{enumerate*}

\subsubsection{Collecting articles}
In this step, we query the selected scientific databases with the set of defined keywords by adapting our basic query to each scientific database, as presented in table~\ref{tab:request}. 
The first search was carried out at the end of May 2021, and a total of 373 studies were collected to be analyzed. A second search was conducted in February 2022 with the objective of updating our analysis report with all new studies published since the first search. By limiting ourselves to studies published in 2021 and 2022 on the selected scientific databases, we were able to add 70 articles to our initial collection.

\begin{table}[ht]
 \renewcommand{\arraystretch}{1.5}
	\caption{Final request for each search engine}\label{tab:request}
	\begin{center}
	\begin{minipage}{\textwidth}
    	\begin{tabular}{|p{0.2\linewidth} | p{0.55\linewidth} | p{0.05\linewidth} | p{0.05\linewidth}|}
         \hline
    		 \textbf{Scientific database} & \textbf{Request} & \textbf{May 2021} & \textbf{Feb. 2022} \tabularnewline
    		\hline
    		ACM	& [Abstract: ontology] AND [[Abstract: ``machine learning"] OR [Abstract: ``deep learning''] OR [Abstract: ``neural network"]] AND [All: artificial intelligence] & 101 & 9 \tabularnewline
    		\hline
    		Science Direct & artificial intelligence AND Title, abstract, keywords: ``ontology" AND (``machine learning" OR ``deep learning" OR ``neural network") & 106 & 23 \tabularnewline
    		\hline
    		Web of Science & TS=(ontology AND (``machine learning" OR ``deep learning" OR ``neural network")) AND WC=``Artificial Intelligence" & 166 & 38 \tabularnewline
             \hline
    	\end{tabular}
    	\end{minipage}
	\end{center}
\end{table}

\subsubsection{Applying inclusion, exclusion, and quality criteria}
\begin{figure}[h]
\centering
\includegraphics[width=0.9\textwidth]{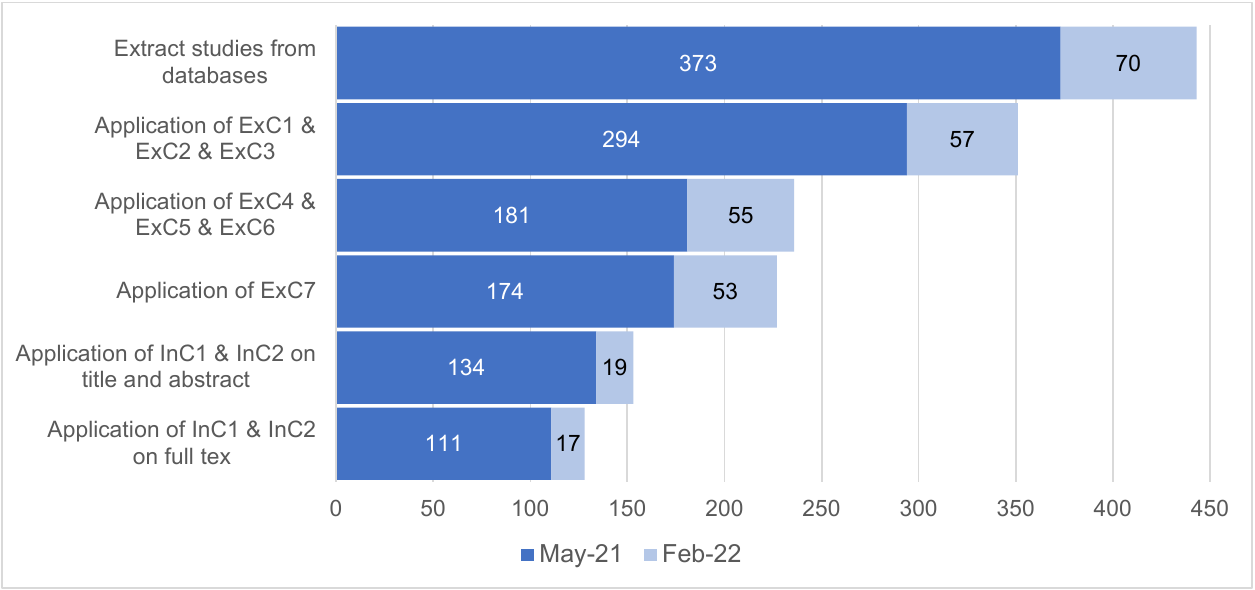}
\caption{Selection of articles}
\label{fig:funnel_articles}
\end{figure}

The different steps for applying the inclusion and exclusion criteria are summarized in Figure~\ref{fig:funnel_articles}. 
After collecting the set of 443 articles, we applied the first four exclusion criteria using Zotero\footnote{\url{https://www.zotero.org/}}, a reference management tool, to remove posters, demonstrations, duplicated and inaccessible papers. The remaining 351 articles are all written in English. In the second step, we read the titles and the abstracts of the obtained studies and apply the last three exclusion criteria to eliminate books, extended papers, and existing surveys. In this step, we also remove studies that did not meet both inclusion criteria based on title and abstract. We obtain 153 studies that we read to verify if they respect our two inclusion criteria. As a result, we selected 128 studies that correspond to the scope of our review.
We also applied the quality criteria defined previously to evaluate the selected primary studies. We assume that the quality of used scientific databases ensures the quality of selected studies, and the relevance of each included article was discussed and validated by all authors of this SLR.

\subsubsection{Analysis}
To answer our research questions, we extracted from each primary study the different attributes described in the table~\ref{tab:extract_data}. Based on the extracted data, we performed some statistical analysis that we present in section~\ref{S:3}.

\begin{table}[ht]
 \renewcommand{\arraystretch}{1.5}
	\caption{Extracted data from final studies}\label{tab:extract_data}
	\begin{center}
	\begin{minipage}{\textwidth}
	\begin{tabular}{| p{0.3\linewidth} | p{0.6\linewidth} |}
	    \hline
		 \textbf{Attribute} & \textbf{Description}\\ 
		 \hline
		 Year                       & Year of publication\\
		 \hline
		 Country                    & Countries where the first author is located\\
		 \hline
		 Machine learning algorithm & The machine learning algorithm(s) used in the paper. \\
		 \hline
		 Ontology reasoning         & Presence of deductive reasoning, at least of formal rules allowing deductive reasoning.\\
		 \hline
		 Artificial intelligence themes & If the studies explore a known theme of Artificial Intelligence as described by the ACM Computing Classification System\\ 
		 \hline
		 Category                  & Category of the article according to our classification of machine learning and ontologies combinations\\
		\hline
	\end{tabular}
	\end{minipage}
	\end{center}
\end{table}
\section{Overview of included studies combining ontologies and machine learning techniques}
\label{S:3}

\begin{figure}[ht]
\centering
\includegraphics[width=0.8\textwidth]{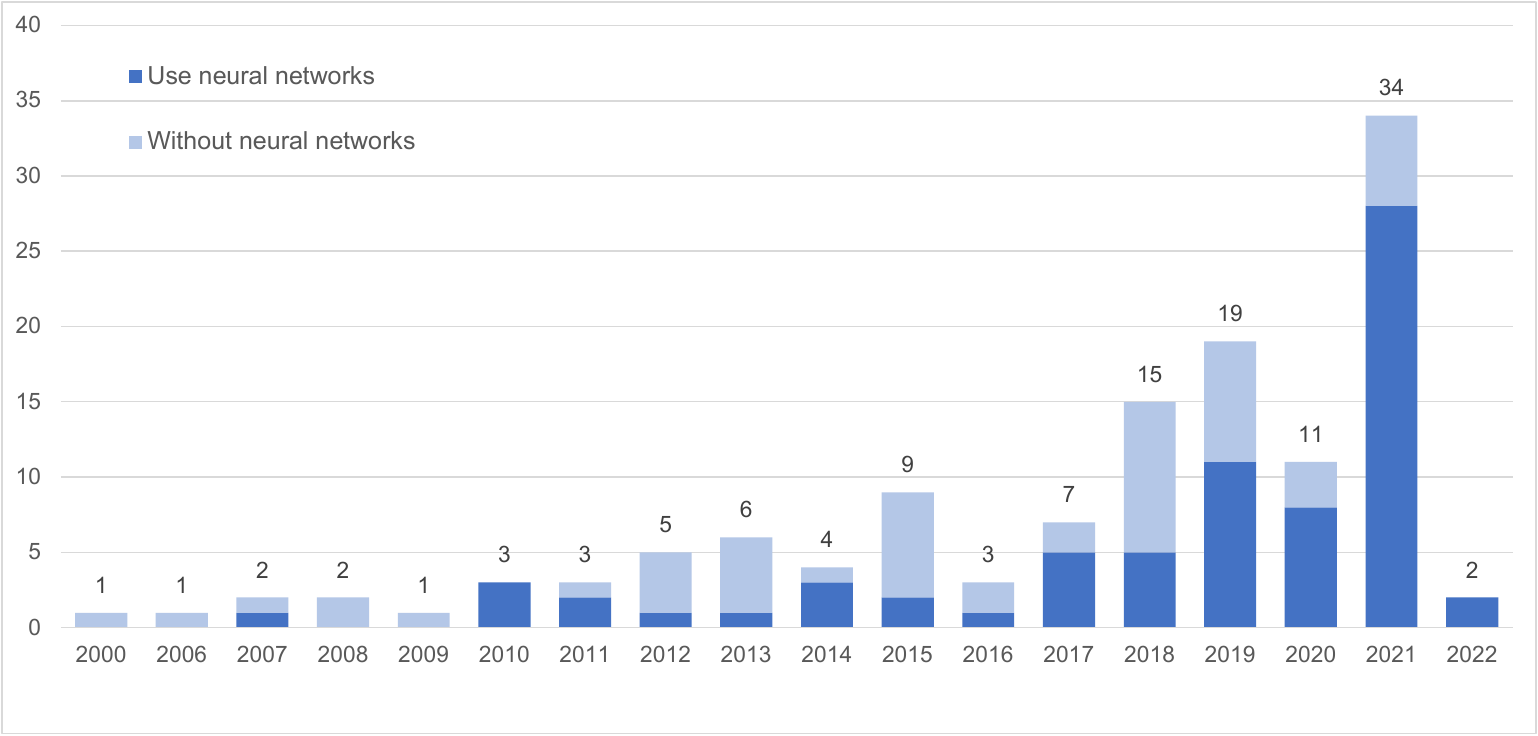}
\caption{Evolution of the number of publications about the combination of ontologies and machine learning.}\label{fig:evolution_nombre_papier}
\end{figure}

In addition to the main analysis for answering the research questions, we performed a demographic analysis of the studies. Figure~\ref{fig:evolution_nombre_papier} presents the number of published papers concerning the combination of ontologies and machine learning techniques. The first paper present in this SLR dates from the year 2000, voluntarily we did not put any restriction on the publication date in our query (cf \ref{scientific_databases_keywords_selection}). Consequently, this SLR allows us to make a state of the art on the combination between ontologies and machine learning for more than 20 years.
Since 2010 the number of studies has increased, and a strong acceleration is taking place in recent years. Indeed, 57\% of the analyzed studies were published after 2018.
Over the years, we see that neural networks are being used more and more. The neural networks group includes algorithms range from the simple perceptron to the most recent techniques such as Transformers. As we see in more detail in the section \ref{ml} neural networks are present in the majority of the papers studied.

\begin{figure}[ht]
\centering
\includegraphics[width=0.7\textwidth]{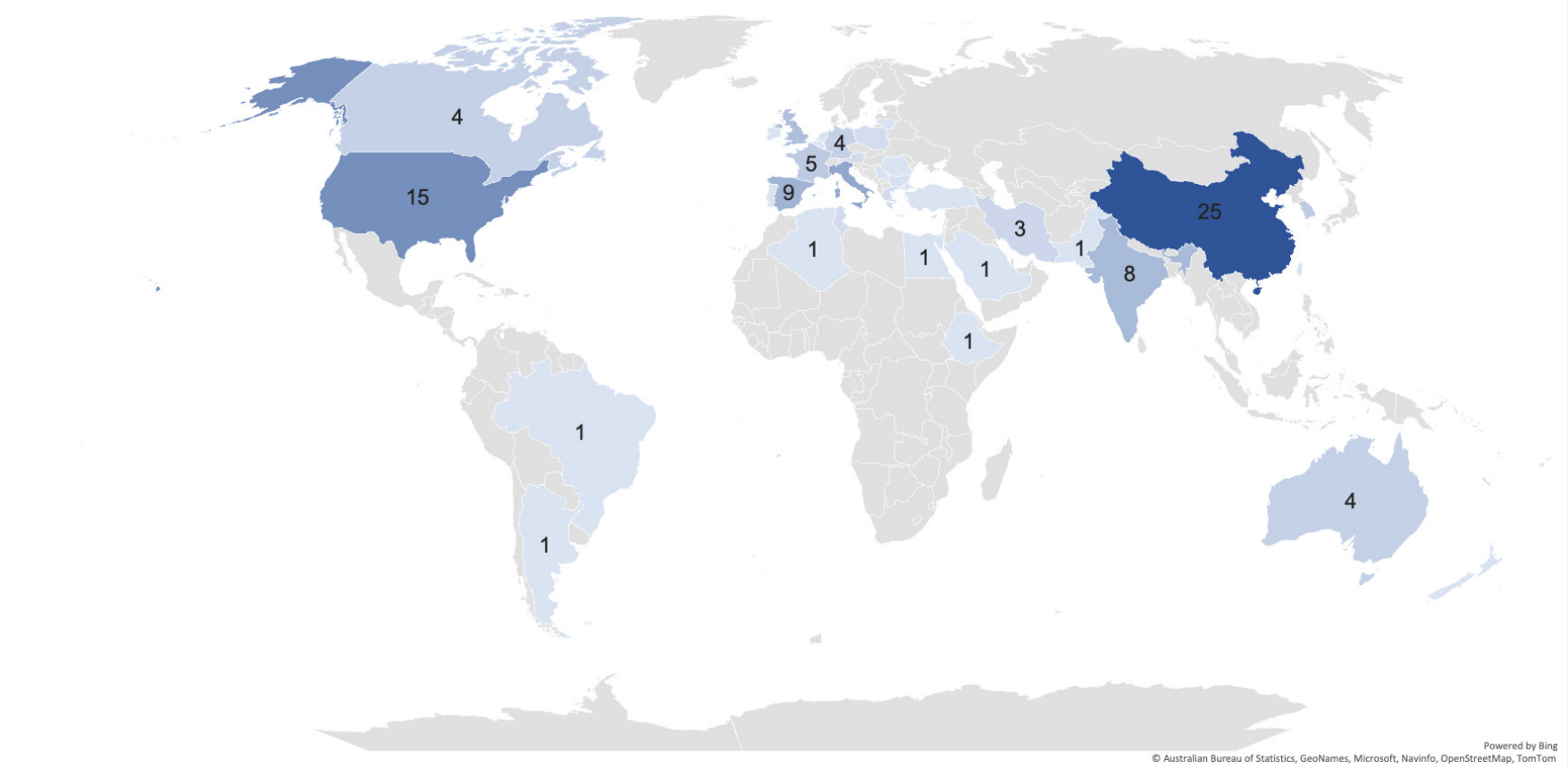}
\caption{Country where the first author is based}\label{fig:countries}
\end{figure}

Figure~\ref{fig:countries} present the geographical distribution of contributors, considering the location of the first author of each article.  The most represented continents are Europe, Asia, and North America. In Europe, leaders are Italy and Spain, each with a dozen papers in the study, but it is above all the multiplicity of contributing countries (United Kingdom, France, Germany, Austria, Poland, Greece, Bulgaria, Romania, Belgium, Lithuania, or Serbia) that allow Europe to come out on top of the most contributing continents. In Asia, it is mostly China that contributes to the second position of our ranking. In North America, the USA provides a large part of the studies.

These data are interesting if we cross them with those presenting the world public budgets of research and development (R\&D)\footnote{http://uis.unesco.org/apps/visualisations/research-and-development-spending/}. The presence of China and the USA at the top of our ranking correlates with the budget invested in R\&D each year. The large contributions of Italy and Spain, on the other hand, are more difficult to explain in terms of their R\&D spending. However, these two countries are present in the Investment Monitor top 40 countries ranking\footnote{\url{https://www.investmentmonitor.ai/ai/ai-index-us-china-artificial-intelligence}}. In Italy, we note that several studies come from Trento University, which also has an interest in more specific neuro-symbolic field.
This demographic study allows us to highlight some university teams that are actively engaged in conducting research in the combination of ontologies and machine learning.

\subsection{Types of combination between ontology and machine learning (RQ1)}
\begin{figure}[ht]
\centering
\includegraphics[width=0.9\textwidth]{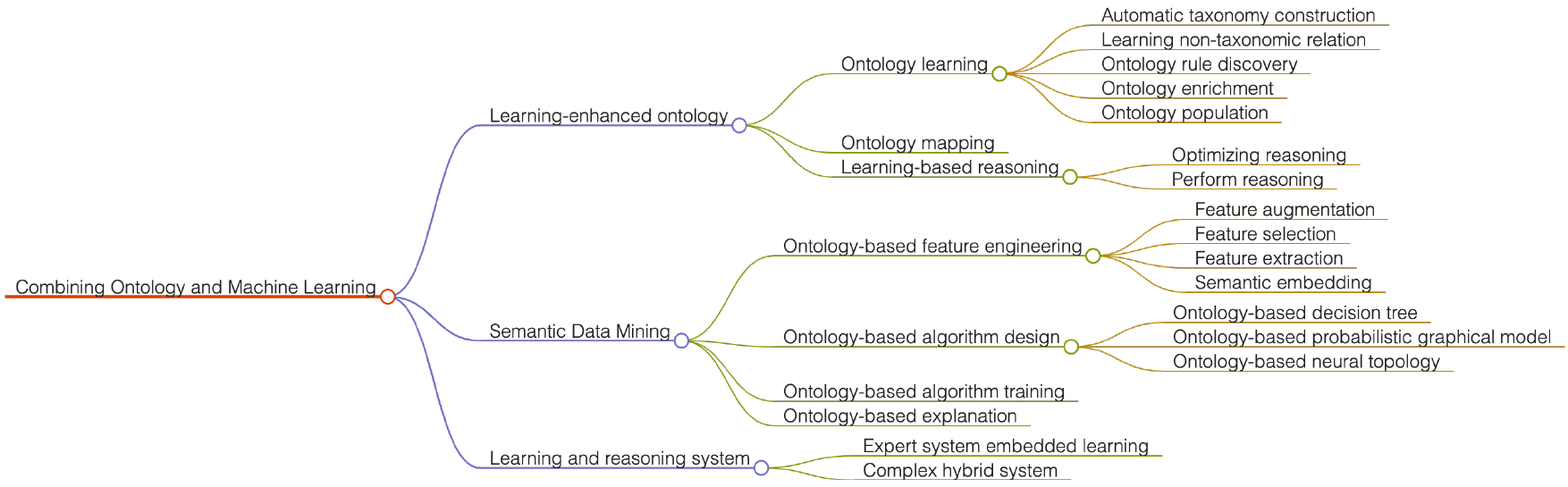}
\caption{Overview of the combination of ontologies and machine learning techniques}\label{fig:categories_combining_OWL_ML}
\end{figure}

After reading the 128 selected articles, we could distinguish three groups of different ontology and machine learning combinations: \emph{Learning-Enhanced Ontology}, \emph{Semantic Data Mining}, and \emph{Learning and reasoning system}. These three main groups and their subgroups have been partially named thanks to recent work that focuses on different forms of combination between inductive learning and deductive reasoning~\citep{von_rueden_informed_2021}. This explains why we sometimes find the term \emph{semantic} instead of \emph{ontology}, whereas in this SLR we only deal with papers that use an ontology for the symbolic part. This will make it easier for the reader to make the connection with other articles dealing with learning and reasoning architecture (such as neuro-symbolic). We detail these three main groups and their subgroups, presented in Figure~\ref{fig:categories_combining_OWL_ML}, in the paragraphs \ref{knowledge_aquisition_discovery}, \ref{semantic_data_mining} and \ref{combining_learning_reasoning} in order to be able to answer research questions RQ1, RQ2, and RQ3 in detail.  

\subsection{Machine learning techniques (RQ2)}
\label{ml}

To answer RQ2, we identified the different machine learning algorithms used in each of the selected articles. 
All the machine learning algorithms used are listed in the "learning algorithm" column of Tables~\ref{tab:learning-enhanced_ontology}, \ref{tab:semantic_DM_algo} and \ref{tab:reason_after_learning}, while the "learning type" column indicates whether they are supervised or otherwise.

Selected articles are divided into the three main learning approaches as follows:
\begin{itemize}
    \item Supervised learning: 107
    \item Unsupervised learning: 37 of which 13 are self-supervised learning
    \item Reinforcement learning: 1
\end{itemize}

As presented in Figure~\ref{fig:evolution_nombre_papier}, neural networks are very common in the selected studies. Neural networks are well involved in the supervised and self-supervised categories and less present in the unsupervised category since their application to clustering problems is more recent. The authors often prefer other more classical clustering algorithms such as k-means, hierarchical ascending classification, principal component analysis, or latent Dirichlet allocation (LDA).

\subsection{Ontology not only subsumption reasoning (RQ3)}

\begin{figure}[ht]
\centering
\includegraphics[width=0.6\textwidth]{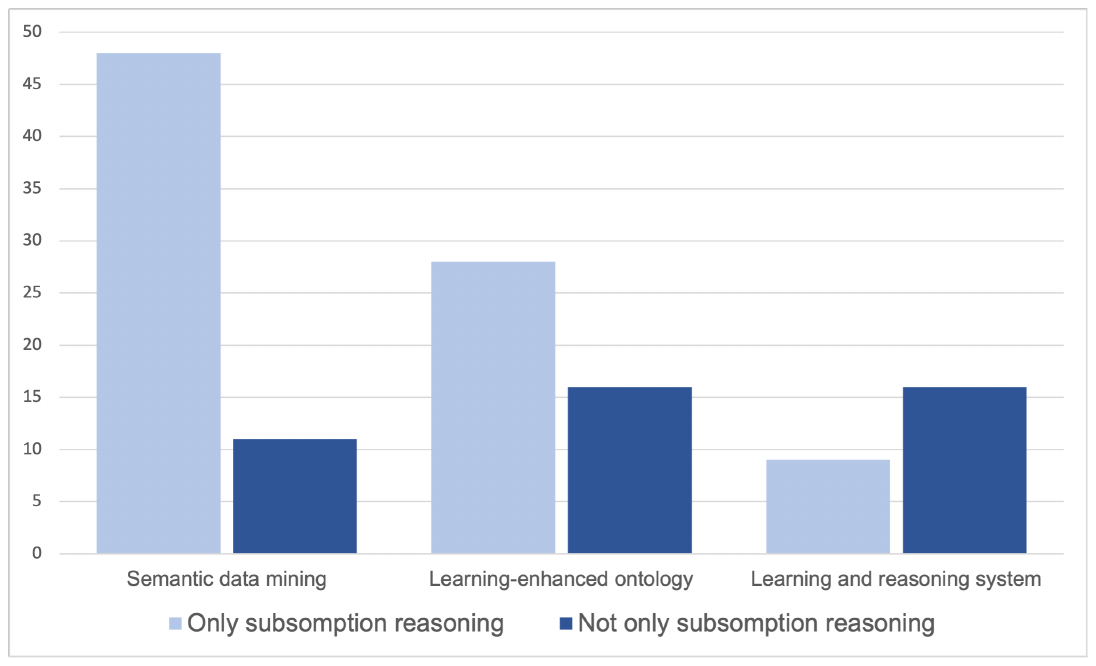}
\caption{Proportion of subsumption and not only subsumption ontology reasoning in studies}\label{fig:ontology_reasoning}
\end{figure}

It is interesting to note that a majority of articles (see Figure~\ref{fig:ontology_reasoning}) only use subsumption rules for deductive reasoning. By deductive reasoning, we mean here ontological reasoning, i.e. the deduction of new facts from general rules. In this review, most of the papers use only the semantic contribution, in particular hierarchical relations, of ontologies and do not focus on the discovery of new facts based on ontological reasoning. Thus, only 37\% of the reviewed articles describe a form of combination between inductive learning and deductive reasoning with non-hierarchical rules. These articles are identified by a check mark in the "Reasoning" column of the tables \ref{tab:learning-enhanced_ontology}, \ref{tab:semantic_DM_algo} and \ref{tab:reason_after_learning}.

It appears that many authors use ontologies as improved taxonomies (with non-heuristic relations between concepts) but do not use more complex rules for the inference part.  

\subsection{Artificial intelligence themes (RQ4)}
ACM Computing Classification System\footnote{\url{https://dl.acm.org/ccs}} defines several main themes involved in artificial intelligence like: \emph{Natural Language Processing (NLP)}, \emph{Computer vision}, \emph{Multi-agents system}, \emph{Time series} and \emph{Planning and scheduling}. Some of the selected articles are dedicated to an application in one of these themes. 

\begin{figure}[ht]
\centering
\includegraphics[width=0.7\textwidth]{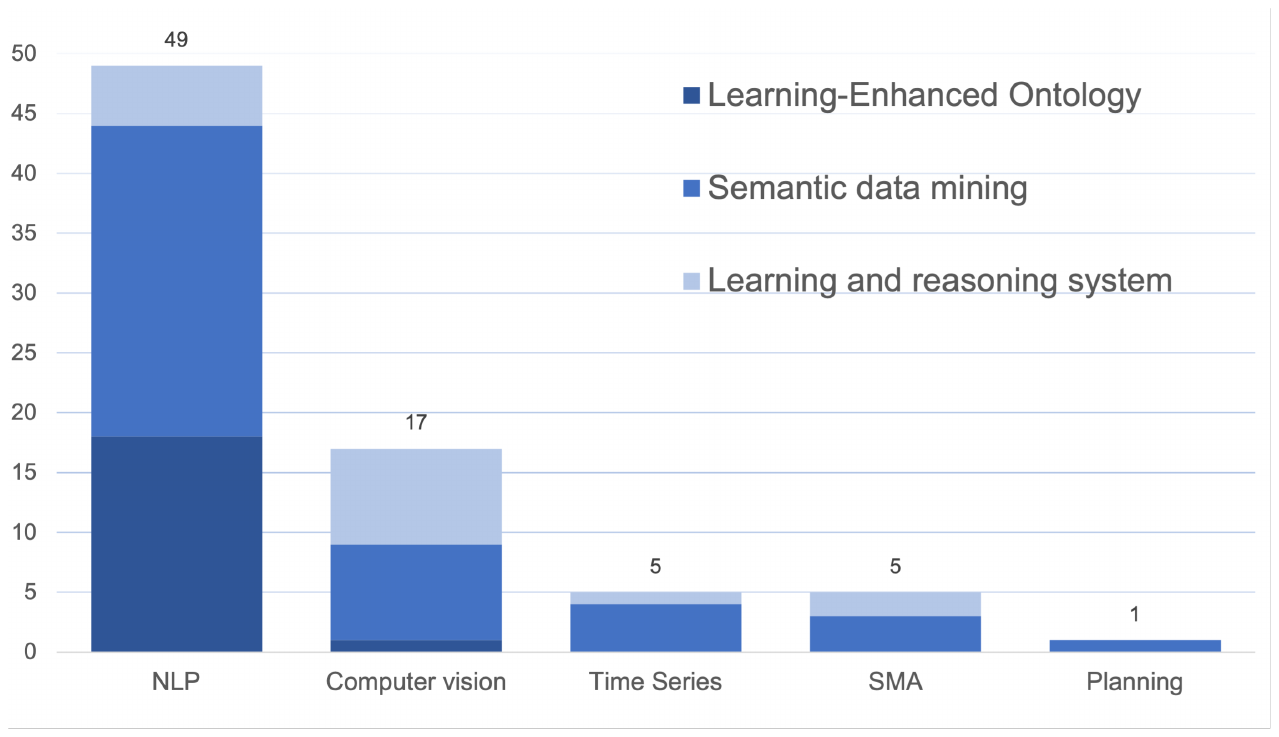}
\caption{Artificial intelligence themes}
\label{fig:AI_themes}
\end{figure}

The most present approach is \emph{NLP}, including articles dealing with \emph{ontology learning} and \emph{informed machine learning} (cf. Figure~\ref{fig:AI_themes}) as detailed in sections \ref{ontology_learning} and \ref{semantic_data_mining}. Some articles deal with  \emph{Computer vision}, in particular for image recognition. Few articles are involved in \emph{Multi-agents system} and \emph{Time series}. Only one article is concerned by \emph{Planning and scheduling}. All relevant AI themes, not only the five main themes, are listed in the "AI Theme" column of Tables \ref{tab:learning-enhanced_ontology}, \ref{tab:semantic_DM_algo} and \ref{tab:reason_after_learning}.

\subsection{Application domains (RQ5)}
\begin{figure}[ht]
\centering
\includegraphics[width=0.6\textwidth]{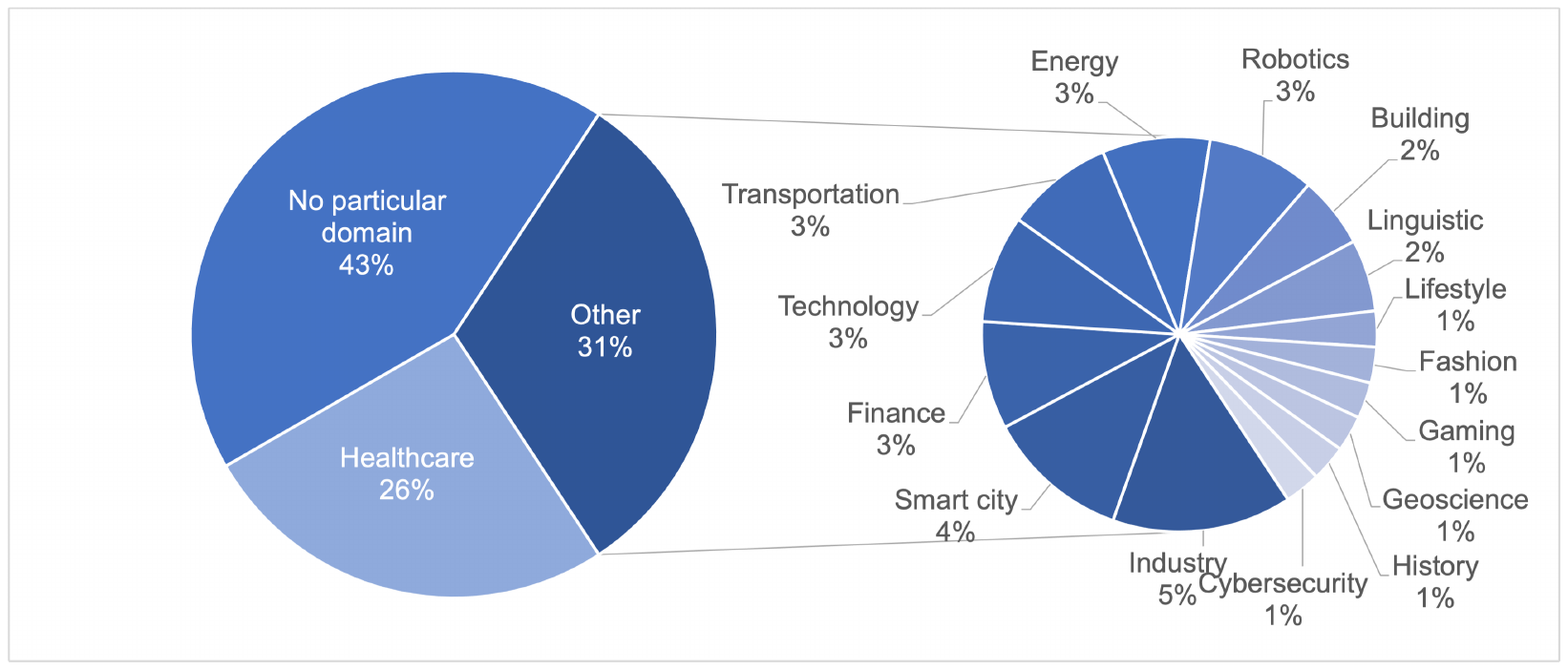}
\caption{Application domains}
\label{fig:application_domains}
\end{figure}

Figure~\ref{fig:application_domains} shows that 43\% of the papers do not focus on a single application domain. Indeed, the authors have chosen to solve a particular problem by basing their work on either generalist or interchangeable datasets in order to allow the reuse of their work in various application domains. 
However, the application domain the most encountered is Health (26\% of the studies), notably because medical ontologies such as SNOMED\footnote{https://bioportal.bioontology.org/ontologies/SNOMEDCT} or biological ontologies such as GeneOnto\footnote{http://geneontology.org/docs/download-ontology/} are often used in this domain. It is also a domain that has very strong constraints in terms of the explainability of results. The use of semantic data and ontological reasoning are quite appropriate for this kind of problematic \citep{rubin_biomedical_2008}.

The other domains present in this review are much more anecdotal, as shown in Figure~\ref{fig:application_domains}. The other domains own less than 10 papers, or even just one. All application domains are listed in the "Application domain" column of Tables \ref{tab:learning-enhanced_ontology}, \ref{tab:semantic_DM_algo} and \ref{tab:reason_after_learning}.

\section{Learning-Enhanced Ontology}
\label{knowledge_aquisition_discovery}
There are several ways to improve the use of ontologies through machine learning. First, ontology creation and maintenance can be (partly) automated thanks to machine learning techniques. In this case, we speak about \emph{ontology learning} \citep{wong_ontology_2012}, in which ontologies can be learned from various resources. Second, \emph{ontology mapping} groups together the categories that allow the use of ontologies to be improved thanks to machine learning (i.e., guaranteeing interoperability). Finally, \emph{learning-based reasoning} presents the set of techniques to facilitate deductive ontological reasoning thanks to machine learning.

The comprehensive details of all papers within this category are outlined in Table \ref{tab:learning-enhanced_ontology}, where they are meticulously categorized by their respective field of application, AI themes they explore, and the machine learning algorithms employed.

\subsection{Ontology learning}
\label{ontology_learning}
Ontology learning is the process through which ontologies are automatically generated or enriched from various sources of data and knowledge~\citep{maedche_ontology_2001}. This concept has already been studied in many recent reviews due to its potential to provide valuable assistance in the creation of ontologies, a traditionally time- and resource-intensive task~\citep{al-aswadi_automatic_2020, khadir_ontology_2021, asim_survey_2018}.

\begin{figure}[ht]
\centering
\includegraphics[width=0.2\textwidth]{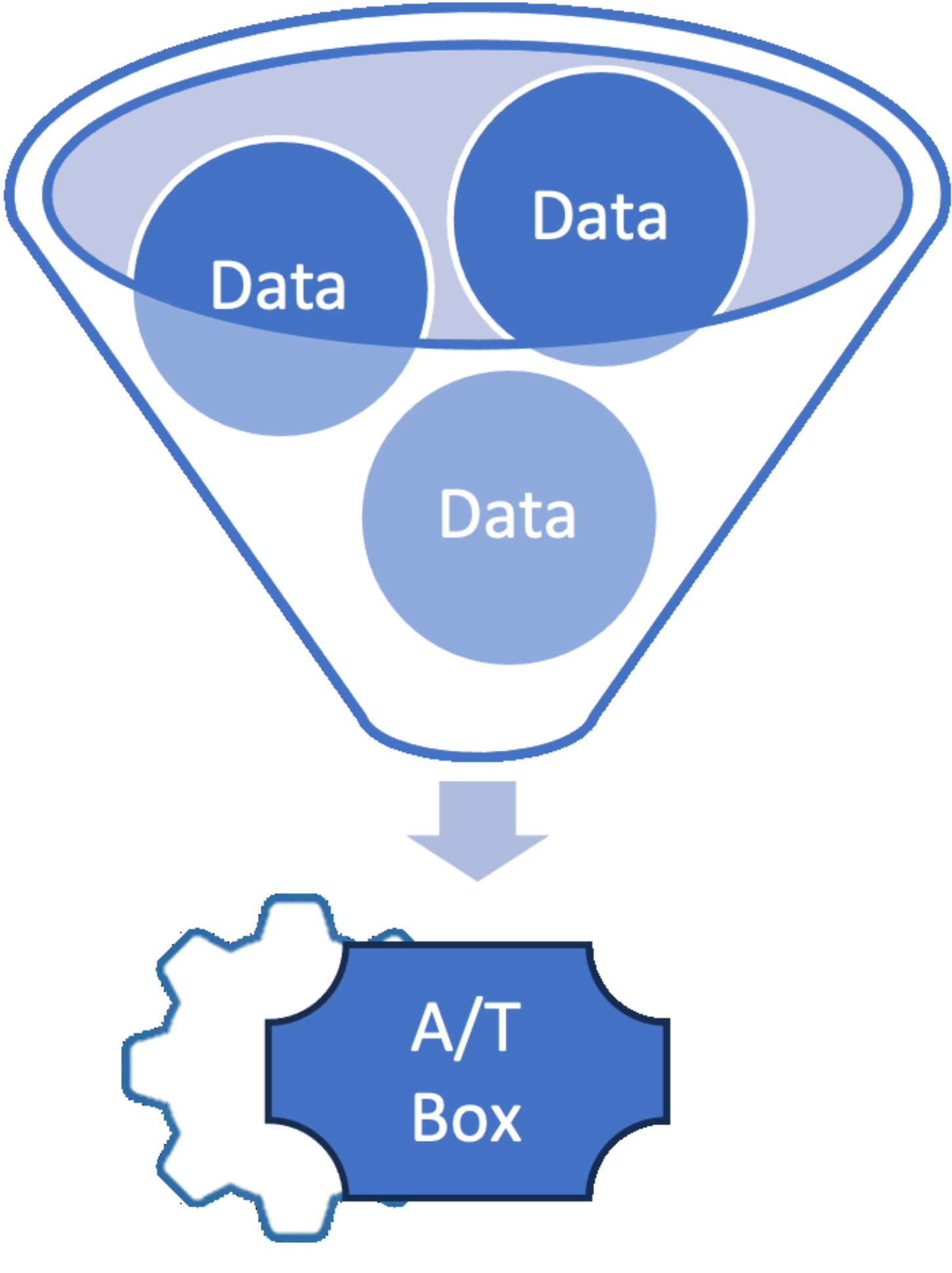}
\caption{Ontology learning mechanism}
\label{fig:11}
\end{figure}

The ontology learning process involves collecting and sometimes analyzing data from diverse sources such as texts, databases, web documents, and even existing ontologies. Using machine learning algorithms, information extracted from these data is then utilized in identifying concepts, relationships, and properties that could potentially be integrated into an ontology. Figure~\ref{fig:11} illustrates this mechanism, showing that data is processed by a machine learning algorithm, symbolized by a funnel, before being transformed into elements of the TBox or ABox of an ontology, represented by the annotated polygon "A/T Box" associated with a cogwheel. The cogwheel symbolizes the final ontology created, upon which inferences can now be made.

\subsubsection{Automatic taxonomy construction}

Automatic Taxonomy Construction (ATC) is the computer process of systematically generating a hierarchical classification system for entities or concepts based on inherent relationships, attributes, or similarities. 

The ATC process comprises several distinct stages, each of which is identified and carefully described by~\cite{getahun_integrated_2017}. Firstly, the pre-processing stage involves preparing the raw data, usually text, by cleaning, tokenizing, and normalizing it to facilitate subsequent analysis. Next, the concept extraction phase focuses on identifying key concepts or terms in the pre-processed data, using techniques such as NLP and pattern recognition. Concept extraction can employ various techniques, ranging from conventional approaches like TF-IDF~\citep{ghoniem_novel_2019} to topic modeling methods such as LDA~\citep{rani_semi-automatic_2017}, or advanced deep learning techniques using Word2vec, including CBOW or Skip-G~\citep{albukhitan_arabic_2017}.
These extracted concepts are then mapped to specific domains or topics in the concept-domain matching phase, where they are ranked according to their relevance, often using a similarity score~\cite{albukhitan_arabic_2017, getahun_integrated_2017, ghoniem_novel_2019}. Next, the concept-pair extraction phase aims to discover relationships between concept pairs, often using semantic analysis and graph-based algorithms to identify associations. Finally, the taxonomic relationship extraction phase aims to establish taxonomic relationships between identified concepts, such as hierarchical relationships like "is-a" or "part-of", thus completing the taxonomy construction. 
The concluding phase, serving as the core of ATC, encompasses the utilization of machine learning techniques such as formal concept analysis (FCA)~\citep{jurkevicius_ontology_2010}, hierarchical agglomerative clustering (HAC)~\cite{getahun_integrated_2017}, and, more recently, recurrent neural networks\citep{petrucci_expressive_2018} . Additionally, methods involving Markov networks, such as Markov Logic Networks (MLN)~\citep{wu_automatically_2008} or Conditional Random Field (CRF)~ \citep{song_automated_2016, jia_practical_2018}, are also employed in this crucial stage.
Together, these integrated steps form a comprehensive framework for the automatic generation of taxonomies from unstructured data. Studies frequently leverage a combination of at least three of the four mentioned steps to achieve this process. 

While numerous papers employ unstructured data, such as text corpora, as the input for ATC, it's important to recognize that ATC can also be applied to structured data, as demonstrated by the research conducted by~\citep{jia_practical_2018}.

\subsubsection{Learning non-hierarchical relations} 

Learning non-hierarchical relations in ontology learning involves identifying and comprehending semantic connections between entities that do not follow a hierarchical structure, typically focusing on association, correlation, or analogy-based methods to discover and label these relations. Incorporating these non-taxonomic relations elevates the achieved taxonomies to the status of ontologies.
To accomplish this,~\cite{albukhitan_arabic_2017} employs an undisclosed clustering algorithm, while~\cite{getahun_integrated_2017} utilizes both a correlation-based method and a concept analogy-based approach leveraging Word2Vec. In contrast,~\cite{petrucci_expressive_2018} utilizes a Seq2Seq algorithm to convert a natural language sentence like \textit{"A bee is an insect that produces honey"} into its formalized logical description, expressed as "$bee \subseteq insect \cap \exists produces.honey$".

\subsubsection{Rule discovery} 

Rule discovery or rule mining aims to discover actionable and interpretable rules that capture interesting patterns or dependencies within the data, aiding in decision-making and knowledge extraction. It facilitates the representation of complex relationships and inferences within ontologies (such as IF-THEN rules), often expressed in a formalized set of machine-readable rules, such as SWRL.

The study by~\cite{jurkevicius_ontology_2010} mentions the use of an artificial neural network (ANN) for rule generation, although details on this aspect are limited in the article.~ \cite{mcglinn_usability_2017} adopt a mix of intelligent rule generation techniques, employing both ANNs and genetic algorithms (GA), as well as data mining rules using decision tree techniques on historical sensor data.~\cite{ko_machine_2021} focus on the construction of design rules for additive manufacturing (AM) using the machine learning algorithm CART (Classification and Regression Tree) on measurement data. In their study in the financial domain,~\cite{yang_construction_2020} use the Apriori algorithm to discover association rules between data items in transactions. This algorithm is particularly suitable for inferring situational elements of risky events such as time and place through the analysis of transaction data.

These first three areas: automatic taxonomy construction, learning non-hierarchical relations, and rule discovery are TBox statements.

\subsubsection{Ontology population}

Ontology population is the process of enriching the TBox within an ontology by adding a substantial base of factual knowledge or instances. This involves inserting concrete data or instances into the ontology's conceptual framework, thus constituting the ABox part of the ontology. The automatic populating of an ontology is an important issue, as it enables the deductive reasoning mechanism to be used rapidly on a sometimes heterogeneous database.

In the study by~\cite{craven_learning_2000}, the focus is on populating the ontology by extracting new instances from web pages using the Naive Bayes algorithm. The algorithm is used to classify and identify instances, providing a method for populating the ontology with factual information gathered from web sources.~\cite{kordjamshidi_global_2015} focus on populating the ontology with spatial information extraction. They use the Support Vector Machine (SVM) algorithm to efficiently populate the ontology with spatial information by extracting relevant details from different sources, helping to enrich the knowledge base.~\cite{markievicz_action_2015} extend the application of SVMs to the field of robotics. The study focuses on the classification of actions described in a corpus of texts relating to chemistry experiments. The ultimate goal is to translate these actions into a robot executable format. In the healthcare field,~\cite{rubrichi_system_2013} propose a methodology for the automatic recognition of drug-related entities in textual descriptions of drugs. They use the CRF (Conditional Random Field) algorithm, derived from Markov methods, to populate the ontology with this drug-related information.~\cite{packer_cost-effective_2015} use hidden Markov models (HMMs) for the ontology population in the historical domain. They present ListReader, a method for training the structure and parameters of an HMM without the need for labeled training data. This approach is particularly beneficial for dealing with OCR errors in historical documents.~\cite{kuang_integrating_2018} address large-scale visual recognition in the context of computer vision. They propose a multi-level deep learning algorithm that combines deep convolutional neural networks (CNNs) and tree classifiers. In the field of biochemistry,~\cite{ayadi_ontology_2019} introduces a new approach to automatically populate the ontology of biomolecular networks. They rely on artificial neural networks (ANNs), in particular deep learning, and preprocessing techniques with Word2Vec.

\subsubsection{Ontology enrichment}

While the terminology section of the ontology, captured by the TBox, tends to be less dynamic than the instances in the ABox, regular maintenance is essential to prevent the ontology from becoming outdated. Ontology enrichment refers to the process of enhancing an ontology by updating its content through the addition or modification of concepts, properties, and relationships~\citep{messaoud_semcado_2015}. This process aims to expand the ontology's knowledge representation to accommodate new information and ensure its relevance to evolving domains or applications. Indeed, it is unrealistic to anticipate the inclusion of all domain and expert knowledge in an initial ontology due to various factors. These may include experts' limitations in formalizing their knowledge comprehensively from the beginning or the possibility that certain problems or required knowledge have not yet been identified~\citep{thomopoulos_iterative_2013}.

In the biomedical study by~\cite{valarakos_building_2006}, a dataset from a domain-specific corpus (PubMed abstracts) is used. Hidden Markov Models (HMM) are used to extract relevant tokens from the dataset. Next, the COmpression-based CLUstering (COCLU) algorithm is applied for ontology enrichment, focusing on non-taxonomic lexical-semantic relations.~\cite{thomopoulos_iterative_2013} focus on ontology enrichment in the context of the food industry. They use classification algorithms such as CART and C4.5 to extract new knowledge, including concepts and relationships, from a food dataset. Note that all new propositions are validated by a domain expert before being incorporated into the ontology.~\cite{messaoud_semcado_2015}  contribute to the enrichment of a medical ontology through causal discovery. Their method, SemCaDo (Semantic Causal Discovery), uses causal Bayesian networks (CBNs) to learn causal discoveries from gene expression datasets and gene ontology. The new knowledge found by the CBN is then used to evolve the ontology.~\cite{song_automated_2016} apply conditional random fields (CRFs) to discover Q\&A from collaborative engineering tasks. The discovered Q\&A are transformed into ontological concepts and relations by a semantic mapping step.~\citep{mihindukulasooriya_rdf_2018} focus on knowledge base (KB) quality assessment. They use the Random Forest (RF) algorithm to add integrity constraints to the KB, thereby improving its quality.
The work of~\cite{hong_constructing_2021} in the field of brain areas and autism uses natural language processing (NLP) techniques. BiLSTM and CRF are used for entity extraction, then BiLSTM is used again for relation extraction. Finally, instances with high confidence scores are manually reviewed by experts.~\cite{merono-penuela_multi-domain_2021}  target ontology evolution and concept drift detection in Web vocabularies. They exploit several algorithms provided by the WEKA API and use strings of RDF vocabulary versions as datasets.~\cite{djellali_using_2013}  propose a semi-automatic approach using truncated singular value decomposition (TSVD) and Fuzzy ART clustering for ontology enrichment. The method involves variable selection and clustering to identify candidate changes, reducing noise and improving clustering accuracy.

\subsection{Ontology mapping}
\label{ontology_mapping}

Ontology mapping, also known as ontology alignment, aims to discover correspondences between terms with similar meanings in two distinct ontologies while ensuring the overall structure coherence of the ontology~\cite{kalfoglou_ontology_2003}. The primary goal of ontology mapping is to establish a semantic correspondence between elements of ontologies to facilitate interoperability and extend their terminological scope by aligning concepts, properties, and instances.

\begin{figure}[ht]
\centering
\includegraphics[width=0.2\textwidth]{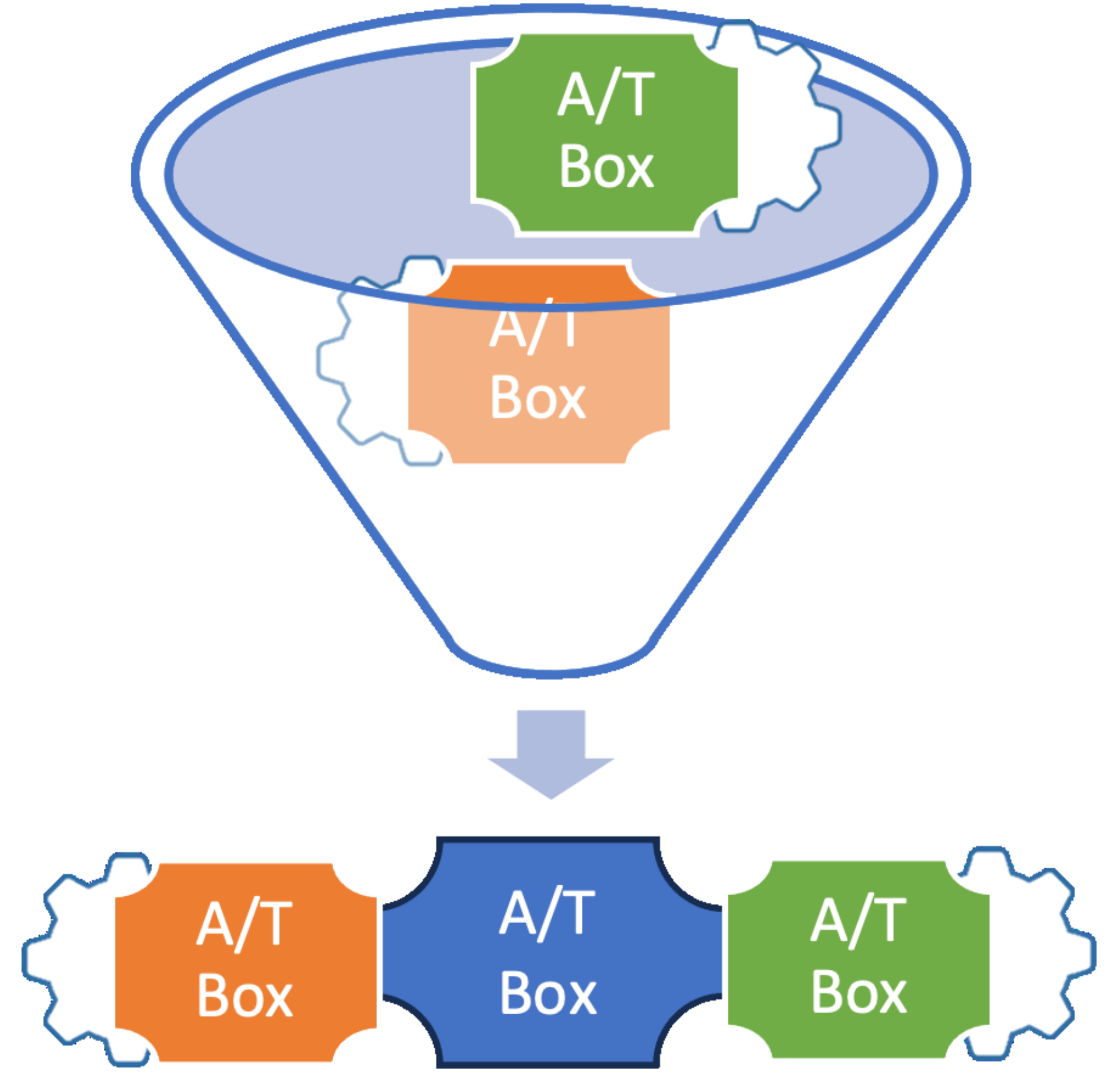}
\caption{Ontology mapping mechanism}
\label{fig:12}
\end{figure}

Concept alignment involves finding equivalences between similar concepts, such as "housing" and "dwelling"; property alignment matches relationships between concepts, such as "has owner" and "owned by"; and instance alignment associates specific individuals from different ontologies representing the same reality.
This process, depicted in Figure~\ref{fig:12}, begins with the preparation of source ontologies, represented by annotated polygons "A/T Box," including selecting relevant features in the TBox and ABox. Then, a machine learning model, symbolized again by a funnel shape, is trained, either supervised (with known correspondences) or unsupervised (automatically discovering potential correspondences). This model is then used to predict correspondences between common elements in both ontologies, as illustrated by the final "A/T Box" polygon situated between the two starting ontologies.
These correspondences are often subjected to quality evaluation, particularly for coherence, and integrated to enhance interoperability between source ontologies. Finally, post-processing may be applied to refine the results.

The use of machine learning in ontology mapping automates the process of discovering semantic correspondences, particularly valuable in environments where heterogeneous ontologies need integration.
To achieve this goal, ensemble methods are sometimes employed, such as Random Forest (using bagging)~\citep{rico_predicting_2018, annane_building_2018} or an approach involving three classifiers~\citep{fanizzi_composite_2011}.
However, neural networks, particularly the multi-layer perceptron (MLP)~\citep{rubiolo_knowledge_2012, shannon_comparative_2021}, and more recently, recurrent neural networks (RNNs) like LSTM~\citep{chakraborty_ontoconnect_2021}, or transformer models such as BERT~\citep{mohan_low_2021}, or specific design architectures like IAC~\citep{mao_adaptive_2010}, are the prevailing choices. Neural networks also contribute to data pre-processing through techniques like Word2Vec~\citep{zhou_semantic_2021}.

Ontology mapping is an essential process for harmonizing distinct ontologies and promoting data and knowledge interoperability. The use of machine learning techniques, including neural networks and ensemble methods, facilitates the ontology alignment process, making it applicable to large datasets. Ultimately, this contributes to fostering the efficient exchange of information in a dynamically evolving digital environment.

\subsection{Learning-based reasoning}

In the field of learning-based reasoning, the papers focus on the integration of machine learning techniques to improve ontological reasoning. Indeed, the main challenge facing ontological reasoning is its slow execution, particularly when deployed in real-life scenarios. In our fast-moving society, especially for real-time systems, machine learning algorithms offer a promising solution to enhance the efficiency of inference engines. They can also provide significantly faster alternatives to traditional inference engines such as Pellet~\citep{sirin_pellet_2007} or HermiT~\citep{shearer_hermit_2008}.

\begin{figure}[ht]
\centering
\includegraphics[width=0.2\textwidth]{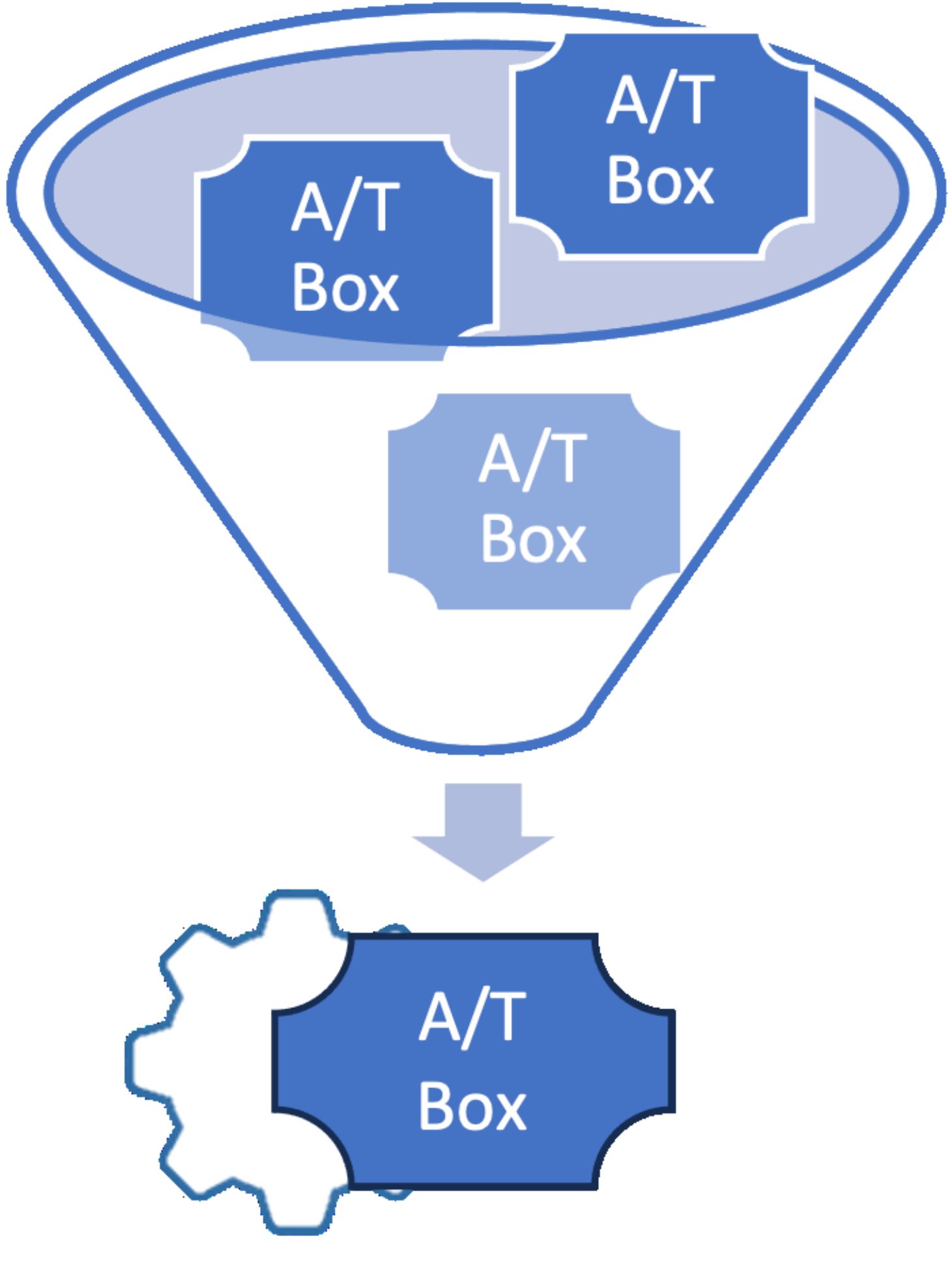}
\caption{Learning-based reasoning mechanism}
\label{fig:13}
\end{figure}

Diverse strategies for harnessing machine learning in ontological reasoning depend on the capability of machine learning algorithms to recognize intricate patterns, associations, and relationships within input ontologies. Figure~\ref{fig:13} illustrates this process, showing how machine learning, represented by a funnel, acquires the ability to perform deductive reasoning based on ontologies represented here by the annotated "A/T Box" polygon. The final deductive reasoning capability is symbolized by the cogwheel next to the "A/T Box" polygon.

\subsubsection{Optimizing reasoning}

Optimizing reasoning can be achieved through various methods, including selecting a deductive reasoner appropriate for the application context, predicting reasoner performance to detect scalability issues, or enhancing the performance of deductive reasoners themselves by incorporating machine learning.

Selecting a reasoner involves choosing an appropriate inference engine or reasoning tool to perform deductive reasoning. Capabilities, efficiency, and compatibility with specific languages or ontology formats may vary from one reasoner to another. The choice of reasoner depends on factors such as ontology complexity, the desired level of reasoning support, and available computing resources.
In the study by~\cite{bock_automatic_2012}, the focus is on selecting an appropriate reasoner for ontological reasoning tasks. The authors, recognizing that no reasoning algorithm universally excels in all description logic and reasoning tasks, implemented an approach within the framework of a reasoning broker called HERAKLES. In HERAKLES, machine learning techniques such as Naive Bayes, k-NN, Support Vector Machine (SVM), and Decision Tree are compared with each other. Through their experimentation, they found that the Decision Tree algorithm outperformed the others, demonstrating superior performance in choosing an appropriate inference engine.

Predicting reasoner performance involves estimating the duration required for a given reasoning task within a specified ontology.  Essentially, it involves forecasting the time needed for a reasoner to complete its tasks, facilitating better planning and management of ontology-related projects and applications.~\cite{pan_predicting_2018} use a combination of the Random Forest (RF) classifier and the Boruta algorithm for feature selection to predict reasoner performance. The challenge is to capture the complexity of ontologies, particularly as ABox intensity increases. The features proposed in the research contribute to greater accuracy in predicting time consumption for ontological reasoning tasks.

Using machine learning techniques can also accelerate the performance of OWL reasoners by reducing the complexity of reasoning tasks.
In the~\cite{mehri_machine_2021} study, the aim is to improve the performance of reasoning systems by applying heuristic optimization techniques assisted by machine learning (ML). The authors use feature reduction techniques, in particular principal component analysis (PCA), to transform features into a set of new non-linearly correlated features. In addition, they use the support vector machine (SVM), well-suited to binary classification in high-dimensional feature spaces.

\subsubsection{Perform reasoning}

The capacity to perform deductive reasoning through a learning model, of logic-based symbolic formalisms, is a recent area of research~\citep{hohenecker_ontology_2020}. The two most recent papers reviewed in this study utilize neural networks to accomplish this task.

The first approach proposed by~\cite{rizzo_tree-based_2017} integrates the use of decision trees, in particular Random Forest (RF), enabling the construction of terminological decision trees to help reasoning processes in the Semantic Web environment.
In the study by~\cite{hohenecker_ontology_2020}, a new model architecture called Recursive Reasoning Network (RRN) is developed to perform this deductive reasoning task.~\cite{makni_deep_2019} focus on noise-tolerant reasoning in ontologies, recognizing the challenge of noise tolerance as a major bottleneck in deductive reasoning. To address this, they employ a recurrent neural network (RNN) to achieve noise-tolerant reasoning capabilities in ontology, this work is particularly interesting for dealing with noisy data commonly encountered in real-world applications.

{ \tiny  
\begin{longtable}{@{}p{0.1\textwidth}p{0.15\textwidth}p{0.1\textwidth}p{0.1\textwidth}p{0.08\textwidth}p{0.1\textwidth}cc@{}}
\caption{Details of articles in the Learning-Enhanced Ontology category}
\label{tab:learning-enhanced_ontology}\\

        \toprule
          Category               			& Sub-category           				& AI Theme 		& Application domain 	& Learning type             	& Learning algorithm 	& Reasoning & Paper    \\ \midrule   \endhead
          
          Ontology learning         		&  Automatic taxonomy construction		& NLP    			& Wikipedia			& Supervised			& Markov Logic Network	& \tikzxmark & \citep{wu_automatically_2008}  \\ 
							&							  	& NLP			& Technology			& 					& CRF				&  \tikzcmark & \citep{song_automated_2016}  \\
          						&							  	& NLP			& Cybersecurity		& 					& CRF				&  \tikzcmark & \citep{jia_practical_2018}  \\ 
         						&								& NLP, Translation	& -					&  					& RNN (Seq2Seq)		& \tikzxmark & \citep{petrucci_expressive_2018} \\ 
          						&								& NLP      			& Biomedical			& 					& RNN, Naive Bayes 	& \tikzxmark & \citep{zhao_domain-specific_2018}  \\  
							&							  	& NLP 			& Biomedical			& 					& SVM 				& \tikzxmark & \citep{ghoniem_novel_2019}  \\  \cmidrule(r){5-8}
							&							  	& NLP    			& -					& Unsupervised		& FCA 				& \tikzxmark & \citep{jurkevicius_ontology_2010}  \\
							&							  	& NLP    			& Tourism				& 					& HAC 				& \tikzxmark & \citep{getahun_integrated_2017}  \\ 
							&							  	& NLP    			& -					& 					& LDA, LSI, SVD 		& \tikzxmark & \citep{rani_semi-automatic_2017}  \\ \cmidrule(r){5-8} 
							&							  	& NLP 			& Linguistic			& Self-supervised 		&  Word2Vec (CBOW)	& \tikzxmark & \citep{albukhitan_arabic_2017}  \\ 
							&							  	& NLP 			& Tourism				& 			 		&  Word2Vec			& \tikzxmark & \citep{getahun_integrated_2017}  \\ \cmidrule(r){2-8}
          						& Learning non-hierarchical relations		& NLP, Translation	& -					& Supervised 			& RNN (Seq2Seq)		& \tikzxmark & \citep{petrucci_expressive_2018} \\ \cmidrule(r){5-8}
							&							  	& NLP 			& Tourism				& Self-supervised		& Word2Vec			& \tikzxmark & \citep{getahun_integrated_2017}  \\ \cmidrule(r){5-8}
							&								& NLP 			& Linguistic			& Unsupervised 		& Clustering			& \tikzxmark & \citep{albukhitan_arabic_2017}  \\  \cmidrule(r){2-8}
							& Rule discovery 					& NLP    			& -	 				& Supervised			& ANN				& \tikzxmark & \citep{jurkevicius_ontology_2010}  \\   
							& 			 					& - 				& Building Energy Management System& 		 	& M5, ANN, GA,  		&  \tikzcmark & \citep{mcglinn_usability_2017} \\ 
							& 			 					& - 				& Additive Manufacturing	& 				 	& CART 				&  \tikzcmark & \citep{ko_machine_2021} \\  \cmidrule(r){5-8} 
							& 			 					& - 				& Finance				& Unsupervised	 	& APRIORI			&  \tikzcmark & \citep{yang_construction_2020} \\ \cmidrule(r){2-8}
							& Ontology population 				& NLP			& Web				& Supervised			& Naïve Bayes			& \tikzxmark & \citep{craven_learning_2000}\\   
							& 								& -				& Healthcare			& 					& CRF				&  \tikzcmark & \citep{rubrichi_system_2013}\\  
							& 								& NLP, Spatial information extraction	& -		& 					& SVM				&  \tikzcmark & \citep{kordjamshidi_global_2015} \\ 
							& 								& NLP			& Robotics			& 					& SVM				& \tikzxmark & \citep{markievicz_action_2015} \\
							& 								& -				& History				& 					& HMM				& \tikzxmark & \citep{packer_cost-effective_2015}\\ 
							& 								& Computer vision, Large-scale visual recognition& -		& 			& Decision tree, CNN	& \tikzxmark & \citep{kuang_integrating_2018}\\
							& 								& NLP			& Biochemistry 			& 					& ANN				& \tikzxmark & \citep{ayadi_ontology_2019}\\ \cmidrule(r){5-8}
							& 								& NLP			& Biochemistry			& Self-supervised		& Word2Vec			& \tikzxmark & \citep{ayadi_ontology_2019}\\  \cmidrule(r){2-8}
							& Ontology enrichment 				& -				& Biomedical 			& Supervised			& HMM				& \tikzxmark & \citep{valarakos_building_2006}  \\ 
							& 								& -				& Food industry			& 					& CART, C4.5			& \tikzxmark & \citep{thomopoulos_iterative_2013}  \\
							& 								& Causal discovery	& Healthcare			& 					& CBN	 (SemCaDo)	&  \tikzcmark & \citep{messaoud_semcado_2015}  \\ 
							&							  	& -				& Technology			& 					& CRF				&  \tikzcmark & \citep{song_automated_2016}  \\ 
							& 								& KB quality assessment& -				& 					& RF					& \tikzxmark & \citep{mihindukulasooriya_rdf_2018}  \\ 
							&							  	& NLP			& Healthcare			& 					& biLSTM, CRF			&  \tikzcmark & \citep{hong_constructing_2021}  \\ 
							& 								& -				& Web vocabularies		& 					& Multiple algorithm		& \tikzxmark & \citep{merono-penuela_multi-domain_2021}  \\ \cmidrule(r){5-8}
							& 								& -				& Biomedical 			& Unsupervised		& COCLU				& \tikzxmark & \citep{valarakos_building_2006}  \\ 
							&							  	& NLP  			& -					& 					& TSVD, Fuzzy ART		& \tikzxmark & \citep{djellali_using_2013}  \\  \cmidrule(r){1-8}
          Ontology mapping             	&         							& Constraint satisfaction problem & Web 		&  Supervised          		& IAC           			& \tikzxmark & \citep{mao_adaptive_2010}\\ 
          				                	&         							& -           			& - 					&           				& Ensemble classifier       	& \tikzxmark & \citep{fanizzi_composite_2011} \\ 
							&         							&  -         			& -  					&           				& MLP        	  		& \tikzxmark & \citep{rubiolo_knowledge_2012} \\
          						&								& Anchoring         	& -   					& 		         		& RF          			& \tikzxmark & \citep{annane_building_2018} \\
							&         							& -          			& -   					&           				& ANN            			& \tikzxmark & \citep{gao_ontology_2018}\\ 
          						&         							& -           			& Wikipedia  			&          				& RF, MLP, SMO     		& \tikzxmark & \citep{rico_predicting_2018} \\
							&         							&  NLP     			& Healthcare   			&           				& MLP        	  		& \tikzxmark & \citep{shannon_comparative_2021} \\  \cmidrule(r){5-8}
							&         							& -            			& Biomedical 			& Self-supervised         	& BERT           			& \tikzxmark & \citep{mohan_low_2021} \\
							&         							& -				& Building				&           				& Word2Vec           		& \tikzxmark & \citep{zhou_semantic_2021} \\ \cmidrule(r){5-8}
							&         							& -         			& Wikipedia   			& Unsupervised         	& PCA          			& \tikzxmark & \citep{rico_predicting_2018}\\ 
							&         							& -            			& Web 				&           				& LSTM           			&  \tikzcmark & \citep{chakraborty_ontoconnect_2021} \\ 
							&         							& Competitive learning & Sensor, IoT 		&           				& ANN           			& \tikzxmark & \citep{xue_matching_2021} \\ \cmidrule(r){1-8} 
	 Learning-based reasoning	& Optimizing reasoning 	       			& -        			& -   					& Supervised        		& Naive Bayes, k-NN, SVM, Decision tree &  \tikzcmark &  \citep{bock_automatic_2012}  \\
							&   								& -        			& -   					& 			       		& RF	 (and Boruta algorithm)&  \tikzcmark & \citep{pan_predicting_2018} \\ 
          						& 				       				& -       			& -   					&         				& SVM       			&  \tikzcmark & \citep{mehri_machine_2021}   \\ \cmidrule(r){5-8}
							& 				       				& -        			& -   					& Unsupervised        		& PCA       			&  \tikzcmark & \citep{mehri_machine_2021} \\ \cmidrule(r){2-8} 
							& Perform reasoning 				& Inductive Logic Programming & Semantic Web &					& Decision Tree, RF		& \tikzcmark  & \citep{rizzo_tree-based_2017}  \\ 
							&        							& Noise-tolerance	& Web   				& Supervised      		& RNN      			&  \tikzcmark &  \citep{makni_deep_2019} \\  
							&         							& -        			& -  					&         				& RRN      			&  \tikzcmark &  \citep{hohenecker_ontology_2020} \\

        \bottomrule
\end{longtable}
}
\section{Semantic Data Mining}
\label{semantic_data_mining}

In this section, we present \emph{Semantic Data Mining} using knowledge from ontology to improve the performance of machine learning algorithms~\citep{lawrynowicz_pattern_2014}. Semantic data mining is a particular form of Informed Machine Learning defined by~\cite{von_rueden_informed_2021} and means "using hybrid information source that consists of data and prior knowledge in machine learning"\footnote{Dedicated terms to reference articles about informed machine learning in the literature. Defines here as \emph{The prior knowledge comes from an independent source, is given by formal representations, and is explicitly integrated into the machine learning pipeline} \citep{von_rueden_informed_2021}.}. The term \emph{informed machine learning} is particularly present in the field of physics \citep{karniadakis_physics-informed_2021}, where the concern is to embed physics into machine learning to improve the results and better adapt the algorithms to the complexity of physical problems. 

In other words, semantic data mining is a combination of data-driven and ontology-driven approaches. This knowledge can be added to machine learning at different stages of the machine learning pipeline \citep{von_rueden_informed_2021}, like during the training data stage (\emph{Ontology-based feature engineering}), the hypothesis set stage (\emph{Ontology-based algorithm design}), the learning algorithm stage (\emph{Ontology-based algorithm training}) or at final hypothesis stage (\emph{Ontology-based explanation}). 

The comprehensive details of all papers within this category are outlined in Table~\ref{tab:semantic_DM_algo}, where they are meticulously categorized by their respective field of application, AI themes they explore, and the machine learning algorithms employed.

\subsection{Ontology-based feature engineering} 

At the first step of machine learning processes (i.e. training data), \emph{Ontology-based feature engineering} allows mixing raw data with prior knowledge in several ways according to feature engineering definition~\citep{duboue_art_2020}: feature augmentation, feature selection, feature extraction or semantic embedding. 

\begin{figure}[ht]
\centering
\includegraphics[width=0.2\textwidth]{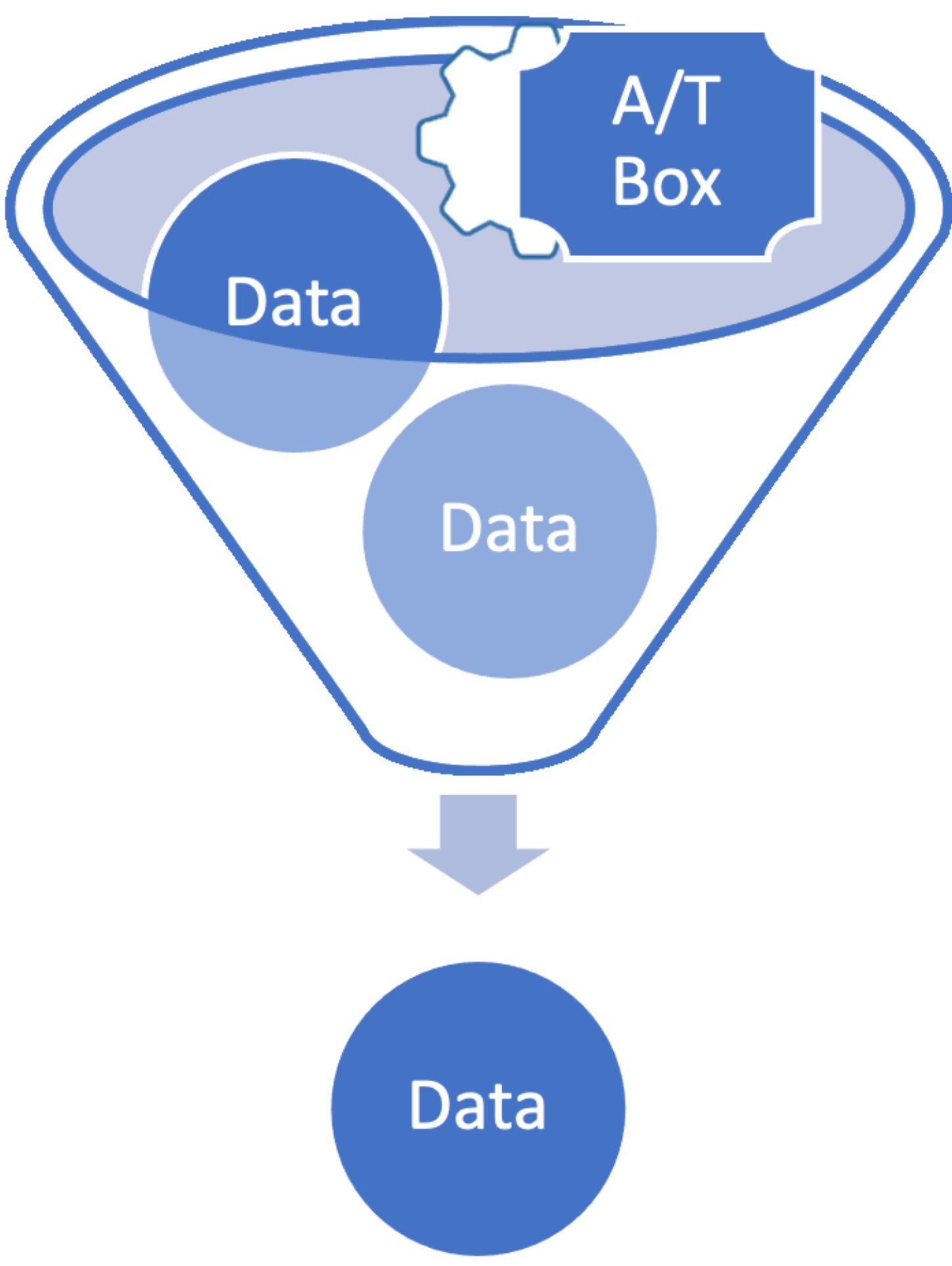}
\caption{Ontology-based feature engineering mechanism}
\label{fig:14}
\end{figure}

In the context of this review, studies categorized in this domain utilize a hybrid source of data, incorporating one or more ontologies, symbolized by the annotated polygon "A/T Box," to produce final results in the form of data, as illustrated in Figure~\ref{fig:14}.

\subsubsection{Feature augmentation}

\emph{Feature augmentation} involves adding new variables (features) derived from the prior knowledge present in the ontology to the original dataset. This process does not always involve logical reasoning; the ontology's semantic structure alone can be sufficient to enrich the original dataset with valuable information that was not available at the outset. The primary objective is to enrich the input features with relevant ontological knowledge, providing the model with a deeper understanding of the domain under consideration. Using ontology to augment features can improve data representation, raise model performance and contribute to more insightful decision-making in a variety of applications.
 
In the field of NLP, and more specifically Automatic Language Processing (ALP), the incorporation of additional data into the initial dataset can prove valuable, as demonstrated by~\cite{pozveh_fnlp-ont_2018} through tasks such as POS labeling, Named Entity Recognition (NER) and Word Disambiguation (WD). In the field of applications for intelligent environments, Ye et al. (2015) used k-means clustering for activity recognition from sensor data augmented with an ontology. In the same line, Salguero et al. (2019) used support vector machines (SVMs) for the recognition of everyday activities from the same type of data. 
As shown in Table~\ref{tab:semantic_DM_algo}, the addition of new features using ontology can be integrated with a wide range of machine learning algorithms.

\subsubsection{Feature selection}
  
\emph{Feature selection} aims at reducing the number of variables to keep only the most relevant without changing the initial variables~\citep{gomathi_ontology_2019}. Unlike traditional feature selection methods, ontology-based feature selection exploits the semantic relationships and structures defined in an ontology to identify and prioritize features for inclusion or exclusion. This approach aims to improve the selection process by incorporating domain-specific semantics, ensuring that selected features align with underlying ontological concepts. Note that this technique can sometimes be useful when faced with the curse of dimensionality~\citep{bellman_adaptive_1961}.
  
  \subsubsection{Feature extraction}
  
\emph{Feature extraction} is the process of transforming raw data into a reduced, relevant representation, highlighting important features for subsequent analysis. Ontology-based feature extraction involves modifying the original variables based on the prior knowledge provided by the ontology to obtain relevant features. This process aims to derive relevant features by exploiting the semantic information embedded in the ontology and allows input variables to be tailored to improve analysis and model performance.

Feature extraction often involves textual data, implying the use of NLP techniques, especially in sentiment analysis where neural networks are frequently employed~\citep{kumar_aspect-based_2020, sabra_hybrid_2020, ahani_evaluating_2021}. 
In computer vision, leveraging ontology can assist in extracting meaningful features from images. Typically, a pre-processing step is employed to transform images into information effectively used by the ontology. Next, the data is often processed by a neural network.~\cite{akila_ontology_2021} used ANN for sports image feature extraction, and \citep{zhao_adaptive_2021} for industrial vision inspection.~\cite{messaoudi_ontology-driven_2021} applied convolutional neural networks (CNN) to healthcare for MRI data classification, while ~\cite{rinaldi_semantic_2021} employed the VGG16 model for feature extraction from both textual documents and pre-classified images.
The feature extraction process can also help reveal meaningful patterns and temporal relationships in time series, facilitating predictions made on this kind of data. In the study conducted by~\cite{liu_photovoltaic_2021} on photovoltaic time series data, the authors used managed recurrent units (GRUs) for feature extraction. GRUs, a type of recurrent neural network (RNN), are particularly effective at capturing temporal dependencies in sequential data.

\subsubsection{Semantic embedding}

In \emph{Semantic embedding}, raw data is refined by semantic knowledge and then transformed into vectors to be exploited mainly by neural networks. This explains the prevalence of neural networks in this category since the data is specifically transformed for them. However, it is noteworthy that~\cite{mabrouk_exploiting_2020} employs semantic embedding for an SVM, while~\cite{zhang_auto_2021} utilizes it for XGBoost. The oldest paper in this SLR that uses this technique is from 2018, we can therefore assume that research in this field is recent.

This category encompasses numerous papers that leverage ontology to create knowledge graph embeddings (KGE)~\citep{chen_knowledge_2021}. KGE employs models like TransE, TransR, DistMult, etc., each with a score function to convert the graph's knowledge into vectors usable by machine learning algorithms. In our study, the primary applications of this technique are in automatic text processing and time series analysis. These transformations are conducted with meticulous consideration for preserving the links between different entities within the KGE.
Ontology embedding expands upon this representation, encompassing a broader scope of ontological knowledge, including aspects like existential rules. In their work,~\cite{benarab_ontology_2019} employs autoencoders for the implementation of ontology embedding.
Word2Vec is also frequently used to pre-process plain text before the semantic embedding stage~\citep{jang_cross-language_2018, ali_intelligent_2021, amador-dominguez_ontology-based_2021}.

\subsection{Ontology-based algorithm design}

In the second stage of the machine learning process (i.e. the hypothesis set), \emph{ontology-based algorithm design} contributes to the incorporation of ontological knowledge into the design and development of machine learning algorithms.

\subsubsection{Ontology-based decision tree}

Ontology-based decision trees refer to an algorithmic design approach that incorporates ontological principles into the construction and use of decision trees. In this context, decision trees, such as random forests, are developed and used in a way that incorporates ontological knowledge. 
\cite{emele_learning_2012} introduced an ontology-based decision tree called STree, derived from C4.5 and enhanced with ontological reasoning, applied to military dialogues. 

\subsubsection{Ontology-based probabilistic graphical model}

An ontology-based probabilistic graphical model refers to an algorithmic design approach that incorporates ontological principles into the construction and use of probabilistic graphical models, such as Bayesian networks or Markov models.

\cite{ruiz-sarmiento_ontology-based_2019} proposed an ontology-based probabilistic graphical model, specifically Ontology-based Conditional Random Fields (obCRFs), for robotics in computer vision tasks. This model enhances standard Conditional Random Fields (CRFs) with additional nodes and relations based on a multi-level ontology structure, aligning with the subsumption ordering of ontologies, to improve object recognition in robot environments. 

\subsubsection{Ontology-based neural topology}

Ontology-based neural topology entails an algorithmic design approach that incorporates ontological principles into the selection or creation of the architectural design for neural networks. This methodology involves integrating ontological insights to guide the structure and configuration of neural networks, aligning them with domain-specific knowledge and semantic relationships. 

\cite{gabriel_neuroevolution_2014} facilitate the choice of an appropriate ANN model structure thanks to an ontology. Rather than going through a grid-search step which is sometimes too time-consuming in complex systems, a model topology (e.g. the number of hidden layers and the number of neurons in the hidden layers for a neural network) can be approximated by prior knowledge. 
\cite{huang_enhancing_2019}  introduced OntoLSTM, an ontology-based long-term memory neural network (LSTM), wherein dense layers are encoded using ontology-derived information. This approach is specifically designed for time series analysis within the context of Industry 4.0.
\cite{kuang_deep_2021} addressed large-scale fashion recognition using a hierarchical deep learning approach called Augmented Hierarchical Deep Learning (AHDL). The proposed hierarchical knowledge distillation method facilitates knowledge transfer between tree node classifiers of hierarchical deep networks, thus improving fashion image representation and classification.	
\cite{fu_tagging_2015}  focused on personal photo tagging using transfer learning with a Convolutional AutoEncoderS (CAES) to which they added a Fully Connected layer with Ontology priors (FCO). Their approach exploits ontology priors in the last layer of a fully connected network to improve personal photo tagging performance.

\subsection{Ontology-based algorithm training}

At the third step of the machine learning process (i.e. learning algorithm), \emph{ontology-based algorithm training} integrated prior knowledge into the machine learning algorithm, typically via a loss function. 

In their work,~\cite{serafini_learning_2017} introduced ontology-based algorithm training through the application of a Logical Tensor Network (LTN). The LTN framework integrates logical reasoning into deep learning architectures using t-norms derived from fuzzy logic. This innovative approach allows logical constraints to be added to the inductive reasoning process.

\subsection{Ontology-based explanation}

In the realm of artificial intelligence research, neural networks are frequently regarded as ``black boxes", where the explicit input-output behavior of the algorithm is observable, but the underlying reasoning mechanism remains opaque. Therefore, the development of explainable artificial intelligence (XAI) becomes crucial. The explainability of an algorithmic model pertains to its ability to present a coherent sequence of interconnected steps that can be interpreted by humans as causes or reasons behind the decision-making process~\citep{donis_ebri_combining_2021}. This capability allows for the clarification of the algorithm or its outputs, enhancing the understanding of how and why certain decisions are made.
In addition to the issue of trust, the lack of explainability in AI models has also given rise to legal challenges in various domains such as military defense, healthcare, insurance, and autonomous vehicles. The inability to provide clear and understandable explanations for AI-driven decisions poses legal complications in these areas. The global explainability of a model aids in identifying the key variables that contribute to the model's output. It enables the determination of the specific role played by a particular variable in the model's final decision or prediction. Global explainability is used to assess the importance of a model's features. The SHAPE algorithm~\citep{lundberg_unified_2017} is commonly employed to identify variables with the most significant impact in a machine learning model. Conversely, the local explainability of a model focuses on the process leading to decisions made for a specific individual~\citep{ribeiro_why_2016, lundberg_unified_2017}. It aims to highlight the impact of each variable on the outcome, thereby making the decision more interpretable and understandable for that particular case. 

\begin{figure}[ht]
\centering
\includegraphics[width=0.4\textwidth]{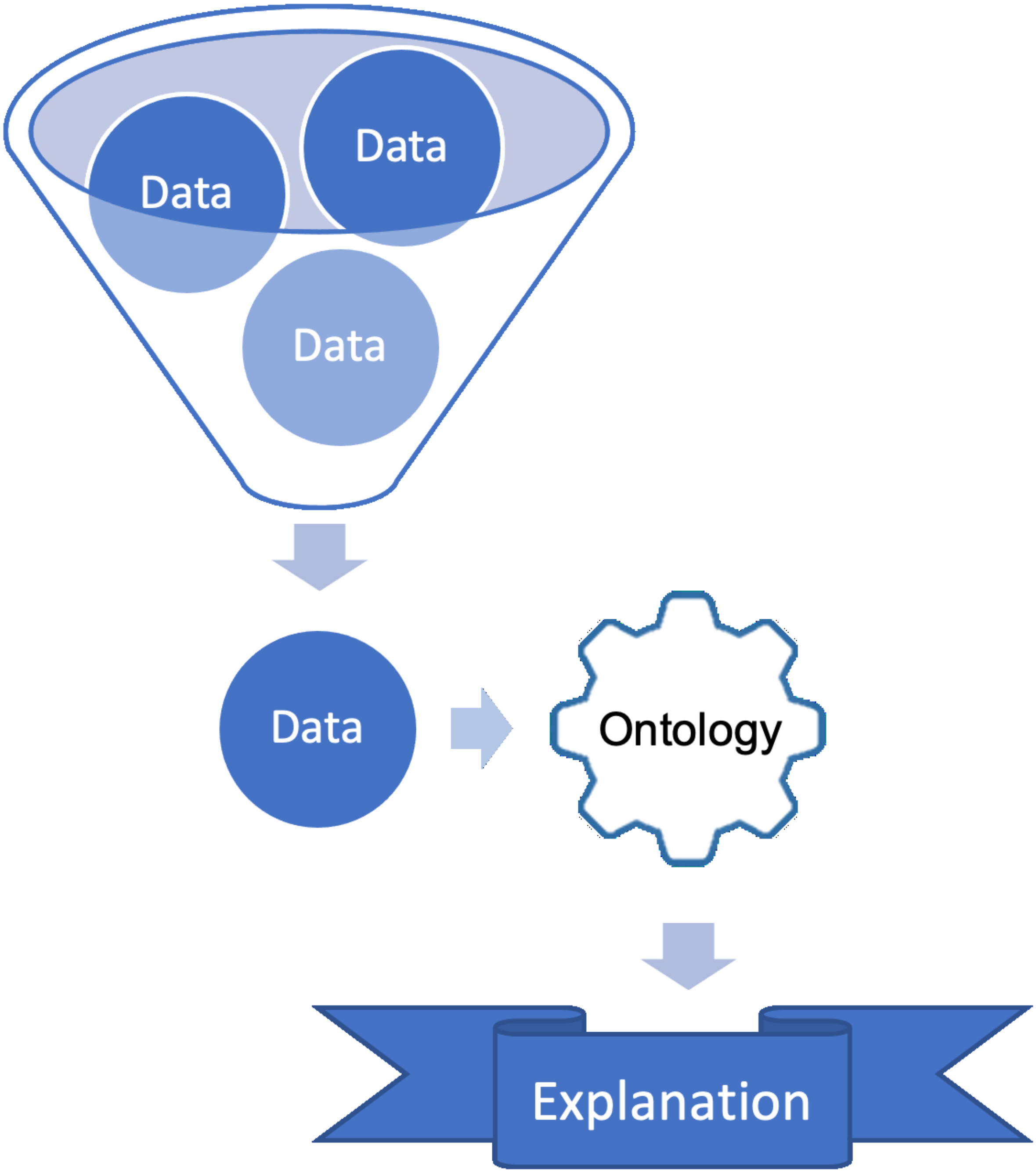}
\caption{Ontology-based explanation mechanism}
\label{fig:15}
\end{figure}

When prior knowledge is not integrated, explanations are primarily based on mathematical correlations between data and results, which does not always guarantee the robustness and reliability of indicators. Black-box explanation through an ontology relies on the idea of using a formal and explicit knowledge structure to clarify the internal workings of a considered AI model, often viewed as a "black box." This means that when a prediction is generated by a conventional machine learning model, as represented by a funnel in Figure~\ref{fig:15}, the ontology is then used to provide explanations for that prediction as shown in the same figure.

For a comprehensive explanation of the model, the ontology can be used to demonstrate how concepts and entities in the ontology are related to the features or input data of the model, thereby describing the overall reasoning process of the model. For a local explanation, the ontology can be employed to highlight how specific concepts or entities in the ontology contributed to the particular prediction for a given individual. In this SLR, two studies use ontologies to enhance the explainability of models: one of these studies aimed to provide a global explanation of the model~\citep{confalonieri_using_2021}, while the other focused on delivering local explanations~\citep{panigutti_doctor_2020}.

These studies demonstrate the usefulness of ontologies in augmenting the interpretability of AI models, both at the global and local levels. The integration of prior knowledge through ontologies helps establish a logical and coherent framework for the model's explanations, aligning them with existing domain knowledge. This not only enhances the trustworthiness of the explanations but also provides a deeper understanding of the reasoning process employed by the model.

Table \ref{tab:semantic_DM_algo} presents machine learning algorithms used in \emph{Ontology-based explanation} category. Only neural networks are presented here, \cite{panigutti_doctor_2020} specifies that it is a Gated Recurrent Unit network (GRU), a special type of Recurrent Neural Network.

{\tiny 
\begin{longtable}{@{}p{0.09\textwidth}p{0.1\textwidth}p{0.1\textwidth}p{0.1\textwidth}p{0.08\textwidth}p{0.1\textwidth}cc@{}}
\caption{Details of articles in the Semantic Data Mining category}
\label{tab:semantic_DM_algo}\\

        \toprule
          Category               						& Sub-category            	& AI Theme		& Application domain		& Learning type		&  Learning algorithm	& Reasoning & Paper    \\ \midrule   \endhead
          Ontology-based feature engineering     		& Feature augmentation	& Automatic Task Detection& Technology		& Supervised		& J48				& \tikzxmark  & \citep{rath_uico_2009} \\ 
          									&  					& NLP			& Cybersecurity (Spam detection) & 			& Bayesian network		& \tikzxmark  & \citep{santos_enhanced_2012}  \\ 
          									&					& Pattern discovery	& -					& 				& RF					& \tikzxmark  & \citep{lawrynowicz_pattern_2014}  \\ 
										& 					& -				& -					& 				& ANN				& \tikzxmark  & \citep{pancerz_encoding_2014}  \\ 
										& 					& NLP, TAL		& Persian text 			& 				& HMM				& \tikzxmark  & \citep{pozveh_fnlp-ont_2018}  \\  
										& 					& - 				& Biomedical			& 				& SVM				& \tikzxmark  & \citep{wan_predicting_2018}  \\ 
										& 					& Activity recognition	& Smart environments	& 				& SVM				& \tikzxmark  & \citep{salguero_methodology_2019}  \\ 
										& 					& NLP			& Healthcare 			& 				& RNN, CNN, HAN		& \tikzxmark  & \citep{abdollahi_substituting_2021}  \\  
										& 					& -				& Healthcare 			& 				& ANN				& \tikzxmark  & \citep{wang_evaluating_2021} \\ \cmidrule(r){5-8}
										& 					& Activity recognition	& Smart home			& Unsupervised	& k-means			& \tikzxmark  & \citep{ye_usmart_2015}  \\ 
										& 					& Recommendation system& Wine			& 				& Farthest First (k-means)	& \tikzxmark  & \citep{oliveira_wine_2021}  \\ \cmidrule(r){2-8}
										& Feature selection		& -				& Healthcare 			& Supervised		& ANN, SVM			& \tikzxmark  & \citep{gomathi_ontology_2019}  \\  \cmidrule(r){2-8}
										& Feature extraction		& MAS, Planning	& Tourism				& Supervised 		& C4.5, k-NN			& \tikzxmark  & \citep{castillo_samap_2008}  \\ 
										& 					& Semantic annotation& -					& 				& Bayesian network		& \tikzxmark  & \citep{rajput_bnosa_2011}  \\ 
										& 					& -				& Healthcare 			& 				& Bayesian networks, ANN, SVM, regression (WEKA)	& \tikzxmark  & \citep{hsieh_transformation_2013}  \\ 
										& 					& NLP, Sentiment analysis& -				& 				& SVM				& \tikzxmark  & \citep{agarwal_concept-level_2015, manuja_intelligent_2015}  \\ 
										& 					& MAS			& Contexte-aware		& 				& ANN				& \tikzcmark & \citep{yilmaz_matching_2017}  \\ 
										& 					& NLP			& Finance, Corporate disclosures& 			& Naïve Bayes			& \tikzxmark  & \citep{evert_combining_2019}  \\ 
										& 					& User Interface 	& Healthcare 			& 				& SVM				& \tikzxmark  & \citep{greenbaum_improving_2019}  \\ 
										& 					& NLP			& Spam filtering		& 				& RF, SVM, C4.5, Naïve Bayes, LR, Adaboost, bagging& \tikzxmark  & \citep{mendez_new_2019}  \\ 
										& 					& -     			& Healthcare 			& 				& LR 				& \tikzxmark  & \citep{radovanovic_framework_2019}  \\ 
										& 					& NLP, Sentiment analysis& -				& 				& CNN				& \tikzxmark  & \citep{kumar_aspect-based_2020}  \\ 
										& 					& NLP, Sentiment analysis	& Healthcare 	& 				& MLP, SVM, ensemble classifier	& \tikzcmark & \citep{sabra_hybrid_2020}  \\ 
          									& 					& NLP, Sentiment analysis	& Healthcare	& 				& k-NN, ANFIS			& \tikzxmark  & \citep{ahani_evaluating_2021}  \\
										& 					& Computer Vision 	& Sport 				& 				& ANN				& \tikzxmark  & \citep{akila_ontology_2021}  \\ 
										& 					& Time series		& Photovoltaic			& 				& GRU				&  \tikzcmark & \citep{liu_photovoltaic_2021}  \\ 
										& 					& Computer Vision	& Healthcare  			& 				& CNN				& \tikzxmark  & \citep{messaoudi_ontology-driven_2021}  \\ 
										& 					& NLP			& Healthcare 			& 				& LSTM				& \tikzxmark  & \citep{nayak_experience_2021}  \\
										& 					& Computer Vision, NLP	& -				& 				& VGG16				& \tikzxmark  & \citep{rinaldi_semantic_2021}  \\
										& 					& Computer Vision	& Industry (Industrial vision)& 				& ANN				& \tikzxmark  & \citep{zhao_adaptive_2021}  \\ 
										& 					& -				& Industry	4.0 (Condition monitoring)	& 		& LSTM				& \tikzcmark & \citep{zhou_semml_2021}  \\  
										& 					& NLP			& -					& 				& ANN				& \tikzxmark  & \citep{deepak_artificially_2022}  \\  \cmidrule(r){5-8}
										& 					& NLP			& -					& Self-supervised	& Word2Vec			& \tikzxmark  & \citep{kumar_aspect-based_2020}  \\ \cmidrule(r){5-8}
										& 					& -				& Biomedical 			& Unsupervised	& FCA				& \tikzxmark  & \citep{akand_learning_2007}  \\ 
										& 					& MAS, Planning	& -					& 				& Clustering, EM		& \tikzxmark  & \citep{castillo_samap_2008}  \\ 
										& 					& NLP			& -					& 				& HAC				& \tikzcmark  & \citep{radinsky_learning_2012}  \\ 
										& 					& NLP			& Finance, Corporate disclosures& 			& LDA, LSI			& \tikzxmark  & \citep{evert_combining_2019}  \\ 
										& 					& NLP, Sentiment analysis	& Healthcare	& 				& EM, LDA, Hot-Deck	& \tikzxmark  & \citep{ahani_evaluating_2021}  \\ 
										& 					& Computer Vision	& -					& 				& k-means			& \tikzxmark  & \citep{akila_ontology_2021}  \\ 
										& 					& NLP 			& Biomedical			& 				& k-means			& \tikzxmark  & \citep{perez-perez_framework_2021}  \\ \cmidrule(r){2-8}
										& Semantic embedding	& NLP			& Dialog state tracking	& Supervised		& bi-LSTM			& \tikzxmark  & \citep{jang_cross-language_2018}  \\ 
										& 					& NLP, Sentiment analysis & Transport		& 				& SVM, LR, MLP, kNN, Naïve Bayes, Decision tree, DNN	& \tikzxmark  & \citep{ali_transportation_2019}  \\ 
										& 					& NLP			& Healthcare 			& 				& CNN				& \tikzxmark  & \citep{gaur_knowledge-aware_2019}  \\ 
										& 					& NLP			& Linguistic 			& 				& bi-LSTM			& \tikzxmark  & \citep{moussallem_utilizing_2019}  \\ 
										& 					& NLP			& Healthcare 			& 				& MLP				& \tikzxmark  & \citep{hassanzadeh_matching_2020}  \\ 
										& 					& NLP			& - 					& 				& SVM				& \tikzxmark  & \citep{mabrouk_exploiting_2020}  \\ 
										& 					& NLP			& Building				& 				& RNN				& \tikzcmark & \citep{ren_information_2020}  \\ 
										& 					& NLP			& -					& 				& bi-LSTM			& \tikzxmark  & \citep{alexandridis_knowledge-based_2021}  \\ 
										& 					& NLP			& Healthcare 			& 				& bi-LSTM			& \tikzxmark  & \citep{ali_intelligent_2021}  \\ 
										& 					& Time series, concept drift& Smart City 		& 				& LR, RF, ASHT, leveraging bagging, SGD & \tikzcmark  & \citep{chen_knowledge_2021}  \\ 
										& 					& Time series		& Healthcare 			& 				& GRU				& \tikzxmark  & \citep{niu_fusion_2022}  \\ \cmidrule(r){5-8}	
										& 					& NLP  			& -					& Self-supervised	& Word2Vec			& \tikzxmark  & \citep{jang_cross-language_2018}   \\ 
										& 					& NLP, Sentiment analysis& Transport		& 				& Topic2Vec 			& \tikzxmark  & \citep{ali_transportation_2019}  \\ 
										& 					& -				& Biomedical			& 				& Autoencoders (multiple neural networks) & \tikzxmark  & \citep{benarab_ontology_2019}  \\
										& 					& NLP			& Geoscience			& 				& GloVe, Word2Vec, Doc2Vec  	& \tikzxmark  & \citep{qiu_geoscience_2019}  \\
										& 					& NLP			& Healthcare 			& 				& Word2Vec (skip-gram) 	& \tikzxmark  & \citep{ali_intelligent_2021}  \\ 
										& 					& NLP  			& -					& 				& Word2Vec			&  \tikzcmark & \citep{amador-dominguez_ontology-based_2021}   \\  \cmidrule(r){5-8}
										& 					& NLP, Sentiment analysis& Transport		& Unsupervised	& LDA				& \tikzxmark  & \citep{ali_transportation_2019}  \\
										& 					& NLP			& Healthcare 			& 				& LDA, Information Gain	& \tikzxmark  & \citep{ali_intelligent_2021}  \\  \cmidrule(r){1-8}
	Ontology-based algorithm design			& Ontology-based decision tree	& NLP (Dialogue)& Military			& Supervised		& C4.5				& \tikzcmark & \citep{emele_learning_2012}  \\  \cmidrule(r){2-8}
										& Ontology-based probabilistic graphical model	& Computer Vision	& Robotics& Supervised	& CRF				& \tikzxmark  & \citep{ruiz-sarmiento_ontology-based_2019}  \\ \cmidrule(r){2-8}
										& Ontology-based neural topology& Time series		& Industry 4.0			& Supervised		& LSTM				& \tikzxmark  & \citep{huang_enhancing_2019}  \\ 
										& 					& Computer Vision	& Fashion				& 				& CNN				& \tikzxmark  & \citep{kuang_deep_2021}  \\  \cmidrule(r){5-8}
										& 					& MAS			& Video Game			& Reinforcement	& ANN				& \tikzxmark  & \citep{gabriel_neuroevolution_2014}  \\  \cmidrule(r){5-8} 
										& 					& Computer Vision, Tagging& Personal Photos & Self-supervised (Transfer learning)	& CAES, FCO			& \tikzxmark  & \citep{fu_tagging_2015}  \\  \cmidrule(r){1-8} 
										
	Ontology-based algorithm training			& 					& Computer Vision	& - 					& Supervised		& LTN				&  \tikzcmark & \citep{serafini_learning_2017}  \\ \cmidrule(r){1-8}
	Ontology-based explanation				&         				& -      			& Healthcare    			& Supervised    	& GRU         	  		& \tikzxmark  & \citep{panigutti_doctor_2020} \\ 
     	     									&         				& -           			& -   					& 	 		  	& ANN        			&  \tikzcmark  & \citep{confalonieri_using_2021} \\

        \bottomrule
\end{longtable}
}
\section{Learning and Reasoning Systems}
\label{combining_learning_reasoning}

This category represents the set of complete applications that use machine learning and ontologies to operate. The application is a computer program capable of performing one or more specific tasks in the same field, e.g. a decision support system for the management of cardiac pathology~\citep{ali_smart_2020}. Studies in this category, shown in Figure~\ref{fig:categories_combining_OWL_ML}, describe complete application systems, not only some specific mechanisms (e.g. ontology learning, semantic feature engineering, etc.).

\subsection{Expert System Embedded Learning}
\begin{figure}[ht]
\centering
\includegraphics[width=0.2\textwidth]{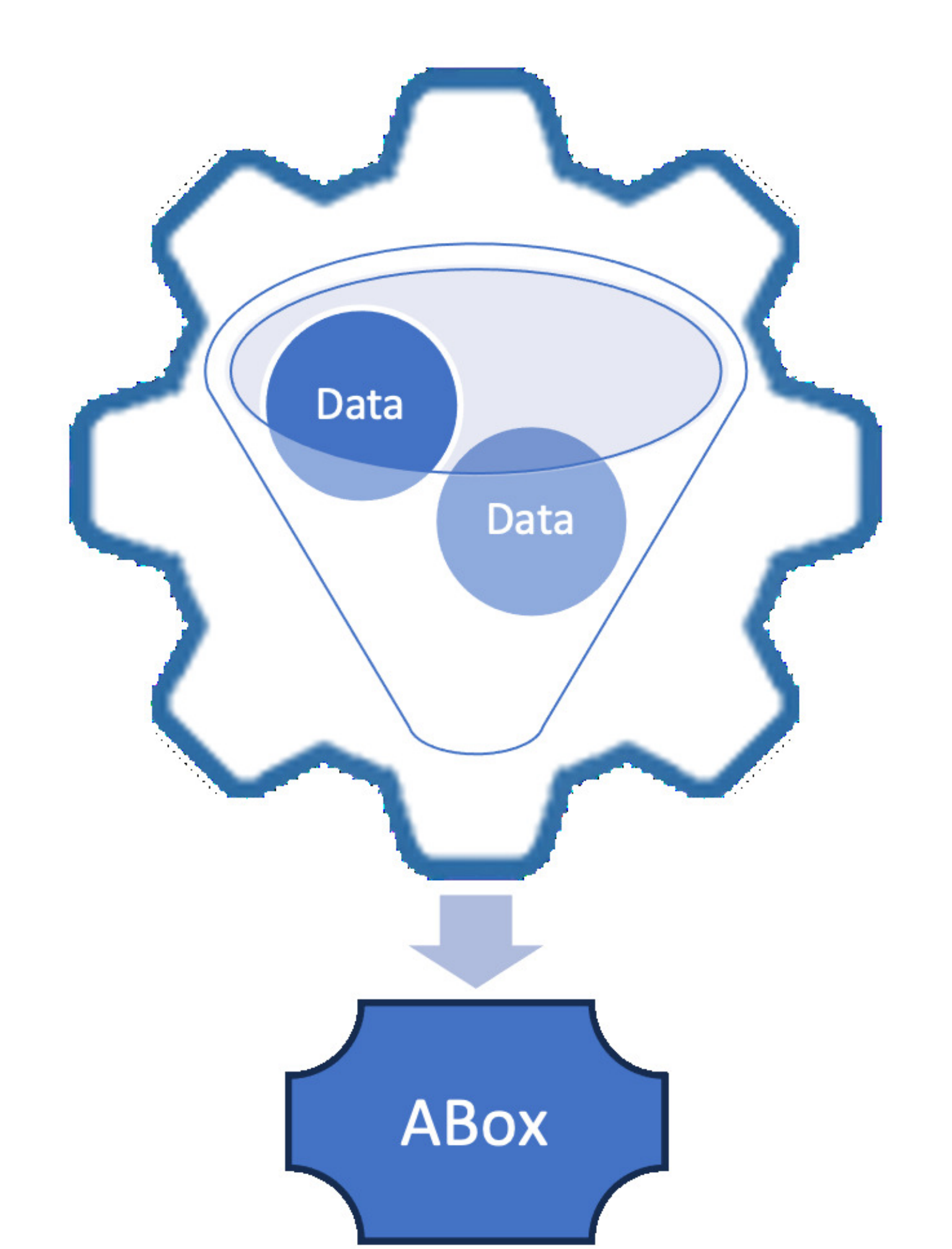}
\caption{Expert System Embedded Learning mechanism}
\label{fig:16}
\end{figure}

The first sub-category covers ontology-based expert systems that exploit machine learning for execution. An expert system comprises various components, including a knowledge base, an inference engine, and an interface~\citep{liebowitz_handbook_1997}. 
In \emph{Expert System Embedded Learning}, the machine learning component, represented in Figure~\ref{fig:16} by a funnel, is integrated into an expert system, here represented by a cogwheel symbolizing an ontology. The integrated learning model(s) can be considered as sub-modules of the expert system and the produced results mainly consist of new facts inferred by the expert system, represented by an annotated polygon "ABox".
This is why we have chosen to use the term "embedded", which is an equivalent of "integrated" because the proper functioning of the expert system is closely linked to the machine learning part.

The two articles in this category use machine learning to find missing values in an expert system, enabling it to perform deductive reasoning. They therefore perform a task similar to that performed by~\cite{makni_deep_2019}, which enables noise-tolerant deductive reasoning. 
However, their approach is different, as they don't use a machine learning algorithm to perform the reasoning, but rather to impute missing values. 
Moreover, in both instances, the decision-making system could theoretically operate without an external learning module, although with reduced performance. This distinction justifies their classification in the Expert System Embedded Learning category.

In the first study,~\cite{khan_validation_2013} introduced Holmes (Hybrid Ontological and Learning MEdical System), a medical system integrating ontology and machine learning for decision-making in patient treatment. Holmes employs Adaboost as its primary learning algorithm, enabling the creation of a semantic decision support system resilient to noise. Specifically, it addresses decision-making scenarios related to the administration of sleeping pills.
In the second study,~\cite{bischof_enriching_2018} presented the Open City Data Pipeline, which aims to collect, integrate, and enrich statistical data from various cities around the world for republication as machine-readable linked data. To handle missing values in the dataset, their imputation pipeline employs principal component analysis (PCA) as a pre-processing step. Subsequently, the authors apply different algorithms, including multiple logistic regression (MLR), k-nearest neighbors (k-NN), or random forest (RF), based on the performance obtained during the preprocessing phase.

\subsection{Hybrid application}
\label{reason_after_learning}

\begin{figure}[ht]
\centering
\includegraphics[width=0.2\textwidth]{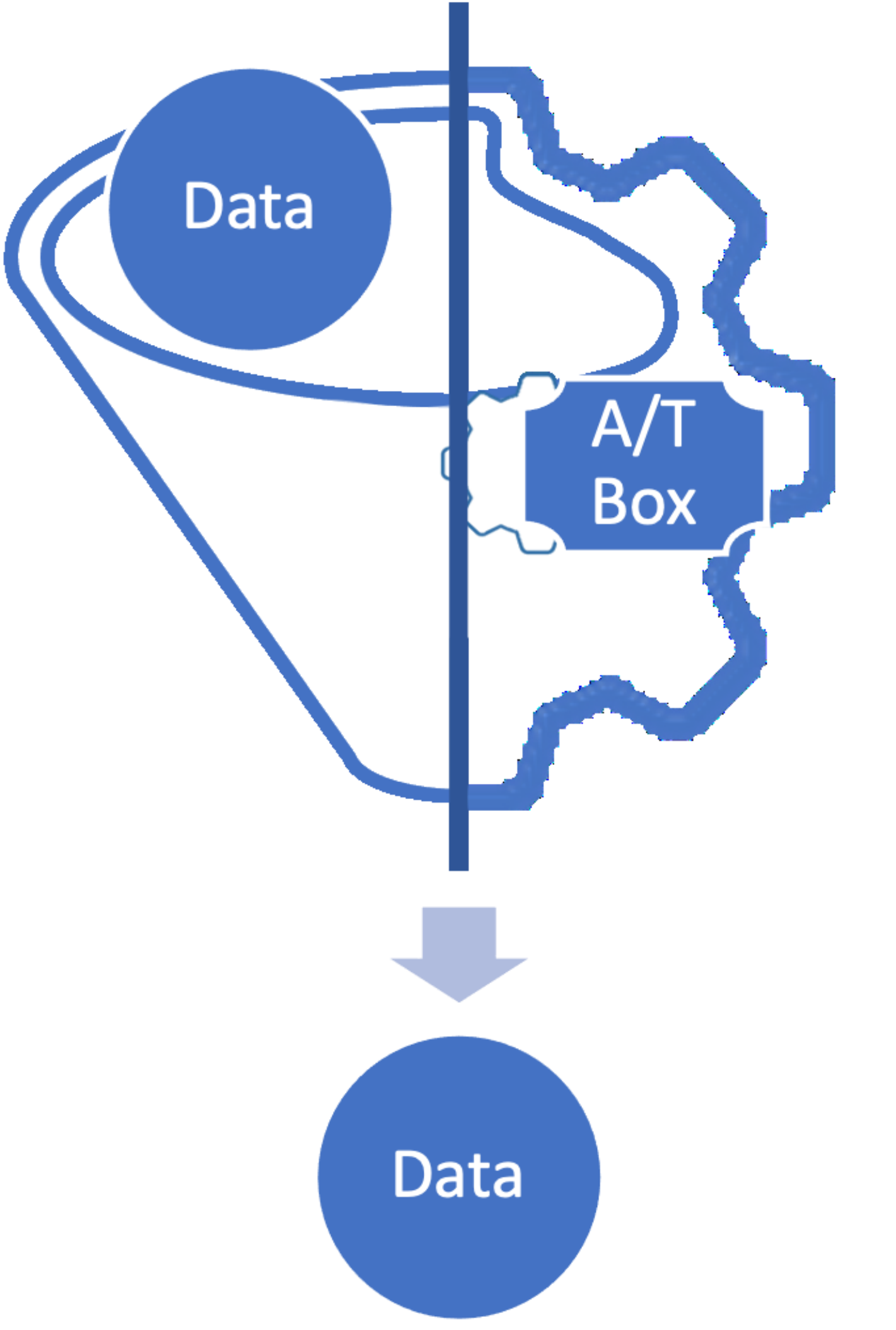}
\caption{Hybrid application mechanism}
\label{fig:17}
\end{figure}

This second sub-category represents the set of hybrid application systems that use machine learning and ontologies in a more complex way, often by communicating the two types of reasoning (inductive and deductive) within the AI system.
Figure~\ref{fig:17} clearly illustrates the hybridization mechanism by symbolizing the fusion between machine learning (funnel) and ontology (cogwheel), enabling result prediction. Through the integration of multiple modules, these systems can capitalize on the advantages of learning-based approaches, such as machine learning, while leveraging symbolic reasoning techniques, such as ontologies.

Similar to the \emph{Expert System Embedded Learning} category, hybrid applications leverage both learning and deductive reasoning. They uses machine learning for tasks such as data recognition, shape analysis, or event detection and subsequently employ deductive reasoning based on the information acquired earlier.
These works go beyond populating an ontology, even though the initial process may be similar. After processing raw data through learning, they incorporate it into the ABox of the ontology, enabling the utilization of deductive reasoning or (at least) semantic processing.
The inclusion of some articles in this category is also justified by the incorporation of at least two distinct forms of hybridization. In other words, the authors not only engage in ontology learning, but also leverage the ontology created to inform a machine learning algorithm.

Some papers are based on the semantic interpretation of images. In the study by~\cite{zarchi_semantic_2014}, the authors focused on semantic image recognition based on image properties. They used artificial neural networks (ANNs) to develop a model capable of recognizing similar images by taking advantage of their semantic properties.
\cite{ye_eventnet_2015} took computer vision applications a step further, focusing on event recognition. They used deep learning to train a model capable of recognizing events from videos, aligning them with ontology concepts. Convolutional neural networks (CNNs) played a crucial role in recognizing multiple events within a single video.
\cite{donadello_integration_2016} addressed the semantic understanding of images through the integration of ontology and machine learning. Their Part Whole Clustering Algorithm (PWCA) used ontology and machine learning, including deep learning and clustering techniques, to identify the different parts of an image belonging to specific objects.
In the context of fine image classification,~\cite{palazzo_exploiting_2021}  processed a complex dataset containing images of various fruit varieties. They tackled the challenge using Bayesian networks and CNNs. The authors encoded specialized knowledge in deep models by building a Bayesian network graph from the domain ontology. 

Two studies investigate messages related to aviation.
In the first paper,~\cite{wang_knowledge_2010} focused on the analysis of aviation-related messages and failure analysis. They used a Back-Propagation Neural Network (BPNN) algorithm, a type of artificial neural network (ANN). The ontology played a crucial role in preparing the variables for the neural network, subsequently facilitating the extraction of knowledge and its representation as rules.
In the second paper,~\cite{wang_atc_2021}  studied natural language processing (NLP) in the context of aviation, in particular Air Traffic Control (ATC). They used a long-term memory network (LSTM), a type of recurrent neural network (RNN), and integrated an ontology to facilitate the translation of aeronautical messages. This approach enabled efficient language processing and understanding in the aviation context.

Two papers can be identified as being in the Smart City domain.~\cite{keyarsalan_designing_2011} focus on fuzzy ontology for traffic light control in a smart city context. They use the radial basis function neural network (RBFNN) for image recognition tasks, such as traffic density estimation. Then, the ontology is used as a decision aid to regulate traffic based on the results returned in real time by the images. ~\citep{patel_video_2021} deal with video surveillance in a smart city scenario, in particular for the detection of abnormal events in a parking zone. They use the You Only Look Once (YOLO) algorithm for object detection in video images, enabling efficient feature extraction. The detected objects are then processed by an ontology to perform semantic reasoning on the images and identify anomalies.

{\tiny 
\begin{longtable}{@{}p{0.11\textwidth}p{0.1\textwidth}p{0.12\textwidth}p{0.11\textwidth}p{0.12\textwidth}cc@{}}
\caption{Details of articles in the Learning and Reasoning Systems category}
\label{tab:reason_after_learning}\\

        \toprule
          Category               					& AI Theme        	& Application domain		& Learning type             	& ML algorithm 							& Reasoning			& Paper    \\ \midrule   \endhead
	Expert System Embedded Learning		& -				& Healthcare			& Supervised			& Adaboost							&  \tikzcmark			& \citep{khan_validation_2013}  \\  
          								& -				& Smart City			& 					& MLR, k-NN, RF						&  \tikzcmark			& \citep{bischof_enriching_2018}  \\\cmidrule(r){4-7}
									& -				& Smart City			& Unsupervised		& PCA								&  \tikzcmark			& \citep{bischof_enriching_2018}  \\ \cmidrule(r){1-7}
	Hybrid application					& MAS			& -					& Supervised			& ANN								&  \tikzcmark  			& \citep{rosaci_cilios_2007}  \\ 
         								& NLP, knowledge acquisition& Aviation, Failure analysis& 			& BPNN 								&  \tikzxmark			& \citep{wang_knowledge_2010} \\
									& Computer Vision, Fuzzy ontology	& Smart City, Traffic Light Control& 	& RBFNN								&  \tikzcmark			& \citep{keyarsalan_designing_2011}  \\ 
         								& Computer Vision, Activity recognition  & -    	& 			      		& SVM       							&  \tikzcmark			& \citep{del_rincon_common-sense_2013} \\
									& Computer Vision	& -					& 					& ANN								& \tikzxmark			& \citep{zarchi_semantic_2014} \\ 
									& Computer Vision	& -					& 					& CNN								& \tikzxmark			& \citep{ye_eventnet_2015}  \\
									& Computer Vision	& - 					& 					& R-CNN								& \tikzcmark			& \citep{donadello_integration_2016}  \\
 									& Time Series		& Smart Grids 			& 					& Shapelet							&  \tikzcmark			& \citep{patri_sensors_2016}  \\
									& MAS			& Web				& 					& RF									&  \tikzcmark 			& \citep{mitchell_never-ending_2018}  \\ 
									& -				& Industry	(alarm system)	& 					& Bayesian network						&  \tikzcmark 			& \citep{silva_context-aware_2018}\\  
									& NLP			& Technology			& 					& Rocchio algorithm						& \tikzxmark			& \citep{shi_ontology-based_2019}  \\
									& NLP			& Finance				& 					& LSTM								& \tikzxmark  			& \citep{zhang_hybrid_2019}  \\
									& Signal analysis 	& Industry				& 					& HMM								&  \tikzcmark			& \citep{zhou_hybrid_2019}  \\  
									& -				& Healthcare			& 					& LogitBoost, ANN						&  \tikzcmark			& \citep{ali_smart_2020}  \\ 
									& Indoor localization & Healthcare, Smart environment  &       			& Decision tree (C5.0)	     				&  \tikzcmark 			&  \citep{woensel_indoor_2020} \\ 
									& NLP			& Transport			& 					& MLP, GRU							&  \tikzcmark			& \citep{cheng_location_2021}  \\
									& Computer Vision	& - 					& 					& R-CNN								& \tikzcmark			& \citep{foo_screw_2021}  \\
									& Computer Vision (fine-grained classification)	& - 	& 				& Bayesian networks	, CNN				& \tikzxmark  			& \citep{palazzo_exploiting_2021}  \\ 
									& NLP			& Aviation				& 					& LSTM								& \tikzxmark			& \citep{wang_atc_2021}  \\ 
									& AmI 			& Healthcare        		&         		   		& ANN 								&  \tikzcmark     		& \citep{chung_ambient_2020}   \\
									& Computer Vision	& Smart City (video surveillance) & 				& YOLO								& \tikzcmark 			& \citep{patel_video_2021}  \\ 
									& Fraud Detection	& Insurance       		&         		   		& Ensemble methods, XGBoost	 		&  \tikzxmark     		& \citep{zhang_auto_2021}   \\ \cmidrule(r){4-7}
									& Computer Vision	& -					& Unsupervised		& k-means							&  \tikzcmark			& \citep{del_rincon_common-sense_2013}  \\
									& Computer Vision	& -					& 					& PWCA								&  \tikzcmark			& \citep{donadello_integration_2016}  \\ 
									& MAS			& Web				& 					& k-means							&  \tikzcmark 			& \citep{mitchell_never-ending_2018}  \\
									& AmI 			& Healthcare        		&         		   		& APRIORI							&  \tikzcmark     		& \citep{chung_ambient_2020}   \\
									& Indoor localization & Healthcare, Smart environment  &       			& PCA     								&  \tikzcmark 			&  \citep{woensel_indoor_2020} \\ 
									& Human-robot interaction & Robotics		& 				 	& SOM								& \tikzxmark 			& \citep{russo_unsupervised_2021} \\ \cmidrule(r){4-7} 
									& NLP			& Finance				&Self-supervised 		& Word2Vec							& \tikzxmark  			& \citep{zhang_hybrid_2019}  \\ 
									& NLP			& Transport			& 					& Word2Vec							&  \tikzcmark			& \citep{cheng_location_2021}  \\
     
        \bottomrule
\end{longtable}
}

\section{Conclusion: challenges and research directions}
\label{conclusion}

\subsection{Hybrid AI: Design Pattern and Taxonomy}
As demonstrated in this study, the integration of ontology and machine learning represents a significant challenge that also brings forth new possibilities. It is important to recognize that the fusion of ontology and machine learning falls within a broader paradigm called AI hybridization, which aims to combine different types of reasoning. \cite{van_bekkum_modular_2021} has described several design patterns for hybrid AI, consisting of seven elementary patterns that characterize the types of input and output data, as well as the mechanisms employed for data processing (prediction, deduction, training, etc.). These elementary patterns are further combined to form more intricate design patterns that delineate various hybridization scenarios. In order to facilitate the utilization of this classification, we have associated each category from this SLR with the corresponding design pattern. The outcomes of this mapping are presented in table~\ref{tab:kautz_classification}.

Notably, design pattern number 4 emerges as the most suitable for describing \emph{Ontology Learning} and \emph{Ontology Mapping}. Design pattern number 4 is specifically utilized for learning with symbolic output and consists of a primary block that learns from textual data and a secondary block capable of deducing insights from a new semantic model. 

Design pattern number 5 is dedicated to the specific objective of mitigating the widely recognized 'black-box" phenomenon inherent in some machine learning algorithms, especially deep neural networks. Within this design pattern, a symbolic model is employed following the training of a learning model to elucidate the obtained results by leveraging prior knowledge. This aligns closely with our designated category of \emph{Ontology-based explanation} where the emphasis lies on utilizing ontological resources to provide comprehensible explanations.

Design pattern number 3b represents systems capable of learning not only from data, but also from symbols, as is the case for \emph{Ontology-based feature engineering}. 

Design pattern number 7 specifically addresses informed learning with prior knowledge, aligning with our designated categories of \emph{Ontology-based algorithm design} and \emph{Ontology-based algorithm training}. The core principle underlying this design pattern is the inclusion of prior knowledge within the pipeline of the machine learning model. By incorporating relevant awareness, the objective is to enhance the performance and generalization capabilities of the model. As observed in our SLR, the integration of knowledge can occur at various stages within the learning pipeline. These include incorporating knowledge into the training data, incorporating it into the model architecture, incorporating it during the learning process of the model, and even incorporating it post-hoc after the learning phase. 

Design pattern number 10 is dedicated to harnessing the power of machine learning, specifically neural networks, to enable logical reasoning. In this design pattern, a neural network is trained to perform logical reasoning tasks, aligning closely with our designated category of \emph{Learning-based Reasoning} This approach offers notable advantages, including enhanced scalability compared to traditional logical reasoning methods that may encounter bottlenecks when dealing with large ontologies. Moreover, learning-based reasoning exhibits greater resilience to noisy or missing data, thus improving the overall robustness of the reasoning process. 

Design pattern number 12 focuses on the design of hybrid AI systems that closely resemble real-life applications. In contrast to a single monolithic component, hybrid AI systems are composed of multiple interconnected modules that communicate with each other. This design pattern aligns with the \emph{Learning and Reasoning system} category identified in our systematic literature review. The objective of these hybrid AI systems is to leverage the synergies between learning and symbolic modules, aiming to produce more reliable models with enhanced transparency and reproducibility. By integrating multiple modules, these systems can benefit from the strengths of both learning-based approaches, such as machine learning, and symbolic reasoning techniques.

Another well-recognized subgroup within Hybrid AI is neuro-symbolic, which concentrates on the integration of symbolic methods with neural networks, especially deep neural networks~\cite{henry_kautz_third_2020} has introduced a comprehensive taxonomy that classifies the diverse neuro-symbolic approaches, providing a structured framework for understanding and categorizing them.
Indeed, in a similar manner, the categories identified in our SLR align with the groups outlined in~\cite{henry_kautz_third_2020}'s taxonomy, specifically when the ML algorithm employed is a neural network. To further aid readers in their mapping efforts, we have included this alignment in table~\ref{tab:kautz_classification}, allowing for a clearer understanding of the correspondence between the SLR categories and Kautz's taxonomy.

Category \emph{Ontology-based feature engineering} corresponds to approach Symbolic Neuro symbolic, which involves transforming raw data using symbolic integration. This technique is commonly used in NLP tasks, where data is converted into vectors using methods such as Word2vec and GloVe. 
 
The Neuro\_\{Symbolic\}  architecture, on the other hand, is more complex: it facilitates the conversion of symbolic rules into neural network models (\emph{Ontology-based algorithm design} or \emph{Ontology-based algorithm design}), as illustrated by the logical tensor networks~\citep{serafini_learning_2017} discussed in this systematic literature review.

The Symbolic[Neuro] architecture combines neural pattern recognition with a symbolic problem-solving framework, resulting in enhanced problem-solving capabilities. This architecture is specifically applied in the category of~\emph{Expert System Embedded Learning}. 

The Neuro\textbar Symbolic architecture is prominently featured in our study, encompassing the categories of \emph{ontology learning}, \emph{ontology mapping}, \emph{ontology-based explanation}, and \emph{hybrid application}. This architecture closely resembles the Symbolic[Neuro] architecture but utilizes coroutines instead of subroutines. It emphasizes the communication between a symbolic system and a neural system, which is particularly relevant to our \emph{hybrid application} category. We have chosen to place the other three categories within this architecture because it best aligns with their respective functionalities, even though the communication between the two systems may be more limited compared to hybrid applications.

Lastly, the \emph{Learning-based reasoning} category is analogous to the Neuro:Symbolic→Neuro architecture, which involves training a neural network on symbolic rules. In learning-based reasoning, the network learns logical rules to perform deductive reasoning on new inputs. The advantage of this approach is that the neural network does not perform reasoning by explicitly following step-by-step rules; instead, it makes predictions based on the expected outcome of deductive reasoning. As mentioned earlier, this approach significantly reduces computational time, particularly when dealing with large ontologies.

\begin{table}[ht]
 \renewcommand{\arraystretch}{1.5}
    \caption{Alignment of our Hybridization Categories with Van Bekkum's Design Patterns and Kautz's Taxonomy}\label{tab:kautz_classification}
    \begin{center}
    \begin{minipage}{\textwidth}
    \begin{tabular}{|p{0.4\linewidth}|p{0.2\linewidth}|p{0.3\linewidth}|}
     \hline
     \textbf{Category} & \textbf{\cite{van_bekkum_modular_2021}} & \textbf{\cite{henry_kautz_third_2020}}\\
    \hline
    \multicolumn{3}{|c|}{Learning-enhanced ontology} \\
    \hline
       Ontology Learning & Design Pattern 4 & Neuro\textbar Symbolic \\
       Ontology Mapping & Design Pattern 4 & Neuro\textbar Symbolic\\
       Learning-based reasoning & Design Pattern 10 & Neuro:Symbolic→Neuro\\
    \hline
    \multicolumn{3}{|c|}{Semantic data mining} \\
    \hline
      Ontology-based feature engineering & Design Pattern 3b & Symbolic Neuro symbolic\\
      Ontology-based algorithm design & Design Pattern 7 & Neuro\_\{Symbolic\}\\
      Ontology-based algorithm training & Design Pattern 7 & Neuro\_\{Symbolic\}\\
      Ontology-based explanation & Design Pattern 5 & Neuro\textbar Symbolic\\
    \hline
    \multicolumn{3}{|c|}{Learning and Reasoning system} \\
    \hline
       Expert System Embedded Learning & Design Pattern 12 & Symbolic[Neuro]\\
       Hybrid application & Design Pattern 12 & Neuro\textbar Symbolic\\\hline
    \end{tabular}
    \end{minipage}
    \end{center}
    
\end{table}

\subsection{Three challenges for the future}

We have identified three main challenges using ontologies combined with machine learning.
The first is the formal proof of the expressiveness and decidability of the ontology. The second is the ability to explain the results of a machine learning algorithm. The third concerns the management of consistency during \emph{ontology learning} and \emph{ontology mapping}.

\subsubsection{Improve expressiveness and decidability of the ontology}
Between taxonomy and formal ontology, this semantic representation of knowledge is a balance between expressiveness and decidability.
Description logics are used to formalize ontology and determine this level of expressiveness/decidability.
Each description logic represents a formal, axiomatized language describing the level of constraints supported. 
Since 2012, OWL2 language, recommended by the W3C\footnote{https://www.w3.org}, allows the expressiveness of $\mathcal{SROIQ(D)}$ logic.
Inference engines can interpret this logic and check consistency, reorganize the structuration of concepts in the TBox, or, thanks to rule-based language (e.g. SWRL) infer new knowledge into the ABox.
A large majority of articles studied make no mention of deductive reasoning beyond subsumption links made possible by ontologies and inference engines. Many of them use an ontology for its contribution at the semantic level. Ontology is used as an improved taxonomy since it has the advantage of also representing non-hierarchical relationships between the different terms of a domain. This corresponds, at best, to $\mathcal{S}$ description logic language.

\subsubsection{eXplainable Artificial Intelligence (XAI)}

In recent times, AI systems have made significant progress in perceiving, learning, decision-making, and even autonomous action. Nevertheless, there remains a level of distrust among humans towards these systems, largely due to their inability to provide explanations for the reasoning behind their decisions~\citep{gunning_darpas_2019}. 

Over the past two decades, the research domain of eXplainable Artificial Intelligence (XAI) has experienced significant growth. This attribute holds critical importance in sensitive industrial sectors such as healthcare, finance, insurance, and defense. Achieving explainability in AI systems has been explored through various techniques, including Local Interpretable Model-agnostic Explanations (LIME) \citep{ribeiro_why_2016}, SHapley Additive exPlanations (SHAP) \citep{lundberg_unified_2017}, as well as symbolic reasoning.
Two papers, outlined in section \ref{reason_after_learning}, have specifically explored this subject by incorporating ontology \citep{confalonieri_using_2021, panigutti_doctor_2020}. These papers propose interesting approaches for obtaining either a global \citep{confalonieri_using_2021} or local explanation\citep{panigutti_doctor_2020}. In these works, a domain-specific ontology pertaining to the targeted field of explanation is employed to enhance the explanatory quality.
These studies highlight the role of ontologies in augmenting the interpretability of AI models, both at the global and local levels. By integrating prior knowledge through ontologies, a logical and cohesive framework is established for the model's explanations, ensuring their alignment with existing domain knowledge. This integration not only enhances the credibility of the explanations but also facilitates a deeper comprehension of the model's reasoning process.

\subsubsection{Consistency checking}
Change management in ontology during \emph{ontology learning} process or \emph{ontology mapping} requires consistency checking. Consistency management allows guaranteeing the reasoning validity of the different ontology releases.
This study of consistency is well carried out by \cite{mitchell_never-ending_2018} which is interested in the enrichment of an ontology, as well as by \cite{del_rincon_common-sense_2013} and \cite{donadello_integration_2016}, classified in the category \emph{ontology population} (as explained in paragraph \ref{ontology_learning}). However, these three works represent 6\% of the papers which should be concerned by the consistency management. Furthermore, we did not find any paper mentioning the study of consistency in the other categories present in this literature review.

\subsection{Conclusion}

We conduct a SLR to explore the integration of inductive reasoning and deductive reasoning in systems that combine machine learning and ontologies. The aim of this study was to determine whether hybrid AI techniques could improve the ability of machines to perceive the complexity and nuances of our real world in order to improve their interactions with it. In our SLR, we reviewed a total of 128 papers that explore the combination of machine learning and ontologies across various domains and with diverse objectives. Through this comprehensive analysis, we identified and highlighted different categories of combinations of machine learning and ontologies that address distinct problems. We also provide a comprehensive examination of all these categories, emphasizing the various machine learning algorithms utilized in the studies.
Additionally, we have established the connections between our categorization and~\cite{van_bekkum_modular_2021}'s design pattern as well as~\cite{henry_kautz_third_2020}'s neuro-symbolic classification to provide insights to interested readers.

How can we enable machines to have an impact on the tangible world if they can’t make sense of the various aspects, complexities, and nuances of the real world? The question remains, but according to this SLR, from Plato to today, the work is still in progress.

\bibliographystyle{unsrtnat}
\bibliography{bibliography}

\begin{thebibliography}{169}
\providecommand{\natexlab}[1]{#1}
\providecommand{\url}[1]{\texttt{#1}}
\expandafter\ifx\csname urlstyle\endcsname\relax
  \providecommand{\doi}[1]{doi: #1}\else
  \providecommand{\doi}{doi: \begingroup \urlstyle{rm}\Url}\fi

\bibitem[Nilsson(2009)]{nilsson_quest_2009}
Nils~J. Nilsson.
\newblock \emph{The Quest for Artificial Intelligence}.
\newblock Cambridge University Press, 1 edition, 2009.
\newblock ISBN 978-0-521-11639-8 978-0-521-12293-1 978-0-511-81934-6.
\newblock \doi{10.1017/CBO9780511819346}.

\bibitem[Wang(2019)]{wang_defining_2019}
Pei Wang.
\newblock On {Defining} {Artificial} {Intelligence}.
\newblock \emph{Journal of Artificial General Intelligence}, 10:\penalty0
  1--37, January 2019.
\newblock ISSN 1946-0163.
\newblock \doi{10.2478/jagi-2019-0002}.

\bibitem[Minsky(1986)]{minsky_society_1986}
Marvin Minsky.
\newblock \emph{The {Society} of {Mind}}.
\newblock Simon and Schuster, New York etc., January 1986.
\newblock ISBN 978-0-671-65713-0.

\bibitem[Dobrev(2012)]{dobrev_definition_2012}
Dimiter Dobrev.
\newblock A definition of artificial intelligence, 2012.

\bibitem[{McCarthy} et~al.(1955){McCarthy}, Minsky, Rochester, Corporation, and
  Shannon]{mccarthy_proposal_1955}
J~{McCarthy}, M~L Minsky, N~Rochester, I~B~M Corporation, and C~E Shannon.
\newblock A {PROPOSAL} {FOR} {THE} {DARTMOUTH} {SUMMER} {RESEARCH} {PROJECT}
  {ON} {ARTIFICIAL} {INTELLIGENCE}.
\newblock \emph{AI Magazine}, 1955.

\bibitem[Kaplan and Haenlein(2019)]{kaplan_siri_2019}
Andreas Kaplan and Michael Haenlein.
\newblock Siri, {Siri}, in my hand: {Who}’s the fairest in the land? {On} the
  interpretations, illustrations, and implications of artificial intelligence.
\newblock \emph{Business Horizons}, 62\penalty0 (1):\penalty0 15--25, February
  2019.
\newblock ISSN 0007-6813.
\newblock \doi{10.1016/j.bushor.2018.08.004}.

\bibitem[Russell and Norvig(2009)]{russell_artificial_2009}
Stuart Russell and Peter Norvig.
\newblock \emph{Artificial Intelligence: A Modern Approach}.
\newblock Pearson, 3rd edition edition, 2009.
\newblock ISBN 978-0-13-604259-4.

\bibitem[Plato(1888)]{plato_republic_1888}
Plato.
\newblock \emph{The {Republic} of {Plato}}.
\newblock Oxford University Press, 1888.

\bibitem[Rafanelli et~al.(2022)Rafanelli, Costantini, and
  Omicini]{rafanelli_position_2022}
Andrea Rafanelli, Stefania Costantini, and Andrea Omicini.
\newblock Position paper: {On} the role of abductive reasoning in semantic
  image segmentation.
\newblock 2022.

\bibitem[Roudaut(2017)]{roudaut_comment_2017}
Frédéric Roudaut.
\newblock Comment on invente les hypothèses : {Peirce} et la théorie de
  l'abduction.
\newblock \emph{Cahiers philosophiques}, 150\penalty0 (3):\penalty0 45--65,
  December 2017.
\newblock ISSN 0241-2799.
\newblock 00002 Bibliographie\_available: 0 Cairndomain: www.cairn.info Cite
  Par\_available: 0 Publisher: Vrin.

\bibitem[Smith(2020)]{smith_aristotles_2020}
Robin Smith.
\newblock Aristotle’s {Logic}.
\newblock In Edward~N. Zalta, editor, \emph{The {Stanford} {Encyclopedia} of
  {Philosophy}}. Metaphysics Research Lab, Stanford University, fall 2020
  edition, 2020.

\bibitem[Haugeland(1989)]{haugeland_artificial_1989}
John Haugeland.
\newblock \emph{Artificial {Intelligence}: {The} {Very} {Idea}}.
\newblock 1989.
\newblock \doi{10.7551/mitpress/1170.001.0001}.

\bibitem[Guarino et~al.(2009)Guarino, Oberle, and Staab]{guarino_what_2009}
Nicola Guarino, Daniel Oberle, and Steffen Staab.
\newblock What {Is} an {Ontology}?
\newblock In Steffen Staab and Rudi Studer, editors, \emph{Handbook on
  {Ontologies}}, International {Handbooks} on {Information} {Systems}, pages
  1--17. Springer, Berlin, Heidelberg, 2009.
\newblock ISBN 978-3-540-92673-3.
\newblock \doi{10.1007/978-3-540-92673-3_0}.

\bibitem[Hitzler and Sarker(2022)]{hitzler_neuro-symbolic_2022}
P.~Hitzler and M.~K. Sarker.
\newblock \emph{Neuro-{Symbolic} {Artificial} {Intelligence}: {The} {State} of
  the {Art}}.
\newblock IOS Press, 2022.
\newblock ISBN 978-1-64368-245-7.

\bibitem[Garcez and Lamb(2020)]{garcez_neurosymbolic_2020}
Artur~d'Avila Garcez and Luis~C. Lamb.
\newblock Neurosymbolic {AI}: {The} 3rd {Wave}.
\newblock \emph{arXiv:2012.05876 [cs]}, December 2020.

\bibitem[{Henry Kautz}(2020)]{henry_kautz_third_2020}
{Henry Kautz}.
\newblock The {Third} {AI} {Summer}, {Henry} {Kautz}, {AAAI} 2020 {Robert} {S}.
  {Engelmore} {Memorial} {Award} {Lecture}, February 2020.

\bibitem[van Bekkum et~al.(2021)van Bekkum, de~Boer, van Harmelen,
  Meyer-Vitali, and Teije]{van_bekkum_modular_2021}
Michael van Bekkum, Maaike de~Boer, Frank van Harmelen, André Meyer-Vitali,
  and Annette~ten Teije.
\newblock Modular {Design} {Patterns} for {Hybrid} {Learning} and {Reasoning}
  {Systems}: a taxonomy, patterns and use cases.
\newblock \emph{arXiv:2102.11965 [cs]}, March 2021.

\bibitem[Kitchenham et~al.(2010)Kitchenham, Pretorius, Budgen, Pearl~Brereton,
  Turner, Niazi, and Linkman]{kitchenham_systematic_2010}
Barbara Kitchenham, Rialette Pretorius, David Budgen, O.~Pearl~Brereton, Mark
  Turner, Mahmood Niazi, and Stephen Linkman.
\newblock Systematic literature reviews in software engineering – {A}
  tertiary study.
\newblock \emph{Information and Software Technology}, 52\penalty0 (8):\penalty0
  792--805, August 2010.
\newblock ISSN 0950-5849.
\newblock \doi{10.1016/j.infsof.2010.03.006}.

\bibitem[Kitchenham and Charters(2007)]{kitchenham_guidelines_2007}
Barbara Kitchenham and Stuart Charters.
\newblock Guidelines for performing systematic literature reviews in software
  engineering.
\newblock Technical report, 01 2007.

\bibitem[Al-Aswadi et~al.(2020)Al-Aswadi, Chan, and
  Gan]{al-aswadi_automatic_2020}
Fatima~N. Al-Aswadi, Huah~Yong Chan, and Keng~Hoon Gan.
\newblock Automatic ontology construction from text: a review from shallow to
  deep learning trend.
\newblock \emph{Artificial Intelligence Review}, 53\penalty0 (6):\penalty0
  3901--3928, August 2020.
\newblock ISSN 1573-7462.
\newblock \doi{10.1007/s10462-019-09782-9}.

\bibitem[Khadir et~al.(2021)Khadir, Aliane, and Guessoum]{khadir_ontology_2021}
Ahlem~Chérifa Khadir, Hassina Aliane, and Ahmed Guessoum.
\newblock Ontology learning: {Grand} tour and challenges.
\newblock \emph{Computer Science Review}, 39:\penalty0 100339, 2021.
\newblock ISSN 1574-0137.
\newblock \doi{10.1016/j.cosrev.2020.100339}.

\bibitem[Dou et~al.(2015)Dou, Wang, and Liu]{dou_semantic_2015}
Dejing Dou, Hao Wang, and Haishan Liu.
\newblock Semantic data mining: {A} survey of ontology-based approaches.
\newblock In \emph{Proceedings of the 2015 {IEEE} 9th {International}
  {Conference} on {Semantic} {Computing} ({IEEE} {ICSC} 2015)}, pages 244--251,
  February 2015.
\newblock \doi{10.1109/ICOSC.2015.7050814}.

\bibitem[Sirichanya and Kraisak(2021)]{sirichanya_semantic_2021}
Chanmee Sirichanya and Kesorn Kraisak.
\newblock Semantic data mining in the information age: {A} systematic review.
\newblock \emph{International Journal of Intelligent Systems}, 36\penalty0
  (8):\penalty0 3880--3916, 2021.
\newblock ISSN 1098-111X.
\newblock \doi{10.1002/int.22443}.
\newblock 00005 \_eprint:
  https://onlinelibrary.wiley.com/doi/pdf/10.1002/int.22443.

\bibitem[Gusenbauer and Haddaway(2020)]{gusenbauer_which_2020}
Michael Gusenbauer and Neal~R. Haddaway.
\newblock Which academic search systems are suitable for systematic reviews or
  meta-analyses? {Evaluating} retrieval qualities of {Google} {Scholar},
  {PubMed}, and 26 other resources.
\newblock \emph{Research Synthesis Methods}, 11\penalty0 (2):\penalty0
  181--217, 2020.
\newblock ISSN 1759-2887.
\newblock \doi{https://doi.org/10.1002/jrsm.1378}.
\newblock 00102 \_eprint:
  https://onlinelibrary.wiley.com/doi/pdf/10.1002/jrsm.1378.

\bibitem[Usman et~al.(2021)Usman, Ali, and Wohlin]{usman_quality_2021}
Muhammad Usman, Nauman~bin Ali, and Claes Wohlin.
\newblock A {Quality} {Assessment} {Instrument} for {Systematic} {Literature}
  {Reviews} in {Software} {Engineering}.
\newblock \emph{arXiv:2109.10134 [cs]}, September 2021.

\bibitem[von Rueden et~al.(2021)von Rueden, Mayer, Beckh, Georgiev,
  Giesselbach, Heese, Kirsch, Pfrommer, Pick, Ramamurthy, Walczak, Garcke,
  Bauckhage, and Schuecker]{von_rueden_informed_2021}
Laura von Rueden, Sebastian Mayer, Katharina Beckh, Bogdan Georgiev, Sven
  Giesselbach, Raoul Heese, Birgit Kirsch, Julius Pfrommer, Annika Pick,
  Rajkumar Ramamurthy, Michal Walczak, Jochen Garcke, Christian Bauckhage, and
  Jannis Schuecker.
\newblock Informed {Machine} {Learning} -- {A} {Taxonomy} and {Survey} of
  {Integrating} {Knowledge} into {Learning} {Systems}.
\newblock \emph{IEEE Transactions on Knowledge and Data Engineering}, pages
  1--1, 2021.
\newblock ISSN 1041-4347, 1558-2191, 2326-3865.
\newblock \doi{10.1109/TKDE.2021.3079836}.

\bibitem[Rubin et~al.(2008)Rubin, Shah, and Noy]{rubin_biomedical_2008}
Daniel~L. Rubin, Nigam~H. Shah, and Natalya~F. Noy.
\newblock Biomedical ontologies: a functional perspective.
\newblock \emph{Briefings in Bioinformatics}, 9\penalty0 (1):\penalty0 75--90,
  January 2008.
\newblock ISSN 1467-5463.
\newblock \doi{10.1093/bib/bbm059}.

\bibitem[Wong et~al.(2012)Wong, Liu, and Bennamoun]{wong_ontology_2012}
Wilson Wong, Wei Liu, and Mohammed Bennamoun.
\newblock Ontology learning from text: {A} look back and into the future.
\newblock \emph{ACM Computing Surveys}, 44\penalty0 (4):\penalty0 20:1--20:36,
  September 2012.
\newblock ISSN 0360-0300.
\newblock \doi{10.1145/2333112.2333115}.

\bibitem[Maedche and Staab(2001)]{maedche_ontology_2001}
A.~Maedche and S.~Staab.
\newblock Ontology learning for the {Semantic} {Web}.
\newblock \emph{IEEE Intelligent Systems}, 16\penalty0 (2):\penalty0 72--79,
  March 2001.
\newblock ISSN 1941-1294.
\newblock \doi{10.1109/5254.920602}.
\newblock 03204 Conference Name: IEEE Intelligent Systems.

\bibitem[Asim et~al.(2018)Asim, Wasim, Khan, Mahmood, and
  Abbasi]{asim_survey_2018}
Muhammad~Nabeel Asim, Muhammad Wasim, Muhammad Usman~Ghani Khan, Waqar Mahmood,
  and Hafiza~Mahnoor Abbasi.
\newblock A survey of ontology learning techniques and applications.
\newblock \emph{Database}, 2018\penalty0 (bay101), January 2018.
\newblock ISSN 1758-0463.
\newblock \doi{10.1093/database/bay101}.

\bibitem[Getahun and Woldemariyam(2017)]{getahun_integrated_2017}
Fekade Getahun and Kidane Woldemariyam.
\newblock Integrated {Ontology} {Learner}: {Towards} {Generic} {Semantic}
  {Annotation} {Framework}.
\newblock In \emph{Proceedings of the 9th {International} {Conference} on
  {Management} of {Digital} {EcoSystems}}, {MEDES} '17, pages 142--149, New
  York, NY, USA, 2017. Association for Computing Machinery.
\newblock ISBN 978-1-4503-4895-9.
\newblock \doi{10.1145/3167020.3167042}.

\bibitem[Ghoniem et~al.(2019)Ghoniem, Alhelwa, and Shaalan]{ghoniem_novel_2019}
Rania~M. Ghoniem, Nawal Alhelwa, and Khaled Shaalan.
\newblock A {Novel} {Hybrid} {Genetic}-{Whale} {Optimization} {Model} for
  {Ontology} {Learning} from {Arabic} {Text}.
\newblock \emph{ALGORITHMS}, 12\penalty0 (9), September 2019.
\newblock \doi{10.3390/a12090182}.
\newblock 00000 Place: ST ALBAN-ANLAGE 66, CH-4052 BASEL, SWITZERLAND
  Publisher: MDPI Type: Article.

\bibitem[Rani et~al.(2017)Rani, Dhar, and Vyas]{rani_semi-automatic_2017}
Monika Rani, Amit~Kumar Dhar, and O.~P. Vyas.
\newblock Semi-automatic terminology ontology learning based on topic modeling.
\newblock \emph{ENGINEERING APPLICATIONS OF ARTIFICIAL INTELLIGENCE},
  63:\penalty0 108--125, August 2017.
\newblock ISSN 0952-1976.
\newblock \doi{10.1016/j.engappai.2017.05.006}.
\newblock 00000 Place: THE BOULEVARD, LANGFORD LANE, KIDLINGTON, OXFORD OX5
  1GB, ENGLAND Publisher: PERGAMON-ELSEVIER SCIENCE LTD Type: Article.

\bibitem[Albukhitan et~al.(2017)Albukhitan, Helmy, and
  Alnazer]{albukhitan_arabic_2017}
Saeed Albukhitan, Tarek Helmy, and Ahmed Alnazer.
\newblock Arabic {Ontology} {Learning} {Using} {Deep} {Learning}.
\newblock In \emph{Proceedings of the {International} {Conference} on {Web}
  {Intelligence}}, {WI} '17, pages 1138--1142, New York, NY, USA, 2017.
  Association for Computing Machinery.
\newblock ISBN 978-1-4503-4951-2.
\newblock \doi{10.1145/3106426.3109052}.

\bibitem[Jurkevičius and Vasilecas(2010)]{jurkevicius_ontology_2010}
Darius Jurkevičius and Olegas Vasilecas.
\newblock Ontology {Creation} {Using} {Formal} {Concepts} {Approach}.
\newblock In \emph{Proceedings of the 11th {International} {Conference} on
  {Computer} {Systems} and {Technologies} and {Workshop} for {PhD} {Students}
  in {Computing} on {International} {Conference} on {Computer} {Systems} and
  {Technologies}}, {CompSysTech} '10, pages 64--70, New York, NY, USA, 2010.
  Association for Computing Machinery.
\newblock ISBN 978-1-4503-0243-2.
\newblock \doi{10.1145/1839379.1839392}.

\bibitem[Petrucci et~al.(2018)Petrucci, Rospocher, and
  Ghidini]{petrucci_expressive_2018}
Giulio Petrucci, Marco Rospocher, and Chiara Ghidini.
\newblock Expressive ontology learning as neural machine translation.
\newblock \emph{JOURNAL OF WEB SEMANTICS}, 52-53:\penalty0 66--82, October
  2018.
\newblock ISSN 1570-8268.
\newblock \doi{10.1016/j.websem.2018.10.002}.
\newblock 00000 Place: PO BOX 211, 1000 AE AMSTERDAM, NETHERLANDS Publisher:
  ELSEVIER SCIENCE BV Type: Article.

\bibitem[Wu and Weld(2008)]{wu_automatically_2008}
Fei Wu and Daniel~S. Weld.
\newblock Automatically {Refining} the {Wikipedia} {Infobox} {Ontology}.
\newblock In \emph{Proceedings of the 17th {International} {Conference} on
  {World} {Wide} {Web}}, {WWW} '08, pages 635--644, New York, NY, USA, 2008.
  Association for Computing Machinery.
\newblock ISBN 978-1-60558-085-2.
\newblock \doi{10.1145/1367497.1367583}.

\bibitem[Song et~al.(2016)Song, Jiang, and Liu]{song_automated_2016}
Bo~Song, Zuhua Jiang, and Lijun Liu.
\newblock Automated experiential engineering knowledge acquisition through
  {Q}\&{A} contextualization and transformation.
\newblock \emph{ADVANCED ENGINEERING INFORMATICS}, 30\penalty0 (3):\penalty0
  467--480, August 2016.
\newblock ISSN 1474-0346.
\newblock \doi{10.1016/j.aei.2016.06.002}.
\newblock 00000 Place: THE BOULEVARD, LANGFORD LANE, KIDLINGTON, OXFORD OX5
  1GB, OXON, ENGLAND Publisher: ELSEVIER SCI LTD Type: Article.

\bibitem[Jia et~al.(2018)Jia, Qi, Shang, Jiang, and Li]{jia_practical_2018}
Yan Jia, Yulu Qi, Huaijun Shang, Rong Jiang, and Aiping Li.
\newblock A {Practical} {Approach} to {Constructing} a {Knowledge} {Graph} for
  {Cybersecurity}.
\newblock \emph{Engineering}, 4\penalty0 (1):\penalty0 53--60, 2018.
\newblock ISSN 2095-8099.
\newblock \doi{https://doi.org/10.1016/j.eng.2018.01.004}.

\bibitem[McGlinn et~al.(2017)McGlinn, Yuce, Wicaksono, Howell, and
  Rezgui]{mcglinn_usability_2017}
Kris McGlinn, Baris Yuce, Hendro Wicaksono, Shaun Howell, and Yacine Rezgui.
\newblock Usability evaluation of a web-based tool for supporting holistic
  building energy management.
\newblock \emph{Automation in Construction}, 84:\penalty0 154--165, 2017.
\newblock ISSN 0926-5805.
\newblock \doi{https://doi.org/10.1016/j.autcon.2017.08.033}.

\bibitem[Ko et~al.(2021)Ko, Witherell, Lu, Kim, and Rosen]{ko_machine_2021}
Hyunwoong Ko, Paul Witherell, Yan Lu, Samyeon Kim, and David~W. Rosen.
\newblock Machine learning and knowledge graph based design rule construction
  for additive manufacturing.
\newblock \emph{Additive Manufacturing}, 37:\penalty0 101620, 2021.
\newblock ISSN 2214-8604.
\newblock \doi{https://doi.org/10.1016/j.addma.2020.101620}.

\bibitem[Yang(2020)]{yang_construction_2020}
Bo~Yang.
\newblock Construction of logistics financial security risk ontology model
  based on risk association and machine learning.
\newblock \emph{Safety Science}, 123:\penalty0 104437, 2020.
\newblock ISSN 0925-7535.
\newblock \doi{https://doi.org/10.1016/j.ssci.2019.08.005}.

\bibitem[Craven et~al.(2000)Craven, DiPasquo, Freitag, McCallum, Mitchell,
  Nigam, and Slattery]{craven_learning_2000}
Mark Craven, Dan DiPasquo, Dayne Freitag, Andrew McCallum, Tom Mitchell, Kamal
  Nigam, and Seán Slattery.
\newblock Learning to construct knowledge bases from the {World} {Wide} {Web}.
\newblock \emph{Artificial Intelligence}, 118\penalty0 (1):\penalty0 69--113,
  2000.
\newblock ISSN 0004-3702.
\newblock \doi{https://doi.org/10.1016/S0004-3702(00)00004-7}.

\bibitem[Kordjamshidi and Moens(2015)]{kordjamshidi_global_2015}
Parisa Kordjamshidi and Marie-Francine Moens.
\newblock Global machine learning for spatial ontology population.
\newblock \emph{JOURNAL OF WEB SEMANTICS}, 30:\penalty0 3--21, January 2015.
\newblock ISSN 1570-8268.
\newblock \doi{10.1016/j.websem.2014.06.001}.
\newblock 00000 Place: RADARWEG 29, 1043 NX AMSTERDAM, NETHERLANDS Publisher:
  ELSEVIER Type: Article.

\bibitem[Markievicz et~al.(2015)Markievicz, Kapociute-Dzikiene, Tamosiunaite,
  and Vitkute-Adzgauskiene]{markievicz_action_2015}
Irena Markievicz, Jurgita Kapociute-Dzikiene, Minija Tamosiunaite, and Daiva
  Vitkute-Adzgauskiene.
\newblock Action {Classification} in {Action} {Ontology} {Building} {Using}
  {Robot}-{Specific} {Texts}.
\newblock \emph{INFORMATION TECHNOLOGY AND CONTROL}, 44\penalty0 (2):\penalty0
  155--164, 2015.
\newblock ISSN 1392-124X.
\newblock \doi{10.5755/j01.itc.44.2.7322}.
\newblock 00000 Place: KAUNAS UNIV TECHNOL, DEPT ELECTRONICS ENGINEERING,
  STUDENTU STR 50, KAUNAS, LT-51368, LITHUANIA Publisher: KAUNAS UNIV
  TECHNOLOGY Type: Article.

\bibitem[Rubrichi et~al.(2013)Rubrichi, Quaglini, Spengler, Russo, and
  Gallinari]{rubrichi_system_2013}
Stefania Rubrichi, Silvana Quaglini, Alex Spengler, Paola Russo, and Patrick
  Gallinari.
\newblock A system for the extraction and representation of summary of product
  characteristics content.
\newblock \emph{ARTIFICIAL INTELLIGENCE IN MEDICINE}, 57\penalty0 (2,
  SI):\penalty0 145--154, February 2013.
\newblock ISSN 0933-3657.
\newblock \doi{10.1016/j.artmed.2012.08.004}.
\newblock 00000 Place: PO BOX 211, 1000 AE AMSTERDAM, NETHERLANDS Publisher:
  ELSEVIER SCIENCE BV Type: Article.

\bibitem[Packer and Embley(2015)]{packer_cost-effective_2015}
Thomas~L. Packer and David~W. Embley.
\newblock Cost-{Effective} {Information} {Extraction} from {Lists} in {OCRed}
  {Historical} {Documents}.
\newblock In \emph{Proceedings of the 3rd {International} {Workshop} on
  {Historical} {Document} {Imaging} and {Processing}}, {HIP} '15, pages 23--30,
  New York, NY, USA, 2015. Association for Computing Machinery.
\newblock ISBN 978-1-4503-3602-4.
\newblock \doi{10.1145/2809544.2809547}.

\bibitem[Kuang et~al.(2018)Kuang, Yu, Li, Zhang, and
  Fan]{kuang_integrating_2018}
Zhenzhong Kuang, Jun Yu, Zongmin Li, Baopeng Zhang, and Jianping Fan.
\newblock Integrating multi-level deep learning and concept ontology for
  large-scale visual recognition.
\newblock \emph{PATTERN RECOGNITION}, 78:\penalty0 198--214, June 2018.
\newblock ISSN 0031-3203.
\newblock \doi{10.1016/j.patcog.2018.01.027}.
\newblock 00000 Place: THE BOULEVARD, LANGFORD LANE, KIDLINGTON, OXFORD OX5
  1GB, OXON, ENGLAND Publisher: ELSEVIER SCI LTD Type: Article.

\bibitem[Ayadi et~al.(2019)Ayadi, Samet, Beuvron, and
  Zanni-Merk]{ayadi_ontology_2019}
Ali Ayadi, Ahmed Samet, François de Bertrand~de Beuvron, and Cecilia
  Zanni-Merk.
\newblock Ontology population with deep learning-based {NLP}: a case study on
  the {Biomolecular} {Network} {Ontology}.
\newblock \emph{Procedia Computer Science}, 159:\penalty0 572--581, 2019.
\newblock ISSN 1877-0509.
\newblock \doi{https://doi.org/10.1016/j.procs.2019.09.212}.

\bibitem[Messaoud et~al.(2015)Messaoud, Leray, and Amor]{messaoud_semcado_2015}
Montassar~Ben Messaoud, Philippe Leray, and Nahla~Ben Amor.
\newblock {SemCaDo}: {A} serendipitous strategy for causal discovery and
  ontology evolution.
\newblock \emph{Knowledge-Based Systems}, 76:\penalty0 79--95, 2015.
\newblock ISSN 0950-7051.
\newblock \doi{https://doi.org/10.1016/j.knosys.2014.12.006}.

\bibitem[Thomopoulos et~al.(2013)Thomopoulos, Destercke, Charnomordic, Johnson,
  and Abécassis]{thomopoulos_iterative_2013}
Rallou Thomopoulos, Sébastien Destercke, Brigitte Charnomordic, Iyan Johnson,
  and Joël Abécassis.
\newblock An iterative approach to build relevant ontology-aware data-driven
  models.
\newblock \emph{Information Sciences}, 221:\penalty0 452--472, 2013.
\newblock ISSN 0020-0255.
\newblock \doi{https://doi.org/10.1016/j.ins.2012.09.015}.

\bibitem[Valarakos et~al.(2006)Valarakos, Karkaletsis, Alexopoulou,
  Papadimitriou, Spyropoulos, and Vouros]{valarakos_building_2006}
Alexandros~G. Valarakos, Vangelis Karkaletsis, Dimitra Alexopoulou, Elsa
  Papadimitriou, Constantine~D. Spyropoulos, and George Vouros.
\newblock Building an allergens ontology and maintaining it using machine
  learning techniques.
\newblock \emph{Computers in Biology and Medicine}, 36\penalty0 (10):\penalty0
  1155--1184, 2006.
\newblock ISSN 0010-4825.
\newblock \doi{https://doi.org/10.1016/j.compbiomed.2005.09.007}.

\bibitem[Mihindukulasooriya et~al.(2018)Mihindukulasooriya, Rashid, Rizzo,
  García-Castro, Corcho, and Torchiano]{mihindukulasooriya_rdf_2018}
Nandana Mihindukulasooriya, Mohammad Rifat~Ahmmad Rashid, Giuseppe Rizzo, Raúl
  García-Castro, Oscar Corcho, and Marco Torchiano.
\newblock {RDF} {Shape} {Induction} {Using} {Knowledge} {Base} {Profiling}.
\newblock In \emph{Proceedings of the 33rd {Annual} {ACM} {Symposium} on
  {Applied} {Computing}}, {SAC} '18, pages 1952--1959, New York, NY, USA, 2018.
  Association for Computing Machinery.
\newblock ISBN 978-1-4503-5191-1.
\newblock \doi{10.1145/3167132.3167341}.

\bibitem[Hong et~al.(2021)Hong, Xu, and Shi]{hong_constructing_2021}
Liang Hong, Haoshuai Xu, and Xiaoyue Shi.
\newblock Constructing {Ontology} of {Brain} {Areas} and {Autism} to {Support}
  {Domain} {Knowledge} {Exploration} and {Discovery}.
\newblock \emph{INTERNATIONAL JOURNAL OF COMPUTATIONAL INTELLIGENCE SYSTEMS},
  14\penalty0 (1):\penalty0 834--846, 2021.
\newblock ISSN 1875-6891.
\newblock \doi{10.2991/ijcis.d.210203.005}.
\newblock 00000 Place: 29 AVENUE LAUMIERE, PARIS, 75019, FRANCE Publisher:
  ATLANTIS PRESS Type: Article.

\bibitem[Meroño-Peñuela et~al.(2021)Meroño-Peñuela, Pernisch, Guéret, and
  Schlobach]{merono-penuela_multi-domain_2021}
Albert Meroño-Peñuela, Romana Pernisch, Christophe Guéret, and Stefan
  Schlobach.
\newblock Multi-{Domain} and {Explainable} {Prediction} of {Changes} in {Web}
  {Vocabularies}.
\newblock In \emph{Proceedings of the 11th on {Knowledge} {Capture}
  {Conference}}, K-{CAP} '21, pages 193--200, New York, NY, USA, 2021.
  Association for Computing Machinery.
\newblock ISBN 978-1-4503-8457-5.
\newblock \doi{10.1145/3460210.3493583}.
\newblock 00000 event-place: Virtual Event, USA.

\bibitem[Djellali(2013)]{djellali_using_2013}
Choukri Djellali.
\newblock Using {Hamming} {Similarity} to {Map} {Ontology} {Learning}: {A}
  {New} {Data} {Mining} {System}.
\newblock In \emph{Proceedings of the 2013 {Research} in {Adaptive} and
  {Convergent} {Systems}}, {RACS} '13, pages 82--87, New York, NY, USA, 2013.
  Association for Computing Machinery.
\newblock ISBN 978-1-4503-2348-2.
\newblock \doi{10.1145/2513228.2513232}.

\bibitem[KALFOGLOU and SCHORLEMMER(2003)]{kalfoglou_ontology_2003}
YANNIS KALFOGLOU and MARCO SCHORLEMMER.
\newblock Ontology mapping: the state of the art.
\newblock \emph{The Knowledge Engineering Review}, 18\penalty0 (1):\penalty0
  1–31, 2003.
\newblock \doi{10.1017/S0269888903000651}.

\bibitem[Rico et~al.(2018)Rico, Mihindukulasooriya, Kontokostas, Paulheim,
  Hellmann, and Gómez-Pérez]{rico_predicting_2018}
Mariano Rico, Nandana Mihindukulasooriya, Dimitris Kontokostas, Heiko Paulheim,
  Sebastian Hellmann, and Asunción Gómez-Pérez.
\newblock Predicting {Incorrect} {Mappings}: {A} {Data}-{Driven} {Approach}
  {Applied} to {DBpedia}.
\newblock In \emph{Proceedings of the 33rd {Annual} {ACM} {Symposium} on
  {Applied} {Computing}}, {SAC} '18, pages 323--330, New York, NY, USA, 2018.
  Association for Computing Machinery.
\newblock ISBN 978-1-4503-5191-1.
\newblock \doi{10.1145/3167132.3167164}.

\bibitem[Annane et~al.(2018)Annane, Bellahsene, Azouaou, and
  Jonquet]{annane_building_2018}
Amina Annane, Zohra Bellahsene, Faical Azouaou, and Clement Jonquet.
\newblock Building an effective and efficient background knowledge resource to
  enhance ontology matching.
\newblock \emph{JOURNAL OF WEB SEMANTICS}, 51:\penalty0 51--68, August 2018.
\newblock ISSN 1570-8268.
\newblock \doi{10.1016/j.websem.2018.04.001}.
\newblock 00000 Place: PO BOX 211, 1000 AE AMSTERDAM, NETHERLANDS Publisher:
  ELSEVIER SCIENCE BV Type: Article.

\bibitem[Fanizzi et~al.(2011)Fanizzi, d'Amato, and
  Esposito]{fanizzi_composite_2011}
Nicola Fanizzi, Claudia d'Amato, and Floriana Esposito.
\newblock Composite {Ontology} {Matching} with {Uncertain} {Mappings}
  {Recovery}.
\newblock \emph{SIGAPP Appl. Comput. Rev.}, 11\penalty0 (2):\penalty0 17--29,
  March 2011.
\newblock ISSN 1559-6915.
\newblock \doi{10.1145/1964144.1964148}.
\newblock Place: New York, NY, USA Publisher: Association for Computing
  Machinery.

\bibitem[Rubiolo et~al.(2012)Rubiolo, Caliusco, Stegmayer, Coronel, and
  Fabrizi]{rubiolo_knowledge_2012}
M.~Rubiolo, M.~L. Caliusco, G.~Stegmayer, M.~Coronel, and M.~Gareli Fabrizi.
\newblock Knowledge discovery through ontology matching: {An} approach based on
  an {Artificial} {Neural} {Network} model.
\newblock \emph{Information Sciences}, 194:\penalty0 107--119, 2012.
\newblock ISSN 0020-0255.
\newblock \doi{https://doi.org/10.1016/j.ins.2011.08.008}.

\bibitem[Shannon et~al.(2021)Shannon, Rayapati, Corns, and
  Wunsch]{shannon_comparative_2021}
George~J. Shannon, Naga Rayapati, Steven~M. Corns, and Donald~C. Wunsch.
\newblock Comparative study using inverse ontology cogency and alternatives for
  concept recognition in the annotated {National} {Library} of {Medicine}
  database.
\newblock \emph{Neural Networks}, 139:\penalty0 86--104, 2021.
\newblock ISSN 0893-6080.
\newblock \doi{https://doi.org/10.1016/j.neunet.2021.01.018}.

\bibitem[Chakraborty et~al.(2021)Chakraborty, Bansal, Virgili, Konar, and
  Yaman]{chakraborty_ontoconnect_2021}
Jaydeep Chakraborty, Srividya~K. Bansal, Luca Virgili, Krishanu Konar, and
  Beyza Yaman.
\newblock {OntoConnect}: {Unsupervised} {Ontology} {Alignment} with {Recursive}
  {Neural} {Network}.
\newblock In \emph{Proceedings of the 36th {Annual} {ACM} {Symposium} on
  {Applied} {Computing}}, pages 1874--1882. Association for Computing
  Machinery, New York, NY, USA, 2021.
\newblock ISBN 978-1-4503-8104-8.

\bibitem[Mohan et~al.(2021)Mohan, Angell, Monath, and McCallum]{mohan_low_2021}
Sunil Mohan, Rico Angell, Nicholas Monath, and Andrew McCallum.
\newblock Low {Resource} {Recognition} and {Linking} of {Biomedical} {Concepts}
  from a {Large} {Ontology}.
\newblock In \emph{Proceedings of the 12th {ACM} {Conference} on
  {Bioinformatics}, {Computational} {Biology}, and {Health} {Informatics}},
  {BCB} '21, New York, NY, USA, 2021. Association for Computing Machinery.
\newblock ISBN 978-1-4503-8450-6.
\newblock \doi{10.1145/3459930.3469524}.
\newblock 00000 event-place: Gainesville, Florida.

\bibitem[Mao et~al.(2010)Mao, Peng, and Spring]{mao_adaptive_2010}
Ming Mao, Yefei Peng, and Michael Spring.
\newblock An adaptive ontology mapping approach with neural network based
  constraint satisfaction.
\newblock \emph{Journal of Web Semantics}, 8\penalty0 (1):\penalty0 14--25,
  2010.
\newblock ISSN 1570-8268.
\newblock \doi{https://doi.org/10.1016/j.websem.2009.11.002}.

\bibitem[Zhou and El-Gohary(2021)]{zhou_semantic_2021}
Peng Zhou and Nora El-Gohary.
\newblock Semantic information alignment of {BIMs} to computer-interpretable
  regulations using ontologies and deep learning.
\newblock \emph{Advanced Engineering Informatics}, 48:\penalty0 101239, 2021.
\newblock ISSN 1474-0346.
\newblock \doi{https://doi.org/10.1016/j.aei.2020.101239}.

\bibitem[Sirin et~al.(2007)Sirin, Parsia, Grau, Kalyanpur, and
  Katz]{sirin_pellet_2007}
Evren Sirin, Bijan Parsia, Bernardo~Cuenca Grau, Aditya Kalyanpur, and Yarden
  Katz.
\newblock Pellet: A practical owl-dl reasoner.
\newblock \emph{Journal of Web Semantics}, 5\penalty0 (2):\penalty0 51--53,
  2007.

\bibitem[Shearer et~al.(2008)Shearer, Motik, and Horrocks]{shearer_hermit_2008}
Robert~DC Shearer, Boris Motik, and Ian Horrocks.
\newblock Hermit: A highly-efficient owl reasoner.
\newblock In \emph{Owled}, volume 432, page~91, 2008.

\bibitem[Bock et~al.(2012)Bock, Lösch, and Wang]{bock_automatic_2012}
Jürgen Bock, Uta Lösch, and Hai Wang.
\newblock Automatic {Reasoner} {Selection} {Using} {Machine} {Learning}.
\newblock In \emph{Proceedings of the 2nd {International} {Conference} on {Web}
  {Intelligence}, {Mining} and {Semantics}}, {WIMS} '12, New York, NY, USA,
  2012. Association for Computing Machinery.
\newblock ISBN 978-1-4503-0915-8.
\newblock \doi{10.1145/2254129.2254159}.

\bibitem[Pan et~al.(2018)Pan, Bobed, Guclu, Bobillo, Kollingbaum, Mena, and
  Li]{pan_predicting_2018}
Jeff~Z. Pan, Carlos Bobed, Isa Guclu, Fernando Bobillo, Martin~J. Kollingbaum,
  Eduardo Mena, and Yuan-Fang Li.
\newblock Predicting {Reasoner} {Performance} on {ABox} {Intensive} {OWL} 2
  {EL} {Ontologies}.
\newblock \emph{INTERNATIONAL JOURNAL ON SEMANTIC WEB AND INFORMATION SYSTEMS},
  14\penalty0 (1, SI):\penalty0 1--30, March 2018.
\newblock ISSN 1552-6283.
\newblock \doi{10.4018/IJSWIS.2018010101}.
\newblock 00000 Place: 701 E CHOCOLATE AVE, STE 200, HERSHEY, PA 17033-1240 USA
  Publisher: IGI GLOBAL Type: Article; Proceedings Paper.

\bibitem[Mehri et~al.(2021)Mehri, Haarslev, and Chinaei]{mehri_machine_2021}
Razieh Mehri, Volker Haarslev, and Hamidreza Chinaei.
\newblock A machine learning approach for optimizing heuristic decision-making
  in {Web} {Ontology} {Language} reasoners.
\newblock \emph{COMPUTATIONAL INTELLIGENCE}, 37\penalty0 (1):\penalty0
  273--314, February 2021.
\newblock ISSN 0824-7935.
\newblock \doi{10.1111/coin.12404}.
\newblock 00000 Place: 111 RIVER ST, HOBOKEN 07030-5774, NJ USA Publisher:
  WILEY Type: Article.

\bibitem[Hohenecker and Lukasiewicz(2020)]{hohenecker_ontology_2020}
Patrick Hohenecker and Thomas Lukasiewicz.
\newblock Ontology {Reasoning} with {Deep} {Neural} {Networks}.
\newblock \emph{JOURNAL OF ARTIFICIAL INTELLIGENCE RESEARCH}, 68:\penalty0
  503--540, 2020.
\newblock ISSN 1076-9757.
\newblock 00000 Place: USC INFORMATION SCIENCES INST, 4676 ADMIRALITY WAY,
  MARINA DEL REY, CA 90292-6695 USA Publisher: AI ACCESS FOUNDATION Type:
  Article.

\bibitem[Rizzo et~al.(2017)Rizzo, d'Amato, Fanizzi, and
  Esposito]{rizzo_tree-based_2017}
Giuseppe Rizzo, Claudia d'Amato, Nicola Fanizzi, and Floriana Esposito.
\newblock Tree-based models for inductive classification on the {Web} {Of}
  {Data}.
\newblock \emph{JOURNAL OF WEB SEMANTICS}, 45:\penalty0 1--22, August 2017.
\newblock ISSN 1570-8268.
\newblock \doi{10.1016/j.websem.2017.05.001}.
\newblock 00000 Place: RADARWEG 29, 1043 NX AMSTERDAM, NETHERLANDS Publisher:
  ELSEVIER Type: Article.

\bibitem[Makni and Hendler(2019)]{makni_deep_2019}
Bassem Makni and James Hendler.
\newblock Deep learning for noise-tolerant {RDFS} reasoning.
\newblock \emph{SEMANTIC WEB}, 10\penalty0 (5):\penalty0 823--862, 2019.
\newblock ISSN 1570-0844.
\newblock \doi{10.3233/SW-190363}.
\newblock 00000 Place: NIEUWE HEMWEG 6B, 1013 BG AMSTERDAM, NETHERLANDS
  Publisher: IOS PRESS Type: Article.

\bibitem[Zhao and Zhang(2018)]{zhao_domain-specific_2018}
Grace Zhao and Xiaowen Zhang.
\newblock Domain-{Specific} {Ontology} {Concept} {Extraction} and {Hierarchy}
  {Extension}.
\newblock In \emph{Proceedings of the 2nd {International} {Conference} on
  {Natural} {Language} {Processing} and {Information} {Retrieval}}, {NLPIR}
  2018, pages 60--64, New York, NY, USA, 2018. Association for Computing
  Machinery.
\newblock ISBN 978-1-4503-6551-2.
\newblock \doi{10.1145/3278293.3278302}.

\bibitem[Gao et~al.(2018)Gao, Chen, Baig, and Zhang]{gao_ontology_2018}
Wei Gao, Yaojun Chen, Abdul~Qudair Baig, and Yunqing Zhang.
\newblock Ontology geometry distance computation using deep learning
  technology.
\newblock \emph{JOURNAL OF INTELLIGENT \& FUZZY SYSTEMS}, 35\penalty0
  (4):\penalty0 4517--4524, 2018.
\newblock ISSN 1064-1246.
\newblock \doi{10.3233/JIFS-169770}.
\newblock 00000 Place: NIEUWE HEMWEG 6B, 1013 BG AMSTERDAM, NETHERLANDS
  Publisher: IOS PRESS Type: Article.

\bibitem[Xue et~al.(2021)Xue, Wang, and Liu]{xue_matching_2021}
Xingsi Xue, Haolin Wang, and Wenyu Liu.
\newblock Matching sensor ontologies with unsupervised neural network with
  competitive learning.
\newblock \emph{PEERJ COMPUTER SCIENCE}, 7, November 2021.
\newblock \doi{10.7717/peerj-cs.763}.

\bibitem[Lawrynowicz and Potoniec(2014)]{lawrynowicz_pattern_2014}
Agnieszka Lawrynowicz and Jedrzej Potoniec.
\newblock Pattern {Based} {Feature} {Construction} in {Semantic} {Data}
  {Mining}.
\newblock \emph{INTERNATIONAL JOURNAL ON SEMANTIC WEB AND INFORMATION SYSTEMS},
  10\penalty0 (1):\penalty0 27--65, 2014.
\newblock ISSN 1552-6283.
\newblock \doi{10.4018/ijswis.2014010102}.
\newblock 00000 Place: 701 E CHOCOLATE AVE, STE 200, HERSHEY, PA 17033-1240 USA
  Publisher: IGI GLOBAL Type: Article.

\bibitem[Karniadakis et~al.(2021)Karniadakis, Kevrekidis, Lu, Perdikaris, Wang,
  and Yang]{karniadakis_physics-informed_2021}
George~Em Karniadakis, Ioannis~G. Kevrekidis, Lu~Lu, Paris Perdikaris, Sifan
  Wang, and Liu Yang.
\newblock Physics-informed machine learning.
\newblock \emph{Nature Reviews Physics}, 3\penalty0 (6):\penalty0 422--440,
  June 2021.
\newblock ISSN 2522-5820.
\newblock \doi{10.1038/s42254-021-00314-5}.
\newblock 00237 Number: 6 Publisher: Nature Publishing Group.

\bibitem[Duboue(2020)]{duboue_art_2020}
Pablo Duboue.
\newblock \emph{The {Art} of {Feature} {Engineering}: {Essentials} for
  {Machine} {Learning}}.
\newblock Cambridge University Press, June 2020.
\newblock ISBN 978-1-108-70938-5.
\newblock 00035 Google-Books-ID: \_BzhDwAAQBAJ.

\bibitem[Pozveh et~al.(2018)Pozveh, Monadjemi, and
  Ahmadi]{pozveh_fnlp-ont_2018}
Zahra~Hosseini Pozveh, Amirhassan Monadjemi, and Ali Ahmadi.
\newblock {FNLP}-{ONT}: {A} feasible ontology for improving {NLP} tasks in
  {Persian}.
\newblock \emph{EXPERT SYSTEMS}, 35\penalty0 (4, SI), August 2018.
\newblock ISSN 0266-4720.
\newblock \doi{10.1111/exsy.12282}.
\newblock 00000 Place: 111 RIVER ST, HOBOKEN 07030-5774, NJ USA Publisher:
  WILEY Type: Article.

\bibitem[Gomathi and Karlekar(2019)]{gomathi_ontology_2019}
N.~Gomathi and Nandkishor~P. Karlekar.
\newblock Ontology and {Hybrid} {Optimization} {Based} {SVNN} for {Privacy}
  {Preserved} {Medical} {Data} {Classification} in {Cloud}.
\newblock \emph{INTERNATIONAL JOURNAL ON ARTIFICIAL INTELLIGENCE TOOLS},
  28\penalty0 (3), May 2019.
\newblock ISSN 0218-2130.
\newblock \doi{10.1142/S021821301950009X}.
\newblock 00000 Place: 5 TOH TUCK LINK, SINGAPORE 596224, SINGAPORE Publisher:
  WORLD SCIENTIFIC PUBL CO PTE LTD Type: Article.

\bibitem[Bellman(1961)]{bellman_adaptive_1961}
Richard~E. Bellman.
\newblock \emph{Adaptive {Control} {Processes}: {A} {Guided} {Tour}}.
\newblock Princeton University Press, December 1961.
\newblock ISBN 978-1-4008-7466-8.
\newblock \doi{10.1515/9781400874668}.

\bibitem[Kumar et~al.(2020)Kumar, Pannu, and Malhi]{kumar_aspect-based_2020}
Ravindra Kumar, Husanbir~Singh Pannu, and Avleen~Kaur Malhi.
\newblock Aspect-based sentiment analysis using deep networks and stochastic
  optimization.
\newblock \emph{NEURAL COMPUTING \& APPLICATIONS}, 32\penalty0 (8):\penalty0
  3221--3235, April 2020.
\newblock ISSN 0941-0643.
\newblock \doi{10.1007/s00521-019-04105-z}.
\newblock 00000 Place: 236 GRAYS INN RD, 6TH FLOOR, LONDON WC1X 8HL, ENGLAND
  Publisher: SPRINGER LONDON LTD Type: Article.

\bibitem[Sabra et~al.(2020)Sabra, Malik, Afzal, Sabeeh, and
  Eddine]{sabra_hybrid_2020}
Susan Sabra, Khalid~Mahmood Malik, Muhammad Afzal, Vian Sabeeh, and
  Ahmad~Charaf Eddine.
\newblock A hybrid knowledge and ensemble classification approach for
  prediction of venous thromboembolism.
\newblock \emph{EXPERT SYSTEMS}, 37\penalty0 (1, SI), February 2020.
\newblock ISSN 0266-4720.
\newblock \doi{10.1111/exsy.12388}.
\newblock 00000 Place: 111 RIVER ST, HOBOKEN 07030-5774, NJ USA Publisher:
  WILEY Type: Article.

\bibitem[Ahani et~al.(2021)Ahani, Nilashi, Zogaan, Samad, Aljehane, Alhargan,
  Mohd, Ahmadi, and Sanzogni]{ahani_evaluating_2021}
Ali Ahani, Mehrbakhsh Nilashi, Waleed~Abdu Zogaan, Sarminah Samad, Nojood~O.
  Aljehane, Ashwaq Alhargan, Saidatulakmal Mohd, Hossein Ahmadi, and Louis
  Sanzogni.
\newblock Evaluating medical travelers’ satisfaction through online review
  analysis.
\newblock \emph{Journal of Hospitality and Tourism Management}, 48:\penalty0
  519--537, 2021.
\newblock ISSN 1447-6770.
\newblock \doi{https://doi.org/10.1016/j.jhtm.2021.08.005}.

\bibitem[Akila et~al.(2021)Akila, Priyadharshini, Ulaganathan, Prempriya,
  Yuvasri, Praba, and {Veeramuthuvenkatesh}]{akila_ontology_2021}
K.~Akila, S.~Indra Priyadharshini, Pradheeba Ulaganathan, P.~Prempriya,
  B.~Yuvasri, T.~Suriya Praba, and {Veeramuthuvenkatesh}.
\newblock Ontology based multiobject segmentation and classification in sports
  videos.
\newblock \emph{JOURNAL OF INTELLIGENT \& FUZZY SYSTEMS}, 41\penalty0
  (5):\penalty0 5399--5409, 2021.
\newblock ISSN 1064-1246.
\newblock \doi{10.3233/JIFS-189862}.

\bibitem[Zhao et~al.(2021)Zhao, Xue, Wang, and Du]{zhao_adaptive_2021}
Delong Zhao, Dun Xue, Xiaoyao Wang, and Fuzhou Du.
\newblock Adaptive vision inspection for multi-type electronic products based
  on prior knowledge.
\newblock \emph{Journal of Industrial Information Integration}, page 100283,
  2021.
\newblock ISSN 2452-414X.
\newblock \doi{https://doi.org/10.1016/j.jii.2021.100283}.

\bibitem[Messaoudi et~al.(2021)Messaoudi, Jaziri, Mtibaa, Gargouri, and
  Vacavant]{messaoudi_ontology-driven_2021}
Rim Messaoudi, Faouzi Jaziri, Achraf Mtibaa, Faiez Gargouri, and Antoine
  Vacavant.
\newblock Ontology-{Driven} {Approach} for {Liver} {MRI} {Classification} and
  {HCC} {Detection}.
\newblock \emph{INTERNATIONAL JOURNAL OF PATTERN RECOGNITION AND ARTIFICIAL
  INTELLIGENCE}, 35\penalty0 (12), September 2021.
\newblock ISSN 0218-0014.
\newblock \doi{10.1142/S0218001421600077}.

\bibitem[Rinaldi et~al.(2021)Rinaldi, Russo, and
  Tommasino]{rinaldi_semantic_2021}
Antonio~M. Rinaldi, Cristiano Russo, and Cristian Tommasino.
\newblock A semantic approach for document classification using deep neural
  networks and multimedia knowledge graph.
\newblock \emph{Expert Systems with Applications}, 169:\penalty0 114320, 2021.
\newblock ISSN 0957-4174.
\newblock \doi{https://doi.org/10.1016/j.eswa.2020.114320}.

\bibitem[Liu et~al.(2021)Liu, Gao, and Ma]{liu_photovoltaic_2021}
Hongfei Liu, Qian Gao, and Pengcheng Ma.
\newblock Photovoltaic generation power prediction research based on high
  quality context ontology and gated recurrent neural network.
\newblock \emph{Sustainable Energy Technologies and Assessments}, 45:\penalty0
  101191, 2021.
\newblock ISSN 2213-1388.
\newblock \doi{https://doi.org/10.1016/j.seta.2021.101191}.

\bibitem[Mabrouk et~al.(2020)Mabrouk, Hlaoua, and
  Omri]{mabrouk_exploiting_2020}
Olfa Mabrouk, Lobna Hlaoua, and Mohamed~Nazih Omri.
\newblock Exploiting ontology information in fuzzy {SVM} social media profile
  classification.
\newblock \emph{APPLIED INTELLIGENCE}, November 2020.
\newblock ISSN 0924-669X.
\newblock \doi{10.1007/s10489-020-01939-2}.
\newblock 00000 Place: VAN GODEWIJCKSTRAAT 30, 3311 GZ DORDRECHT, NETHERLANDS
  Publisher: SPRINGER Type: Article; Early Access.

\bibitem[Zhang et~al.(2021)Zhang, Wu, Chen, Lu, Na, and Qi]{zhang_auto_2021}
Long Zhang, Tianxing Wu, Xiuqi Chen, Bingjie Lu, Chongning Na, and Guilin Qi.
\newblock Auto {Insurance} {Knowledge} {Graph} {Construction} and {Its}
  {Application} to {Fraud} {Detection}.
\newblock In \emph{The 10th {International} {Joint} {Conference} on {Knowledge}
  {Graphs}}, {IJCKG}'21, pages 64--70, New York, NY, USA, 2021. Association for
  Computing Machinery.
\newblock ISBN 978-1-4503-9565-6.
\newblock \doi{10.1145/3502223.3502231}.

\bibitem[Chen et~al.(2021)Chen, Lecue, Pan, Deng, and
  Chen]{chen_knowledge_2021}
Jiaoyan Chen, Freddy Lecue, Jeff~Z. Pan, Shumin Deng, and Huajun Chen.
\newblock Knowledge graph embeddings for dealing with concept drift in machine
  learning.
\newblock \emph{JOURNAL OF WEB SEMANTICS}, 67, February 2021.
\newblock ISSN 1570-8268.
\newblock \doi{10.1016/j.websem.2020.100625}.
\newblock 00000 Place: RADARWEG 29, 1043 NX AMSTERDAM, NETHERLANDS Publisher:
  ELSEVIER Type: Article.

\bibitem[Benarab et~al.(2019)Benarab, Rafique, and Sun]{benarab_ontology_2019}
Achref Benarab, Fahad Rafique, and Jianguo Sun.
\newblock An {Ontology} {Embedding} {Approach} {Based} on {Multiple} {Neural}
  {Networks}.
\newblock In \emph{Proceedings of the 2019 11th {International} {Conference} on
  {Machine} {Learning} and {Computing}}, {ICMLC} '19, pages 186--190, New York,
  NY, USA, 2019. Association for Computing Machinery.
\newblock ISBN 978-1-4503-6600-7.
\newblock \doi{10.1145/3318299.3318365}.

\bibitem[Jang et~al.(2018)Jang, Ham, Lee, and Kim]{jang_cross-language_2018}
Youngsoo Jang, Jiyeon Ham, Byung-Jun Lee, and Kee-Eung Kim.
\newblock Cross-{Language} {Neural} {Dialog} {State} {Tracker} for {Large}
  {Ontologies} {Using} {Hierarchical} {Attention}.
\newblock \emph{IEEE/ACM Trans. Audio, Speech and Lang. Proc.}, 26\penalty0
  (11):\penalty0 2072--2082, November 2018.
\newblock ISSN 2329-9290.
\newblock \doi{10.1109/TASLP.2018.2852492}.

\bibitem[Ali et~al.(2021)Ali, El-Sappagh, Islam, Ali, Attique, Imran, and
  Kwak]{ali_intelligent_2021}
Farman Ali, Shaker El-Sappagh, S.~M.~Riazul Islam, Amjad Ali, Muhammad Attique,
  Muhammad Imran, and Kyung-Sup Kwak.
\newblock An intelligent healthcare monitoring framework using wearable sensors
  and social networking data.
\newblock \emph{Future Generation Computer Systems}, 114:\penalty0 23--43,
  2021.
\newblock ISSN 0167-739X.
\newblock \doi{https://doi.org/10.1016/j.future.2020.07.047}.

\bibitem[Amador-Domínguez et~al.(2021)Amador-Domínguez, Serrano, Manrique,
  Hohenecker, and Lukasiewicz]{amador-dominguez_ontology-based_2021}
Elvira Amador-Domínguez, Emilio Serrano, Daniel Manrique, Patrick Hohenecker,
  and Thomas Lukasiewicz.
\newblock An ontology-based deep learning approach for triple classification
  with out-of-knowledge-base entities.
\newblock \emph{Information Sciences}, 564:\penalty0 85--102, 2021.
\newblock ISSN 0020-0255.
\newblock \doi{https://doi.org/10.1016/j.ins.2021.02.018}.

\bibitem[Emele et~al.(2012)Emele, Norman, Sensoy, and
  Parsons]{emele_learning_2012}
Chukwuemeka~David Emele, Timothy~J. Norman, Murat Sensoy, and Simon Parsons.
\newblock Learning strategies for task delegation in norm-governed
  environments.
\newblock \emph{AUTONOMOUS AGENTS AND MULTI-AGENT SYSTEMS}, 25\penalty0 (3,
  SI):\penalty0 499--525, November 2012.
\newblock ISSN 1387-2532.
\newblock \doi{10.1007/s10458-012-9194-9}.
\newblock 00000 Place: VAN GODEWIJCKSTRAAT 30, 3311 GZ DORDRECHT, NETHERLANDS
  Publisher: SPRINGER Type: Article.

\bibitem[Ruiz-Sarmiento et~al.(2019)Ruiz-Sarmiento, Galindo, Monroy, Moreno,
  and Gonzalez-Jimenez]{ruiz-sarmiento_ontology-based_2019}
Jose-Raul Ruiz-Sarmiento, Cipriano Galindo, Javier Monroy, Francisco-Angel
  Moreno, and Javier Gonzalez-Jimenez.
\newblock Ontology-based conditional random fields for object recognition.
\newblock \emph{KNOWLEDGE-BASED SYSTEMS}, 168:\penalty0 100--108, March 2019.
\newblock ISSN 0950-7051.
\newblock \doi{10.1016/j.knosys.2019.01.005}.
\newblock 00000 Place: RADARWEG 29, 1043 NX AMSTERDAM, NETHERLANDS Publisher:
  ELSEVIER Type: Article.

\bibitem[Gabriel et~al.(2014)Gabriel, Negru, and
  Zaharie]{gabriel_neuroevolution_2014}
Iuhasz Gabriel, Viorel Negru, and Daniela Zaharie.
\newblock Neuroevolution {Based} {Multi}-{Agent} {System} with {Ontology}
  {Based} {Template} {Creation} for {Micromanagement} in {Real}-{Time}
  {Strategy} {Games}.
\newblock \emph{INFORMATION TECHNOLOGY AND CONTROL}, 43\penalty0 (1):\penalty0
  98--109, 2014.
\newblock ISSN 1392-124X.
\newblock \doi{10.5755/j01.itc.43.1.4600}.
\newblock 00000 Place: KAUNAS UNIV TECHNOL, DEPT ELECTRONICS ENGINEERING,
  STUDENTU STR 50, KAUNAS, LT-51368, LITHUANIA Publisher: KAUNAS UNIV
  TECHNOLOGY Type: Article.

\bibitem[Huang et~al.(2019)Huang, Zanni-Merk, and
  Crémilleux]{huang_enhancing_2019}
Xin Huang, Cecilia Zanni-Merk, and Bruno Crémilleux.
\newblock Enhancing {Deep} {Learning} with {Semantics}: an application to
  manufacturing time series analysis.
\newblock \emph{Procedia Computer Science}, 159:\penalty0 437--446, 2019.
\newblock ISSN 1877-0509.
\newblock \doi{https://doi.org/10.1016/j.procs.2019.09.198}.

\bibitem[Kuang et~al.(2021)Kuang, Zhang, Yu, Li, and Fan]{kuang_deep_2021}
Zhenzhong Kuang, Xin Zhang, Jun Yu, Zongmin Li, and Jianping Fan.
\newblock Deep embedding of concept ontology for hierarchical fashion
  recognition.
\newblock \emph{NEUROCOMPUTING}, 425:\penalty0 191--206, February 2021.
\newblock ISSN 0925-2312.
\newblock \doi{10.1016/j.neucom.2020.04.085}.
\newblock 00001 Place: RADARWEG 29, 1043 NX AMSTERDAM, NETHERLANDS Publisher:
  ELSEVIER Type: Article.

\bibitem[Fu et~al.(2015)Fu, Mei, Yang, Lu, and Rui]{fu_tagging_2015}
Jianlong Fu, Tao Mei, Kuiyuan Yang, Hanqing Lu, and Yong Rui.
\newblock Tagging {Personal} {Photos} with {Transfer} {Deep} {Learning}.
\newblock In \emph{Proceedings of the 24th {International} {Conference} on
  {World} {Wide} {Web}}, {WWW} '15, pages 344--354, Republic and Canton of
  Geneva, CHE, 2015. International World Wide Web Conferences Steering
  Committee.
\newblock ISBN 978-1-4503-3469-3.
\newblock \doi{10.1145/2736277.2741112}.

\bibitem[Serafini et~al.(2017)Serafini, Donadello, and
  Garcez]{serafini_learning_2017}
Luciano Serafini, Ivan Donadello, and Artur~d'Avila Garcez.
\newblock Learning and {Reasoning} in {Logic} {Tensor} {Networks}: {Theory} and
  {Application} to {Semantic} {Image} {Interpretation}.
\newblock In \emph{Proceedings of the {Symposium} on {Applied} {Computing}},
  {SAC} '17, pages 125--130, New York, NY, USA, 2017. Association for Computing
  Machinery.
\newblock ISBN 978-1-4503-4486-9.
\newblock \doi{10.1145/3019612.3019642}.

\bibitem[Donís~Ebri(2021)]{donis_ebri_combining_2021}
Pablo Donís~Ebri.
\newblock Combining ontologies and {Machine} {Learning} for {Explainable}
  {Artificial} {Intelligence}.
\newblock Master's thesis, E.T.S. de Ingenieros Informáticos (UPM), 2021.

\bibitem[Lundberg and Lee(2017)]{lundberg_unified_2017}
Scott~M Lundberg and Su-In Lee.
\newblock A {Unified} {Approach} to {Interpreting} {Model} {Predictions}.
\newblock In \emph{Advances in {Neural} {Information} {Processing} {Systems}},
  volume~30. Curran Associates, Inc., 2017.

\bibitem[Ribeiro et~al.(2016)Ribeiro, Singh, and Guestrin]{ribeiro_why_2016}
Marco~Tulio Ribeiro, Sameer Singh, and Carlos Guestrin.
\newblock "{Why} {Should} {I} {Trust} {You}?": {Explaining} the {Predictions}
  of {Any} {Classifier}.
\newblock \emph{arXiv:1602.04938 [cs, stat]}, August 2016.
\newblock 08107 arXiv: 1602.04938.

\bibitem[Confalonieri et~al.(2021)Confalonieri, Weyde, Besold, and
  Martín]{confalonieri_using_2021}
Roberto Confalonieri, Tillman Weyde, Tarek~R. Besold, and Fermín Moscoso
  del~Prado Martín.
\newblock Using ontologies to enhance human understandability of global
  post-hoc explanations of black-box models.
\newblock \emph{Artificial Intelligence}, 296:\penalty0 103471, 2021.
\newblock ISSN 0004-3702.
\newblock \doi{https://doi.org/10.1016/j.artint.2021.103471}.

\bibitem[Panigutti et~al.(2020)Panigutti, Perotti, and
  Pedreschi]{panigutti_doctor_2020}
Cecilia Panigutti, Alan Perotti, and Dino Pedreschi.
\newblock Doctor {XAI}: {An} {Ontology}-{Based} {Approach} to {Black}-{Box}
  {Sequential} {Data} {Classification} {Explanations}.
\newblock In \emph{Proceedings of the 2020 {Conference} on {Fairness},
  {Accountability}, and {Transparency}}, {FAT}* '20, pages 629--639, New York,
  NY, USA, 2020. Association for Computing Machinery.
\newblock ISBN 978-1-4503-6936-7.
\newblock \doi{10.1145/3351095.3372855}.

\bibitem[Rath et~al.(2009)Rath, Devaurs, and Lindstaedt]{rath_uico_2009}
Andreas~S. Rath, Didier Devaurs, and Stefanie~N. Lindstaedt.
\newblock {UICO}: {An} {Ontology}-{Based} {User} {Interaction} {Context}
  {Model} for {Automatic} {Task} {Detection} on the {Computer} {Desktop}.
\newblock In \emph{Proceedings of the 1st {Workshop} on {Context},
  {Information} and {Ontologies}}, {CIAO} '09, New York, NY, USA, 2009.
  Association for Computing Machinery.
\newblock ISBN 978-1-60558-528-4.
\newblock \doi{10.1145/1552262.1552270}.

\bibitem[Santos et~al.(2012)Santos, Laorden, Sanz, and
  Bringas]{santos_enhanced_2012}
Igor Santos, Carlos Laorden, Borja Sanz, and Pablo~G. Bringas.
\newblock Enhanced {Topic}-based {Vector} {Space} {Model} for semantics-aware
  spam filtering.
\newblock \emph{EXPERT SYSTEMS WITH APPLICATIONS}, 39\penalty0 (1):\penalty0
  437--444, January 2012.
\newblock ISSN 0957-4174.
\newblock \doi{10.1016/j.eswa.2011.07.034}.
\newblock 00000 Place: THE BOULEVARD, LANGFORD LANE, KIDLINGTON, OXFORD OX5
  1GB, ENGLAND Publisher: PERGAMON-ELSEVIER SCIENCE LTD Type: Article;
  Proceedings Paper.

\bibitem[Pancerz and Lewicki(2014)]{pancerz_encoding_2014}
Krzysztof Pancerz and Arkadiusz Lewicki.
\newblock Encoding symbolic features in simple decision systems over
  ontological graphs for {PSO} and neural network based classifiers.
\newblock \emph{NEUROCOMPUTING}, 144:\penalty0 338--345, November 2014.
\newblock ISSN 0925-2312.
\newblock \doi{10.1016/j.neucom.2014.04.038}.
\newblock 00000 Place: PO BOX 211, 1000 AE AMSTERDAM, NETHERLANDS Publisher:
  ELSEVIER SCIENCE BV Type: Article.

\bibitem[Wan and Mak(2018)]{wan_predicting_2018}
Shibiao Wan and Man-Wai Mak.
\newblock Predicting subcellular localization of multi-location proteins by
  improving support vector machines with an adaptive-decision scheme.
\newblock \emph{INTERNATIONAL JOURNAL OF MACHINE LEARNING AND CYBERNETICS},
  9\penalty0 (3, SI):\penalty0 399--411, March 2018.
\newblock ISSN 1868-8071.
\newblock \doi{10.1007/s13042-015-0460-4}.
\newblock 00000 Place: TIERGARTENSTRASSE 17, D-69121 HEIDELBERG, GERMANY
  Publisher: SPRINGER HEIDELBERG Type: Article.

\bibitem[Salguero et~al.(2019)Salguero, Medina, Delatorre, and
  Espinilla]{salguero_methodology_2019}
A.~G. Salguero, J.~Medina, P.~Delatorre, and M.~Espinilla.
\newblock Methodology for improving classification accuracy using ontologies:
  application in the recognition of activities of daily living.
\newblock \emph{JOURNAL OF AMBIENT INTELLIGENCE AND HUMANIZED COMPUTING},
  10\penalty0 (6, SI):\penalty0 2125--2142, June 2019.
\newblock ISSN 1868-5137.
\newblock \doi{10.1007/s12652-018-0769-4}.
\newblock 00000 Place: TIERGARTENSTRASSE 17, D-69121 HEIDELBERG, GERMANY
  Publisher: SPRINGER HEIDELBERG Type: Article.

\bibitem[Abdollahi et~al.(2021)Abdollahi, Gao, Mei, Ghosh, Li, and
  Narag]{abdollahi_substituting_2021}
Mahdi Abdollahi, Xiaoying Gao, Yi~Mei, Shameek Ghosh, Jinyan Li, and Michael
  Narag.
\newblock Substituting clinical features using synthetic medical phrases:
  {Medical} text data augmentation techniques.
\newblock \emph{Artificial Intelligence in Medicine}, 120:\penalty0 102167,
  2021.
\newblock ISSN 0933-3657.
\newblock \doi{https://doi.org/10.1016/j.artmed.2021.102167}.

\bibitem[Wang et~al.(2021{\natexlab{a}})Wang, Li, Song, Yan, Shi, and
  Yao]{wang_evaluating_2021}
Zeheng Wang, Liang Li, Miao Song, Jing Yan, Junjie Shi, and Yuanzhe Yao.
\newblock Evaluating the {Traditional} {Chinese} {Medicine} ({TCM})
  {Officially} {Recommended} in {China} for {COVID}-19 {Using}
  {Ontology}-{Based} {Side}-{Effect} {Prediction} {Framework} ({OSPF}) and
  {Deep} {Learning}.
\newblock \emph{Journal of Ethnopharmacology}, 272:\penalty0 113957,
  2021{\natexlab{a}}.
\newblock ISSN 0378-8741.
\newblock \doi{https://doi.org/10.1016/j.jep.2021.113957}.

\bibitem[Ye et~al.(2015{\natexlab{a}})Ye, Stevenson, and
  Dobson]{ye_usmart_2015}
Juan Ye, Graeme Stevenson, and Simon Dobson.
\newblock {USMART}: {An} {Unsupervised} {Semantic} {Mining} {Activity}
  {Recognition} {Technique}.
\newblock \emph{ACM TRANSACTIONS ON INTERACTIVE INTELLIGENT SYSTEMS},
  4\penalty0 (4, SI), January 2015{\natexlab{a}}.
\newblock ISSN 2160-6455.
\newblock \doi{10.1145/2662870}.
\newblock 00000 Place: 2 PENN PLAZA, STE 701, NEW YORK, NY 10121-0701 USA
  Publisher: ASSOC COMPUTING MACHINERY Type: Article.

\bibitem[Oliveira et~al.(2021)Oliveira, Rocha~Silva, and
  Bernardino]{oliveira_wine_2021}
Luis Oliveira, Rodrigo Rocha~Silva, and Jorge Bernardino.
\newblock Wine {Ontology} {Influence} in a {Recommendation} {System}.
\newblock \emph{BIG DATA AND COGNITIVE COMPUTING}, 5\penalty0 (2), June 2021.
\newblock \doi{10.3390/bdcc5020016}.

\bibitem[Castillo et~al.(2008)Castillo, Armengol, Onaindía, Sebastiá,
  González-Boticario, Rodríguez, Fernández, Arias, and
  Borrajo]{castillo_samap_2008}
Luis Castillo, Eva Armengol, Eva Onaindía, Laura Sebastiá, Jesús
  González-Boticario, Antonio Rodríguez, Susana Fernández, Juan~D. Arias,
  and Daniel Borrajo.
\newblock samap: {An} user-oriented adaptive system for planning tourist
  visits.
\newblock \emph{Expert Systems with Applications}, 34\penalty0 (2):\penalty0
  1318--1332, 2008.
\newblock ISSN 0957-4174.
\newblock \doi{https://doi.org/10.1016/j.eswa.2006.12.029}.

\bibitem[Rajput and Haider(2011)]{rajput_bnosa_2011}
Quratulain Rajput and Sajjad Haider.
\newblock {BNOSA}: {A} {Bayesian} network and ontology based semantic
  annotation framework.
\newblock \emph{Journal of Web Semantics}, 9\penalty0 (2):\penalty0 99--112,
  2011.
\newblock ISSN 1570-8268.
\newblock \doi{https://doi.org/10.1016/j.websem.2011.04.002}.

\bibitem[Hsieh et~al.(2013)Hsieh, Lee, and Chen]{hsieh_transformation_2013}
Nan-Chen Hsieh, Kuo-Chen Lee, and Wei Chen.
\newblock The transformation of surgery patient care with a clinical research
  information system.
\newblock \emph{EXPERT SYSTEMS WITH APPLICATIONS}, 40\penalty0 (1):\penalty0
  211--221, January 2013.
\newblock ISSN 0957-4174.
\newblock \doi{10.1016/j.eswa.2012.07.020}.
\newblock 00000 Place: THE BOULEVARD, LANGFORD LANE, KIDLINGTON, OXFORD OX5
  1GB, ENGLAND Publisher: PERGAMON-ELSEVIER SCIENCE LTD Type: Article.

\bibitem[Agarwal et~al.(2015)Agarwal, Poria, Mittal, Gelbukh, and
  Hussain]{agarwal_concept-level_2015}
Basant Agarwal, Soujanya Poria, Namita Mittal, Alexander Gelbukh, and Amir
  Hussain.
\newblock Concept-{Level} {Sentiment} {Analysis} with {Dependency}-{Based}
  {Semantic} {Parsing}: {A} {Novel} {Approach}.
\newblock \emph{COGNITIVE COMPUTATION}, 7\penalty0 (4):\penalty0 487--499,
  August 2015.
\newblock ISSN 1866-9956.
\newblock \doi{10.1007/s12559-014-9316-6}.
\newblock 00000 Place: ONE NEW YORK PLAZA, SUITE 4600, NEW YORK, NY, UNITED
  STATES Publisher: SPRINGER Type: Article.

\bibitem[Manuja and Garg(2015)]{manuja_intelligent_2015}
Manoj Manuja and Deepak Garg.
\newblock Intelligent text classification system based on self-administered
  ontology.
\newblock \emph{TURKISH JOURNAL OF ELECTRICAL ENGINEERING AND COMPUTER
  SCIENCES}, 23\penalty0 (5):\penalty0 1393--1404, 2015.
\newblock ISSN 1300-0632.
\newblock \doi{10.3906/elk-1305-112}.
\newblock 00000 Place: ATATURK BULVARI NO 221, KAVAKLIDERE, ANKARA, 00000,
  TURKEY Publisher: TUBITAK SCIENTIFIC \& TECHNICAL RESEARCH COUNCIL TURKEY
  Type: Article.

\bibitem[Yilmaz(2017)]{yilmaz_matching_2017}
Ozgun Yilmaz.
\newblock Matching points of interest with user context: an {ANN} approach.
\newblock \emph{TURKISH JOURNAL OF ELECTRICAL ENGINEERING AND COMPUTER
  SCIENCES}, 25\penalty0 (4):\penalty0 2784--2795, 2017.
\newblock ISSN 1300-0632.
\newblock \doi{10.3906/elk-1503-61}.
\newblock 00000 Place: ATATURK BULVARI NO 221, KAVAKLIDERE, ANKARA, 00000,
  TURKEY Publisher: TUBITAK SCIENTIFIC \& TECHNICAL RESEARCH COUNCIL TURKEY
  Type: Article.

\bibitem[Evert et~al.(2019)Evert, Heinrich, Henselmann, Rabenstein, Scherr,
  Schmitt, and Schroeder]{evert_combining_2019}
Stefan Evert, Philipp Heinrich, Klaus Henselmann, Ulrich Rabenstein, Elisabeth
  Scherr, Martin Schmitt, and Lutz Schroeder.
\newblock Combining {Machine} {Learning} and {Semantic} {Features} in the
  {Classification} of {Corporate} {Disclosures}.
\newblock \emph{JOURNAL OF LOGIC LANGUAGE AND INFORMATION}, 28\penalty0 (2,
  SI):\penalty0 309--330, June 2019.
\newblock ISSN 0925-8531.
\newblock \doi{10.1007/s10849-019-09283-6}.
\newblock 00000 Place: VAN GODEWIJCKSTRAAT 30, 3311 GZ DORDRECHT, NETHERLANDS
  Publisher: SPRINGER Type: Article.

\bibitem[Greenbaum et~al.(2019)Greenbaum, Jernite, Halpern, Calder, Nathanson,
  Sontag, and Horng]{greenbaum_improving_2019}
Nathaniel~R. Greenbaum, Yacine Jernite, Yoni Halpern, Shelley Calder, Larry~A.
  Nathanson, David~A. Sontag, and Steven Horng.
\newblock Improving documentation of presenting problems in the emergency
  department using a domain-specific ontology and machine learning-driven user
  interfaces.
\newblock \emph{International Journal of Medical Informatics}, 132:\penalty0
  103981, 2019.
\newblock ISSN 1386-5056.
\newblock \doi{https://doi.org/10.1016/j.ijmedinf.2019.103981}.

\bibitem[Mendez et~al.(2019)Mendez, Cotos-Yanez, and
  Ruano-Ordas]{mendez_new_2019}
Jose~R. Mendez, Tomas~R. Cotos-Yanez, and David Ruano-Ordas.
\newblock A new semantic-based feature selection method for spam filtering.
\newblock \emph{APPLIED SOFT COMPUTING}, 76:\penalty0 89--104, March 2019.
\newblock ISSN 1568-4946.
\newblock \doi{10.1016/j.asoc.2018.12.008}.
\newblock 00000 Place: RADARWEG 29, 1043 NX AMSTERDAM, NETHERLANDS Publisher:
  ELSEVIER Type: Article.

\bibitem[Radovanovic et~al.(2019)Radovanovic, Delibasic, Jovanovic, Vukicevic,
  and Suknovic]{radovanovic_framework_2019}
Sandro Radovanovic, Boris Delibasic, Milos Jovanovic, Milan Vukicevic, and
  Milija Suknovic.
\newblock A {Framework} for {Integrating} {Domain} {Knowledge} in {Logistic}
  {Regression} with {Application} to {Hospital} {Readmission} {Prediction}.
\newblock \emph{INTERNATIONAL JOURNAL ON ARTIFICIAL INTELLIGENCE TOOLS},
  28\penalty0 (6, SI), September 2019.
\newblock ISSN 0218-2130.
\newblock \doi{10.1142/S0218213019600066}.
\newblock 00000 Place: 5 TOH TUCK LINK, SINGAPORE 596224, SINGAPORE Publisher:
  WORLD SCIENTIFIC PUBL CO PTE LTD Type: Article.

\bibitem[Nayak et~al.(2021)Nayak, Zaveri, Serrano, and
  Dumontier]{nayak_experience_2021}
Stuti Nayak, Amrapali Zaveri, Pedro~Hernandez Serrano, and Michel Dumontier.
\newblock Experience: {Automated} {Prediction} of {Experimental} {Metadata}
  from {Scientific} {Publications}.
\newblock \emph{J. Data and Information Quality}, 13\penalty0 (4), August 2021.
\newblock ISSN 1936-1955.
\newblock \doi{10.1145/3451219}.
\newblock Place: New York, NY, USA Publisher: Association for Computing
  Machinery.

\bibitem[Zhou et~al.(2021)Zhou, Svetashova, Gusmao, Soylu, Cheng, Mikut,
  Waaler, and Kharlamov]{zhou_semml_2021}
Baifan Zhou, Yulia Svetashova, Andre Gusmao, Ahmet Soylu, Gong Cheng, Ralf
  Mikut, Arild Waaler, and Evgeny Kharlamov.
\newblock {SemML}: {Facilitating} development of {ML} models for condition
  monitoring with semantics.
\newblock \emph{Journal of Web Semantics}, 71:\penalty0 100664, 2021.
\newblock ISSN 1570-8268.
\newblock \doi{https://doi.org/10.1016/j.websem.2021.100664}.

\bibitem[Deepak et~al.(2022)Deepak, Surya, Trivedi, Kumar, Lingampalli, and
  vijayan]{deepak_artificially_2022}
Gerard Deepak, Deepak Surya, Ishdutt Trivedi, Ayush Kumar, Amrutha Lingampalli,
  and Santhana vijayan.
\newblock An artificially intelligent approach for automatic speech processing
  based on triune ontology and adaptive tribonacci deep neural networks.
\newblock \emph{Computers \& Electrical Engineering}, 98:\penalty0 107736,
  2022.
\newblock ISSN 0045-7906.
\newblock \doi{https://doi.org/10.1016/j.compeleceng.2022.107736}.

\bibitem[Akand et~al.(2007)Akand, Bain, and Temple]{akand_learning_2007}
Elma Akand, Michael Bain, and Mark Temple.
\newblock Learning from {Ontological} {Annotation}: {An} {Application} of
  {Formal} {Concept} {Analysis} to {Feature} {Construction} in the {Gene}
  {Ontology}.
\newblock In \emph{Proceedings of the {Third} {Australasian} {Workshop} on
  {Advances} in {Ontologies} - {Volume} 85}, {AOW} '07, pages 15--23, AUS,
  2007. Australian Computer Society, Inc.
\newblock ISBN 978-1-920682-66-8.

\bibitem[Radinsky et~al.(2012)Radinsky, Davidovich, and
  Markovitch]{radinsky_learning_2012}
Kira Radinsky, Sagie Davidovich, and Shaul Markovitch.
\newblock Learning to {Predict} from {Textual} {Data}.
\newblock \emph{JOURNAL OF ARTIFICIAL INTELLIGENCE RESEARCH}, 45:\penalty0
  641--684, 2012.
\newblock ISSN 1076-9757.
\newblock \doi{10.1613/jair.3865}.
\newblock 00000 Place: USC INFORMATION SCIENCES INST, 4676 ADMIRALITY WAY,
  MARINA DEL REY, CA 90292-6695 USA Publisher: AI ACCESS FOUNDATION Type:
  Article.

\bibitem[Pérez-Pérez et~al.(2021)Pérez-Pérez, Igrejas, Fdez-Riverola, and
  Lourenço]{perez-perez_framework_2021}
Martín Pérez-Pérez, Gilberto Igrejas, Florentino Fdez-Riverola, and Anália
  Lourenço.
\newblock A framework to extract biomedical knowledge from gluten-related
  tweets: {The} case of dietary concerns in digital era.
\newblock \emph{Artificial Intelligence in Medicine}, 118:\penalty0 102131,
  2021.
\newblock ISSN 0933-3657.
\newblock \doi{https://doi.org/10.1016/j.artmed.2021.102131}.

\bibitem[Ali et~al.(2019)Ali, Kwak, Khan, El-Sappagh, Ali, Ullah, Kim, and
  Kwak]{ali_transportation_2019}
Farman Ali, Daehan Kwak, Pervez Khan, Shaker El-Sappagh, Amjad Ali, Sana Ullah,
  Kye~Hyun Kim, and Kyung-Sup Kwak.
\newblock Transportation sentiment analysis using word embedding and
  ontology-based topic modeling.
\newblock \emph{KNOWLEDGE-BASED SYSTEMS}, 174:\penalty0 27--42, June 2019.
\newblock ISSN 0950-7051.
\newblock \doi{10.1016/j.knosys.2019.02.033}.
\newblock 00000 Place: RADARWEG 29, 1043 NX AMSTERDAM, NETHERLANDS Publisher:
  ELSEVIER Type: Article.

\bibitem[Gaur et~al.(2019)Gaur, Alambo, Sain, Kursuncu, Thirunarayan, Kavuluru,
  Sheth, Welton, and Pathak]{gaur_knowledge-aware_2019}
Manas Gaur, Amanuel Alambo, Joy~Prakash Sain, Ugur Kursuncu, Krishnaprasad
  Thirunarayan, Ramakanth Kavuluru, Amit Sheth, Randy Welton, and Jyotishman
  Pathak.
\newblock Knowledge-{Aware} {Assessment} of {Severity} of {Suicide} {Risk} for
  {Early} {Intervention}.
\newblock In \emph{The {World} {Wide} {Web} {Conference}}, {WWW} '19, pages
  514--525, New York, NY, USA, 2019. Association for Computing Machinery.
\newblock ISBN 978-1-4503-6674-8.
\newblock \doi{10.1145/3308558.3313698}.

\bibitem[Moussallem et~al.(2019)Moussallem, Ngonga~Ngomo, Buitelaar, and
  Arcan]{moussallem_utilizing_2019}
Diego Moussallem, Axel-Cyrille Ngonga~Ngomo, Paul Buitelaar, and Mihael Arcan.
\newblock Utilizing {Knowledge} {Graphs} for {Neural} {Machine} {Translation}
  {Augmentation}.
\newblock In \emph{Proceedings of the 10th {International} {Conference} on
  {Knowledge} {Capture}}, K-{CAP} '19, pages 139--146, New York, NY, USA, 2019.
  Association for Computing Machinery.
\newblock ISBN 978-1-4503-7008-0.
\newblock \doi{10.1145/3360901.3364423}.

\bibitem[Hassanzadeh et~al.(2020)Hassanzadeh, Karimi, and
  Nguyen]{hassanzadeh_matching_2020}
Hamed Hassanzadeh, Sarvnaz Karimi, and Anthony Nguyen.
\newblock Matching patients to clinical trials using semantically enriched
  document representation.
\newblock \emph{Journal of Biomedical Informatics}, 105:\penalty0 103406, 2020.
\newblock ISSN 1532-0464.
\newblock \doi{https://doi.org/10.1016/j.jbi.2020.103406}.

\bibitem[Ren et~al.(2020)Ren, Wang, and Liu]{ren_information_2020}
Jin Ren, Hengsheng Wang, and Tong Liu.
\newblock Information {Retrieval} {Based} on {Knowledge}-{Enhanced} {Word}
  {Embedding} {Through} {Dialog}: {A} {Case} {Study}.
\newblock \emph{INTERNATIONAL JOURNAL OF COMPUTATIONAL INTELLIGENCE SYSTEMS},
  13\penalty0 (1):\penalty0 275--290, 2020.
\newblock ISSN 1875-6891.
\newblock \doi{10.2991/ijcis.d.200310.002}.
\newblock 00000 Place: 29 AVENUE LAUMIERE, PARIS, 75019, FRANCE Publisher:
  ATLANTIS PRESS Type: Article.

\bibitem[Alexandridis et~al.(2021)Alexandridis, Aliprantis, Michalakis,
  Korovesis, Tsantilas, and Caridakis]{alexandridis_knowledge-based_2021}
Georgios Alexandridis, John Aliprantis, Konstantinos Michalakis, Konstantinos
  Korovesis, Panagiotis Tsantilas, and George Caridakis.
\newblock A {Knowledge}-{Based} {Deep} {Learning} {Architecture} for
  {Aspect}-{Based} {Sentiment} {Analysis}.
\newblock \emph{INTERNATIONAL JOURNAL OF NEURAL SYSTEMS}, 31\penalty0 (10),
  October 2021.
\newblock ISSN 0129-0657.
\newblock \doi{10.1142/S0129065721500465}.

\bibitem[Niu et~al.(2022)Niu, Lu, Peng, and Zeng]{niu_fusion_2022}
Ke~Niu, You Lu, Xueping Peng, and Jingni Zeng.
\newblock Fusion of sequential visits and medical ontology for mortality
  prediction.
\newblock \emph{Journal of Biomedical Informatics}, 127:\penalty0 104012, 2022.
\newblock ISSN 1532-0464.
\newblock \doi{https://doi.org/10.1016/j.jbi.2022.104012}.

\bibitem[Qiu et~al.(2019)Qiu, Xie, Wu, and Li]{qiu_geoscience_2019}
Qinjun Qiu, Zhong Xie, Liang Wu, and Wenjia Li.
\newblock Geoscience keyphrase extraction algorithm using enhanced word
  embedding.
\newblock \emph{EXPERT SYSTEMS WITH APPLICATIONS}, 125:\penalty0 157--169, July
  2019.
\newblock ISSN 0957-4174.
\newblock \doi{10.1016/j.eswa.2019.02.001}.
\newblock 00000 Place: THE BOULEVARD, LANGFORD LANE, KIDLINGTON, OXFORD OX5
  1GB, ENGLAND Publisher: PERGAMON-ELSEVIER SCIENCE LTD Type: Article.

\bibitem[Ali et~al.(2020)Ali, El-Sappagh, Islam, Kwak, Ali, Imran, and
  Kwak]{ali_smart_2020}
Farman Ali, Shaker El-Sappagh, S.~M.~Riazul Islam, Daehan Kwak, Amjad Ali,
  Muhammad Imran, and Kyung-Sup Kwak.
\newblock A smart healthcare monitoring system for heart disease prediction
  based on ensemble deep learning and feature fusion.
\newblock \emph{INFORMATION FUSION}, 63:\penalty0 208--222, November 2020.
\newblock ISSN 1566-2535.
\newblock \doi{10.1016/j.inffus.2020.06.008}.
\newblock 00000 Place: RADARWEG 29, 1043 NX AMSTERDAM, NETHERLANDS Publisher:
  ELSEVIER Type: Article.

\bibitem[Liebowitz(1997)]{liebowitz_handbook_1997}
Jay Liebowitz.
\newblock \emph{The {Handbook} of {Applied} {Expert} {Systems}}.
\newblock CRC Press, 1997.
\newblock ISBN 978-0-429-61249-7.

\bibitem[Khan et~al.(2013)Khan, Doucette, and Cohen]{khan_validation_2013}
Atif Khan, John~A. Doucette, and Robin Cohen.
\newblock Validation of an {Ontological} {Medical} {Decision} {Support}
  {System} for {Patient} {Treatment} {Using} a {Repository} of {Patient}
  {Data}: {Insights} into the {Value} of {Machine} {Learning}.
\newblock \emph{ACM TRANSACTIONS ON INTELLIGENT SYSTEMS AND TECHNOLOGY},
  4\penalty0 (4), September 2013.
\newblock ISSN 2157-6904.
\newblock \doi{10.1145/2508037.2508049}.
\newblock 00000 Place: 2 PENN PLAZA, STE 701, NEW YORK, NY 10121-0701 USA
  Publisher: ASSOC COMPUTING MACHINERY Type: Article.

\bibitem[Bischof et~al.(2018)Bischof, Harth, Kaempgen, Polleres, and
  Schneider]{bischof_enriching_2018}
Stefan Bischof, Andreas Harth, Benedikt Kaempgen, Axel Polleres, and Patrik
  Schneider.
\newblock Enriching integrated statistical open city data by combining
  equational knowledge and missing value imputation.
\newblock \emph{JOURNAL OF WEB SEMANTICS}, 48:\penalty0 22--47, January 2018.
\newblock ISSN 1570-8268.
\newblock \doi{10.1016/j.websem.2017.09.003}.
\newblock 00000 Place: RADARWEG 29, 1043 NX AMSTERDAM, NETHERLANDS Publisher:
  ELSEVIER Type: Article.

\bibitem[Zarchi et~al.(2014)Zarchi, Monadjemi, and
  Jamshidi]{zarchi_semantic_2014}
Mohsen~Sardari Zarchi, Amirhasan Monadjemi, and Kamal Jamshidi.
\newblock A semantic model for general purpose content-based image retrieval
  systems.
\newblock \emph{Computers \& Electrical Engineering}, 40\penalty0 (7):\penalty0
  2062--2071, 2014.
\newblock ISSN 0045-7906.
\newblock \doi{https://doi.org/10.1016/j.compeleceng.2014.07.008}.

\bibitem[Ye et~al.(2015{\natexlab{b}})Ye, Li, Xu, Liu, and
  Chang]{ye_eventnet_2015}
Guangnan Ye, Yitong Li, Hongliang Xu, Dong Liu, and Shih-Fu Chang.
\newblock {EventNet}: {A} {Large} {Scale} {Structured} {Concept} {Library} for
  {Complex} {Event} {Detection} in {Video}.
\newblock In \emph{Proceedings of the 23rd {ACM} {International} {Conference}
  on {Multimedia}}, {MM} '15, pages 471--480, New York, NY, USA,
  2015{\natexlab{b}}. Association for Computing Machinery.
\newblock ISBN 978-1-4503-3459-4.
\newblock \doi{10.1145/2733373.2806221}.

\bibitem[Donadello and Serafini(2016)]{donadello_integration_2016}
Ivan Donadello and Luciano Serafini.
\newblock Integration of numeric and symbolic information for semantic image
  interpretation.
\newblock \emph{INTELLIGENZA ARTIFICIALE}, 10\penalty0 (1):\penalty0 33--47,
  2016.
\newblock ISSN 1724-8035.
\newblock \doi{10.3233/IA-160093}.
\newblock 00000 Place: NIEUWE HEMWEG 6B, 1013 BG AMSTERDAM, NETHERLANDS
  Publisher: IOS PRESS Type: Article.

\bibitem[Palazzo et~al.(2021)Palazzo, Murabito, Pino, Rundo, Giordano, Shah,
  and Spampinato]{palazzo_exploiting_2021}
S.~Palazzo, F.~Murabito, C.~Pino, F.~Rundo, D.~Giordano, M.~Shah, and
  C.~Spampinato.
\newblock Exploiting structured high-level knowledge for domain-specific visual
  classification.
\newblock \emph{PATTERN RECOGNITION}, 112, April 2021.
\newblock ISSN 0031-3203.
\newblock \doi{10.1016/j.patcog.2020.107806}.
\newblock 00001 Place: THE BOULEVARD, LANGFORD LANE, KIDLINGTON, OXFORD OX5
  1GB, OXON, ENGLAND Publisher: ELSEVIER SCI LTD Type: Article.

\bibitem[Wang et~al.(2010)Wang, Wu, Liu, and Gao]{wang_knowledge_2010}
Jun Wang, Yunpeng Wu, Xuening Liu, and Xiaoying Gao.
\newblock Knowledge acquisition method from domain text based on theme logic
  model and artificial neural network.
\newblock \emph{Expert Systems with Applications}, 37\penalty0 (1):\penalty0
  267--275, 2010.
\newblock ISSN 0957-4174.
\newblock \doi{https://doi.org/10.1016/j.eswa.2009.05.009}.

\bibitem[Wang et~al.(2021{\natexlab{b}})Wang, Mao, Wu, Xu, Jiang, and
  Yin]{wang_atc_2021}
Xuan Wang, Yi~Mao, Xiaoyong Wu, Qucheng Xu, Weiyu Jiang, and Suwan Yin.
\newblock An {ATC} instruction processing-based trajectory prediction algorithm
  designing.
\newblock \emph{NEURAL COMPUTING \& APPLICATIONS}, January 2021{\natexlab{b}}.
\newblock ISSN 0941-0643.
\newblock \doi{10.1007/s00521-021-05713-4}.
\newblock 00000 Place: 236 GRAYS INN RD, 6TH FLOOR, LONDON WC1X 8HL, ENGLAND
  Publisher: SPRINGER LONDON LTD Type: Article; Early Access.

\bibitem[Keyarsalan and Montazer(2011)]{keyarsalan_designing_2011}
Maryam Keyarsalan and Gholam~Ali Montazer.
\newblock Designing an intelligent ontological system for traffic light control
  in isolated intersections.
\newblock \emph{Engineering Applications of Artificial Intelligence},
  24\penalty0 (8):\penalty0 1328--1339, 2011.
\newblock ISSN 0952-1976.
\newblock \doi{https://doi.org/10.1016/j.engappai.2011.03.005}.

\bibitem[Patel et~al.(2021)Patel, Merlino, Bruneo, Puliafito, Vyas, and
  Ojha]{patel_video_2021}
Ashish~Singh Patel, Giovanni Merlino, Dario Bruneo, Antonio Puliafito, O.~P.
  Vyas, and Muneendra Ojha.
\newblock Video representation and suspicious event detection using semantic
  technologies.
\newblock \emph{SEMANTIC WEB}, 12\penalty0 (3):\penalty0 467--491, 2021.
\newblock ISSN 1570-0844.
\newblock \doi{10.3233/SW-200393}.
\newblock 00000 Place: NIEUWE HEMWEG 6B, 1013 BG AMSTERDAM, NETHERLANDS
  Publisher: IOS PRESS Type: Article.

\bibitem[Rosaci(2007)]{rosaci_cilios_2007}
D.~Rosaci.
\newblock {CILIOS}: {Connectionist} inductive learning and inter-ontology
  similarities for recommending information agents.
\newblock \emph{Information Systems}, 32\penalty0 (6):\penalty0 793--825, 2007.
\newblock ISSN 0306-4379.
\newblock \doi{https://doi.org/10.1016/j.is.2006.06.003}.

\bibitem[del Rincon et~al.(2013)del Rincon, Santofimia, and
  Nebel]{del_rincon_common-sense_2013}
Jesus~Martinez del Rincon, Maria~J. Santofimia, and Jean-Christophe Nebel.
\newblock Common-sense reasoning for human action recognition.
\newblock \emph{PATTERN RECOGNITION LETTERS}, 34\penalty0 (15, SI):\penalty0
  1849--1860, November 2013.
\newblock ISSN 0167-8655.
\newblock \doi{10.1016/j.patrec.2012.10.020}.
\newblock 00000 Place: RADARWEG 29, 1043 NX AMSTERDAM, NETHERLANDS Publisher:
  ELSEVIER Type: Article.

\bibitem[Patri et~al.(2016)Patri, Panangadan, Sorathia, and
  Prasanna]{patri_sensors_2016}
Om~Prasad Patri, Anand~V. Panangadan, Vikrambhai~S. Sorathia, and Viktor~K.
  Prasanna.
\newblock Sensors to {Events}: {Semantic} {Modeling} and {Recognition} of
  {Events} from {Data} {Streams}.
\newblock \emph{INTERNATIONAL JOURNAL OF SEMANTIC COMPUTING}, 10\penalty0 (4,
  SI):\penalty0 461--501, December 2016.
\newblock ISSN 1793-351X.
\newblock \doi{10.1142/S1793351X16400171}.
\newblock 00000 Place: 5 TOH TUCK LINK, SINGAPORE 596224, SINGAPORE Publisher:
  WORLD SCIENTIFIC PUBL CO PTE LTD Type: Article.

\bibitem[Mitchell et~al.(2018)Mitchell, Cohen, Hruschka, Talukdar, Yang,
  Betteridge, Carlson, Dalvi, Gardner, Kisiel, Krishnamurthy, Lao, Mazaitis,
  Mohamed, Nakashole, Platanios, Ritter, Samadi, Settles, Wang, Wijaya, Gupta,
  Chen, Saparov, Greaves, and Welling]{mitchell_never-ending_2018}
T.~Mitchell, W.~Cohen, E.~Hruschka, P.~Talukdar, B.~Yang, J.~Betteridge,
  A.~Carlson, B.~Dalvi, M.~Gardner, B.~Kisiel, J.~Krishnamurthy, N.~Lao,
  K.~Mazaitis, T.~Mohamed, N.~Nakashole, E.~Platanios, A.~Ritter, M.~Samadi,
  B.~Settles, R.~Wang, D.~Wijaya, A.~Gupta, X.~Chen, A.~Saparov, M.~Greaves,
  and J.~Welling.
\newblock Never-{Ending} {Learning}.
\newblock \emph{Commun. ACM}, 61\penalty0 (5):\penalty0 103--115, April 2018.
\newblock ISSN 0001-0782.
\newblock \doi{10.1145/3191513}.
\newblock Place: New York, NY, USA Publisher: Association for Computing
  Machinery.

\bibitem[Silva et~al.(2018)Silva, Pereira, and Götz]{silva_context-aware_2018}
Márcio J.~da Silva, Carlos~E. Pereira, and Marcelo Götz.
\newblock Context-{Aware} {Recommendation} for {Industrial} {Alarm} {System}.
\newblock \emph{IFAC-PapersOnLine}, 51\penalty0 (11):\penalty0 229--234, 2018.
\newblock ISSN 2405-8963.
\newblock \doi{https://doi.org/10.1016/j.ifacol.2018.08.266}.

\bibitem[Shi et~al.(2019)Shi, Ji, Gao, Gao, Wang, Liao, and
  Shi]{shi_ontology-based_2019}
Jianjun Shi, Weixing Ji, Zhiwei Gao, Yujin Gao, Yizhuo Wang, Xinyi Liao, and
  Feng Shi.
\newblock Ontology-based code snippets management in a cloud environment.
\newblock \emph{JOURNAL OF AMBIENT INTELLIGENCE AND HUMANIZED COMPUTING},
  10\penalty0 (8, SI):\penalty0 2971--2985, August 2019.
\newblock ISSN 1868-5137.
\newblock \doi{10.1007/s12652-018-0701-y}.
\newblock 00000 Place: TIERGARTENSTRASSE 17, D-69121 HEIDELBERG, GERMANY
  Publisher: SPRINGER HEIDELBERG Type: Article.

\bibitem[Zhang et~al.(2019)Zhang, Wang, Zhu, Wang, and Ghei]{zhang_hybrid_2019}
Wei Zhang, Meng Wang, Yanchun Zhu, Jian Wang, and Nasor Ghei.
\newblock A hybrid neural network approach for fine-grained emotion
  classification and computing.
\newblock \emph{JOURNAL OF INTELLIGENT \& FUZZY SYSTEMS}, 37\penalty0
  (3):\penalty0 3081--3091, 2019.
\newblock ISSN 1064-1246.
\newblock \doi{10.3233/JIFS-179111}.
\newblock 00000 Place: NIEUWE HEMWEG 6B, 1013 BG AMSTERDAM, NETHERLANDS
  Publisher: IOS PRESS Type: Article.

\bibitem[Zhou et~al.(2019)Zhou, Yan, Liu, and Xin]{zhou_hybrid_2019}
Qiang Zhou, Ping Yan, Huayi Liu, and Yang Xin.
\newblock A hybrid fault diagnosis method for mechanical components based on
  ontology and signal analysis.
\newblock \emph{JOURNAL OF INTELLIGENT MANUFACTURING}, 30\penalty0
  (4):\penalty0 1693--1715, April 2019.
\newblock ISSN 0956-5515.
\newblock \doi{10.1007/s10845-017-1351-1}.
\newblock 00000 Place: VAN GODEWIJCKSTRAAT 30, 3311 GZ DORDRECHT, NETHERLANDS
  Publisher: SPRINGER Type: Article.

\bibitem[Woensel et~al.(2020)Woensel, Roy, Abidi, and
  Abidi]{woensel_indoor_2020}
William~Van Woensel, Patrice~C. Roy, Syed Sibte~Raza Abidi, and Samina~Raza
  Abidi.
\newblock Indoor location identification of patients for directing virtual
  care: {An} {AI} approach using machine learning and knowledge-based methods.
\newblock \emph{Artificial Intelligence in Medicine}, 108:\penalty0 101931,
  2020.
\newblock ISSN 0933-3657.
\newblock \doi{https://doi.org/10.1016/j.artmed.2020.101931}.

\bibitem[Cheng and Chen(2021)]{cheng_location_2021}
Ruozhen Cheng and Jing Chen.
\newblock A location conversion method for roads through deep learning-based
  semantic matching and simplified qualitative direction knowledge
  representation.
\newblock \emph{Engineering Applications of Artificial Intelligence},
  104:\penalty0 104400, 2021.
\newblock ISSN 0952-1976.
\newblock \doi{https://doi.org/10.1016/j.engappai.2021.104400}.

\bibitem[Foo et~al.(2021)Foo, Kara, and Pagnucco]{foo_screw_2021}
Gwendolyn Foo, Sami Kara, and Maurice Pagnucco.
\newblock Screw detection for disassembly of electronic waste using reasoning
  and re-training of a deep learning model.
\newblock \emph{Procedia CIRP}, 98:\penalty0 666--671, 2021.
\newblock ISSN 2212-8271.
\newblock \doi{https://doi.org/10.1016/j.procir.2021.01.172}.

\bibitem[Chung et~al.(2020)Chung, Yoo, and Choe]{chung_ambient_2020}
Kyungyong Chung, Hyun Yoo, and Do-Eun Choe.
\newblock Ambient context-based modeling for health risk assessment using deep
  neural network.
\newblock \emph{JOURNAL OF AMBIENT INTELLIGENCE AND HUMANIZED COMPUTING},
  11\penalty0 (4, SI):\penalty0 1387--1395, April 2020.
\newblock ISSN 1868-5137.
\newblock \doi{10.1007/s12652-018-1033-7}.
\newblock 00000 Place: TIERGARTENSTRASSE 17, D-69121 HEIDELBERG, GERMANY
  Publisher: SPRINGER HEIDELBERG Type: Article.

\bibitem[Russo et~al.(2021)Russo, Madani, and Rinaldi]{russo_unsupervised_2021}
Cristiano Russo, Kurosh Madani, and Antonio~Maria Rinaldi.
\newblock An {Unsupervised} {Approach} for {Knowledge} {Construction} {Applied}
  to {Personal} {Robots}.
\newblock \emph{IEEE TRANSACTIONS ON COGNITIVE AND DEVELOPMENTAL SYSTEMS},
  13\penalty0 (1):\penalty0 6--15, March 2021.
\newblock ISSN 2379-8920.
\newblock \doi{10.1109/TCDS.2020.2983406}.
\newblock 00000 Place: 445 HOES LANE, PISCATAWAY, NJ 08855-4141 USA Publisher:
  IEEE-INST ELECTRICAL ELECTRONICS ENGINEERS INC Type: Article.

\bibitem[Gunning and Aha(2019)]{gunning_darpas_2019}
David Gunning and David Aha.
\newblock {DARPA}’s {Explainable} {Artificial} {Intelligence} ({XAI})
  {Program}.
\newblock \emph{AI Magazine}, 40\penalty0 (2):\penalty0 44--58, 2019.
\newblock ISSN 2371-9621.
\newblock \doi{10.1609/aimag.v40i2.2850}.

\end{thebibliography}

\end{document}